%% file: arxiv.tex
\documentclass[11pt, dvipsnames, DIV=12]{scrartcl}

\usepackage[english]{babel}
\usepackage[T1]{fontenc}
\usepackage{natbib}
\usepackage[dvipsnames]{xcolor}
\usepackage{tikz}
\usepackage{graphicx}
\usepackage{placeins}
\usetikzlibrary{arrows.meta}

\usepackage{xurl}
\usepackage{siunitx}
\usepackage{dirtytalk}
\usepackage{subcaption}
\usepackage{booktabs}
\usepackage{multirow}
\usepackage{enumitem}
\usepackage{caption}
\usepackage{mathtools}
\mathtoolsset{mathic=true}
\usepackage[all,warning]{onlyamsmath}
\usepackage{amssymb}
\usepackage{amsfonts}
\usepackage{amsthm}

\newtheorem{theorem}{Theorem}[section]
\newtheorem{lemma}[theorem]{Lemma}
\newtheorem{proposition}[theorem]{Proposition}
\newtheorem{corollary}[theorem]{Corollary}
\newcommand{\stmtlabel}[1]{\label{#1}}
\newtheorem*{restatedlemmacolored}{Lemma~\ref{lem:colored}}
\newtheorem*{restatedcoruniversality}{Corollary~\ref{cor:neural-universality}}
\theoremstyle{definition}
\newtheorem{definition}[theorem]{Definition}

\usepackage{thm-restate}
\makeatletter
\@ifundefined{newcounteralias}{}{%
  \renewcommand\thmt@autorefsetup{%
    \@xa\def\csname\thmt@envname autorefname\@xa\endcsname\@xa{\thmt@thmname}%
  }%
}
\makeatother

\usepackage{fixmath}
\usepackage{mleftright}
\usepackage{bm}

\usepackage[
    pagebackref,
    hidelinks,
    pdfdisplaydoctitle,
    pdfpagelabels,
    pdfauthor={Nadav Dym, Tal Amir, Gia Chi Dang, Christopher Morris},
    pdftitle={A Complete Geometric Message-Passing Architecture for Point Clouds},
    pdfsubject={},
    pdfkeywords={MPNNs, GNNs, SO(3), equivariance, complete, universality}
]{hyperref}

\usepackage{microtype}
\usepackage[capitalise,noabbrev]{cleveref}
\usepackage{lmodern}
\usepackage{ellipsis}
\usepackage{authblk}

\makeatletter
\def\input@path{{./}{tex/}}
\makeatother

\newcommand{\new}[1]{\emph{#1}}

\renewcommand{\vec}[1]{\boldsymbol{#1}}
\newcommand{\R}{\mathbb R}
\newcommand{\RR}{\mathbb R}
\newcommand{\Q}{\mathbb Q}
\renewcommand{\dh}{d[h]}
\newcommand{\dz}{d[z]}
\newcommand{\dE}{d[E]}
\newcommand{\dpsi}{d[\psi]}
\newcommand{\dd}{\vec{d}}

\newcommand{\SO}{\operatorname{SO}}
\newcommand{\OO}{\operatorname{O}}

\newcommand{\multiset}[1]{\{\!\!\{#1\}\!\!\}}
\newcommand{\cloud}{\mathcal X}
\newcommand{\cloudspace}{\mathcal V [3,d,n]}
\newcommand{\othercloud}{\mathcal X'}
\newcommand{\funspace}{\mathcal F}

\title{A Complete Geometric Message-Passing Architecture for Point Clouds}
\author[1]{Nadav Dym}
\author[1]{Tal Amir}
\author[2]{Gia Chi Dang}
\author[2]{Christopher Morris}

\affil[1]{Technion -- Israel Institute of Technology}
\affil[2]{RWTH Aachen University}

\date{}

\begin{document}

\maketitle

\begin{abstract}
	Three-dimensional point clouds are a fundamental data modality, and a large family of machine learning architectures has been developed to respect their natural symmetries to rigid motions and permutations. A principled way to compare these architectures, which vary widely in complexity and cost, is through their \emph{expressive power}. That is, an architecture is \new{complete} if it can distinguish any two inputs not related by the symmetries it respects. Simple distance-based message-passing networks are known to be incomplete, and while there exist complete alternatives, they typically rely on computationally intensive mechanisms like high-order representations of the rotation group or higher-order Weisfeiler--Leman-type tests.

	In this work, we devise a novel, simple, equivariant message-passing architecture for point clouds, and prove that it is complete. We complement our theoretical result with experiments showing that our architecture separates challenging point clouds that distance-based message passing provably cannot distinguish. Our results suggest that high-order equivariant features may not be necessary to enhance the expressivity of point cloud architectures, and can help settle completeness for other equivariant architectures for which such results are currently unknown. We give initial evidence for this by proving completeness of a slightly modified version of SEGNN.
\end{abstract}

\section{Introduction}
\new{Three-dimensional point clouds} are a fundamental data modality. They represent shapes in computer vision and molecules in cheminformatics and physical simulation~\citep{schutt2017schnet,qi2017pointnet}. Learning tasks defined on this modality, from shape classification and generation to the prediction of molecular properties and interatomic forces, are correspondingly of broad interest. As a result, a large and growing family of machine learning architectures has been designed for point clouds, many of them built to respect the clouds' natural symmetries, i.e., invariance or equivariance to rigid motions and to permutations of the points~\citep{thomas2018tensor,satorras2021egnn,VN,batatia2022mace,schutt2017schnet,brandstetter2022segnn}.

This proliferation of architectures, which vary widely in complexity and computational cost, raises a natural question: what, if anything, is gained by using a more expressive and more costly architecture? A principled way to address this question is to analyze an architecture's \emph{expressive power} theoretically, through the notion of \new{completeness}. An architecture is complete if it can distinguish between any two inputs that are not related by the symmetries it respects. Completeness is motivated by its close ties to approximation, i.e., an architecture that cannot separate two distinct point clouds cannot approximate a function that assigns them different values, while, conversely, a complete architecture can be shown to be a universal approximator of invariant functions~\citep{chen2019equivalence,dym2025low}. Thus, completeness is a natural requirement for general-purpose point cloud networks where universal approximation properties are desirable.

Distance-based \emph{message passing neural networks (MPNNs)} for point clouds are a family of simple and natural architectures for which  (in)completeness is well understood. Such networks achieve invariance by applying the permutation-invariant message-passing mechanism of graph neural networks to pairwise distances, a rigid-motion invariant. Concretely, MPNNs iteratively update rotation- and translation-invariant node features $\vec{h}_i^{(t)}$, $t \geq 0$, via
\begin{equation}\label{eq:dist}
	\vec{h}_i^{(t+1)} \coloneq \psi\left( \vec{h}_i^{(t)},\multiset{ (\vec{h}_j^{(t)}, \|\vec{x}_i-\vec{x}_j\|) \colon  j \in [n] } \right) \in \mathbb{R}^{d_t},
\end{equation}
$d_t > 0$, where $\vec{x}_i$ denotes the position of the $i$-th node,  $n$ denotes the number of points in the point cloud, the mapping $\psi$ is permutation-invariant, and a global, permutation-invariant representation of the point cloud is obtained by summing, or otherwise pooling, the final node features $\vec{h}_i$. Popular architectures of this kind include SchNet and Molecular Graph Convolutions~\citep{schutt2017schnet,kearnes2016molecular}.

Distance-based architectures were shown to be incomplete for three-dimensional point clouds~\citep{incompleteness1,incompleteness2}, which motivates the search for more expressive architectures. Many modern successful architectures retain the message-passing mechanism, but, besides the invariant features $\vec{h}_i^{(t)}$, they also maintain rotation-\emph{equivariant} features $\vec{y}_i^{(t)}$. For the purpose of our exposition, it is useful to divide these architectures into two families.
\begin{enumerate}
	\item \textbf{Euclidean equivariant MPNN.} Message passing networks whose equivariant features $\vec{y}_i$ are themselves vectors in $\RR^3$. These vectors are typically obtained from the previous generation's features by elementary equivariant operations, such as weighted linear combinations or cross products. Successful methods in this category include EGNN and Vector Neurons \citep{satorras2021egnn,VN}. Prior to this work, no completeness results were known for such networks.

	\item \textbf{High-Order equivariant MPNN.} These methods instead use the irreducible representations of the rotation group, which gives a rich family of equivariant features of dimension $2\ell+1$, with $\ell$ any non-negative integer. There are many methods using this approach, including Tensor field networks, SEGNNs, and MACE~\citep{thomas2018tensor,brandstetter2022segnn,batatia2022mace}. Such methods are typically more time-consuming than those of the first category, but they often attain very good results. Theoretically, the use of high-order MPNNs is justified by the fact that, for high enough order features, this construction is complete~\citep{dym2021universality,maennel2024covariants}. However, the theoretical estimates of the representation order necessary to attain this guarantee are unrealistically large~\citep{hordan2026degree}.
\end{enumerate}

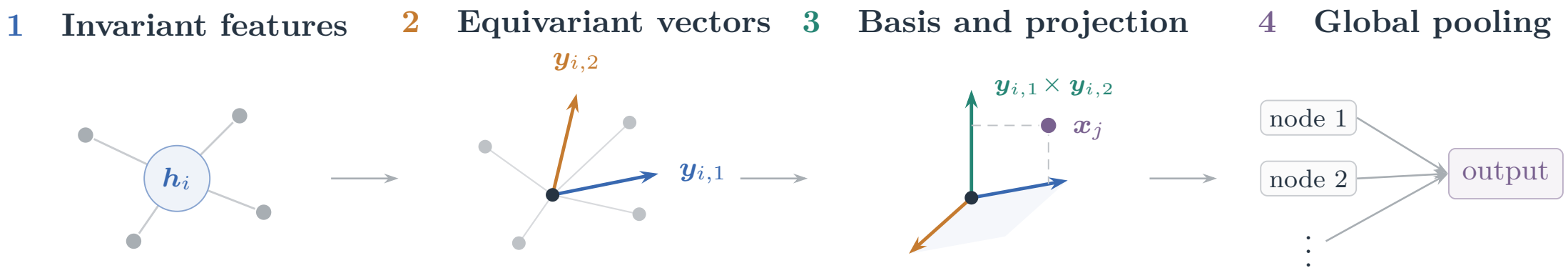
\begin{figure}
	\centering
	\resizebox{\linewidth}{!}{%
		\definecolor{ccink}{HTML}{263442}
		\definecolor{ccblue}{HTML}{3868B0}
		\definecolor{ccorange}{HTML}{C57B30}
		\definecolor{ccgreen}{HTML}{258574}
		\definecolor{ccpurple}{HTML}{79618F}

		\begin{tikzpicture}[
			x=1cm,y=1cm,text=ccink,
			font=\fontsize{8.5}{10}\selectfont,
			steptitle/.style={
					font=\fontsize{9.5}{11}\selectfont\bfseries
				},
			vector/.style={
			-{Stealth[length=1.9mm,width=1.3mm]},
			line width=1pt
			},
			flow/.style={
			-{Stealth[length=1.5mm,width=1mm]},
			draw=ccink!40,
			line width=.55pt
			}
			]
			\path (0,3.95) -- (17,3.95);

			\node[steptitle] at (2.05,3.65)
			{{\color{ccblue}1}\quad Invariant features};
			\node[steptitle] at (6.30,3.65)
			{{\color{ccorange}2}\quad Equivariant vectors};
			\node[steptitle] at (10.55,3.65)
			{{\color{ccgreen}3}\quad Basis and projection};
			\node[steptitle] at (14.80,3.65)
			{{\color{ccpurple}4}\quad Global pooling};

			\begin{scope}[shift={(2.05,2.05)}]
				\foreach \px/\py in {
						-.95/.45,.65/.65,.90/-.35,-.45/-.65
					} {
						\draw[ccink!25,line width=.6pt]
						(\px,\py) -- (0,0);
						\filldraw[
							fill=ccink!40,
							draw=white,
							line width=.5pt
						] (\px,\py) circle (2.5pt);
					}
				\node[
					circle,
					draw=ccblue!60,
					fill=ccblue!8,
					inner sep=3pt,
					text=ccblue
				] at (0,0) {$\vec{h}_i$};
			\end{scope}

			\begin{scope}[shift={(5.95,1.88)}]
				\foreach \px/\py in {
						-.70/.50,.80/.75,.90/-.20,-.35/-.50
					} {
						\draw[ccink!18,line width=.45pt]
						(0,0) -- (\px,\py);
						\fill[ccink!30]
						(\px,\py) circle (2pt);
					}
				\draw[vector,ccblue] (0,0) -- (1.10,.22)
				node[anchor=west,xshift=2pt]
				{$\vec{y}_{i,1}$};
				\draw[vector,ccorange] (0,0) -- (.25,1.05)
				node[anchor=south,yshift=2pt]
				{$\vec{y}_{i,2}$};
				\fill[ccink] (0,0) circle (2pt);
			\end{scope}

			\begin{scope}[shift={(10.30,1.85)}]
				\fill[ccblue!5]
				(0,0) -- (1.00,.18) -- (.35,-.40)
				-- (-.65,-.58) -- cycle;

				\draw[vector,ccblue]
				(0,0) -- (1.00,.18);
				\draw[vector,ccorange]
				(0,0) -- (-.65,-.58);
				\draw[vector,ccgreen]
				(0,0) -- (0,1.12);

				\draw[ccink!30,densely dashed]
				(.80,.75) -- (0,.75);
				\draw[ccink!30,densely dashed]
				(.80,.75) -- (.80,.144);

				\filldraw[
					fill=ccpurple,
					draw=white,
					line width=.5pt
				] (.80,.75) circle (2.5pt);

				\node[anchor=west,text=ccpurple]
				at (.92,.70) {$\vec{x}_j$};
				\node[
					anchor=west,
					text=ccgreen,
					font=\fontsize{7}{9}\selectfont
				] at (.10,1.15)
				{$\vec{y}_{i,1}\!\times\vec{y}_{i,2}$};

				\fill[ccink] (0,0) circle (2pt);
			\end{scope}

			\node[
				draw=ccink!25,
				rounded corners=2pt,
				fill=ccink!2,
				inner sep=2.5pt,
				font=\fontsize{7.5}{9}\selectfont
			] (p1) at (13.80,2.68) {node $1$};

			\node[
				draw=ccink!25,
				rounded corners=2pt,
				fill=ccink!2,
				inner sep=2.5pt,
				font=\fontsize{7.5}{9}\selectfont
			] (p2) at (13.80,2.05) {node $2$};

			\node[font=\fontsize{9}{10}\selectfont]
			(pn) at (13.80,1.40) {$\vdots$};

			\node[
				draw=ccpurple!50,
				rounded corners=2pt,
				fill=ccpurple!7,
				text=ccpurple,
				inner sep=4pt
			] (out) at (15.85,2.10) {output};

			\draw[flow] (p1.east) -- (out.west);
			\draw[flow] (p2.east) -- (out.west);
			\draw[flow] (pn.east) -- (out.west);

			\draw[flow] (3.65,2.05) -- (4.35,2.05);
			\draw[flow] (7.90,2.05) -- (8.60,2.05);
			\draw[flow] (12.15,2.05) -- (12.85,2.05);
		\end{tikzpicture}%
	}
	\caption{\emph{Four steps of complete geometric message passing.}
		(1) Each node aggregates information from its neighbors into
		an invariant feature $\vec{h}_i$.
		(2) Two equivariant vectors $\vec{y}_{i,1}$ and
		$\vec{y}_{i,2}$ are formed from weighted combinations of positions.
		(3) Together with their cross product
		$\vec{y}_{i,1}\times\vec{y}_{i,2}$, these vectors define a basis
		onto which every other point $\vec{x}_j$ is projected.
		(4) Pooling the invariant feature and projection obtained at
		every node gives a rotation- and permutation-invariant
		representation of the point cloud.
	}
	\label{fig:architecture-overview}
\end{figure}

\paragraph{Main contribution.} Our main contribution in this paper is a new, \emph{provably complete} and simple architecture from the Euclidean equivariant category. In its simplest form, the architecture constructs two equivariant features $\vec{y}_{i,1}$ and $\vec{y}_{i,2}$ at each index, and uses them, together with their cross product $\vec{y}_{i,1}\times\vec{y}_{i,2}$, to define a basis for $\RR^3$ onto which the remainder of the point cloud is then projected; see~\cref{fig:architecture-overview}. The key technical difficulty in proving completeness is analyzing the case where the three vectors $\vec{y}_{i,1},\vec{y}_{i,2}$ and $\vec{y}_{i,1}\times\vec{y}_{i,2}$ fail to form a basis. We show that this occurs only for balanced configurations, in the sense of~\citet{leech1957equilibrium}, and we use Leech's classification to prove completeness in this setting as well.

In addition, we show that a somewhat modified version of the \new{steerable graph neural network} (SEGNN) of
\citet{brandstetter2022segnn} can emulate our architecture with \emph{irreducible features of order one}, and thus is complete.

Finally, we complement our theoretical analysis with synthetic experiments showing that our architecture can separate challenging pairs from the literature that distance-based message passing provably cannot distinguish.

\paragraph{Significance.} We present a simple equivariant architecture with completeness guarantees. In this manuscript, we report only preliminary, toy-scale results for this architecture, but we are optimistic that it may perform well on real-world tasks. We believe, however, that the significance of this work extends beyond the specific architecture itself. Ultimately, this result is a theoretical breakthrough showing that completeness can be attained in much lower complexity than what was previously suspected. We believe our proof will help establish completeness for many other architectures for which completeness is currently unknown, either by showing that their architecture can simulate ours, or by incorporating ideas from our proof.  Our result on SEGNN is a first result in this direction.

Additionally, we have proved that Euclidean equivariant MPNNs are sufficient to approximate all invariant functions, so that at least theoretically, high-order equivariant MPNNs are not needed. This should encourage researchers to revisit whether these more complex constructions are truly needed in practice, or whether some engineering refinements could make Euclidean architectures just as effective as high-order ones.

\subsection{Related work}
We now discuss related work that has not been mentioned in the introduction.

\paragraph{Equivariant frame constructions.} Our networks are based on the construction of a local equivariant basis, or frame. This idea has appeared in works such as~\citep{luo2022orientations,du2022clofnet}. What distinguishes our work is that we can establish global separation even when the equivariant vectors remain dependent.

\paragraph{Completeness via 2-WL.} An alternative way to construct a local basis is to construct, for every pair of points $i,j$, a basis derived from the points $\vec{x}_i,\vec{x}_j$. This idea is the basis of some complete constructions for point clouds \citep{hordan2024complete,nigam2024completeness,kurlin2027atomic}, and proofs that geometric order-two Weisfeiler--Leman tests~\citep{hordan2024complete,dellerose2023wl}, and subgraph networks~\citep{li2025geongnn}, are complete. They are also used to define a frame averaging operator in \citep{dymFrames,PozdnyFrames}. While these approaches can also guarantee completeness, their disadvantage is their high complexity, stemming from the need to store and update features for every pair of nodes. In comparison, our complete model maintains features only for individual nodes.

\paragraph{Theoretical Comparison of Equivariant Architectures.}
\cite{Joshi} and \cite{sverdlov2025sparse} show that sparse geometric MPNNs which use equivariant features are more expressive than those that only rely on invariant features. Our analysis gives a similar picture with regard to completeness when using all pairwise interactions. \cite{cen2024are} argue that high-order representations are necessary for symmetric inputs, where Euclidean permutation-invariant and rotation-equivariant features vanish. The implications of these results are for specific applications where one is interested in recovering the orientation of the input. However, for tasks that involve computation of invariant functions like molecular energy or point cloud classification, our results prove that Euclidean Equivariant message passing is universal. These results can probably be extended to universality results for equivariant functions, using the work of~\citet{villar2021scalars,pmlr-v235-hordan24a}, but we leave this for future work.

\section{Background}\label{sec:background}

We introduce our notation and the
symmetry and completeness notions used throughout the paper.

\paragraph{Basic notation.}
Let $\mathbb N\coloneq\{1,2,\ldots\}$ and $[n]\coloneq\{1,\ldots,n\}$, for $n\in\mathbb N$. We write vectors and matrices in boldface using
$\vec{x}$ and $\vec{A}$, respectively; scalars and individual matrix
entries are unbolded. Vectors are columns, and tuples of vectors and
scalars denote their concatenation when used as neural-network inputs.
We use $\vec{u}\cdot\vec{v}=\vec{u}^{\top}\vec{v}$ and $\|\vec{u}\|$
for the Euclidean inner product and norm, and $\vec{I}_k$ for the
$k\times k$ identity matrix. The unit sphere is
$S^2\coloneq\{\vec{u}\in\R^3 \colon \|\vec{u}\|=1\}$.
The notation $\multiset{\cdots}$ denotes a multiset, retaining
repeated elements and their multiplicities. Function names are not
bolded, including when their outputs are vectors.

\paragraph{Attributed point clouds.}
For fixed $n,d\in\mathbb N$, a \emph{charged} or \emph{attributed point
	cloud} is an ordered tuple
\[
	\cloud=((\vec{x}_i,\vec{q}_i))_{i\in[n]}
	\in\cloudspace, \text{ where } \cloudspace \coloneq(\R^3\times\R^d)^n.
\]
Here, $\vec{x}_i\in\R^3$ denotes the position of the $i$-th point, and
$\vec{q}_i\in\R^d$  carries  rotation-invariant information on that point. We will refer to $\vec{q}_i$ as a  \emph{charge}. We write
$r_i\coloneq\|\vec{x}_i\|$ and, when $r_i>0$, call
$\vec{u}_i\coloneq\vec{x}_i/r_i$ the \new{direction} of $\vec{x}_i$.

\paragraph{Rotations, permutations, and congruence.}
A proper rotation is a matrix $\vec{R}\in\R^{3\times3} $ that is orthogonal and has a positive determinant.  The \new{group of proper rotations} in three dimensions is denoted by
$\SO(3)$. Let $S_n$ be the permutation group on $[n]$, and set
$G_n\coloneq\SO(3)\times S_n$. Its left action on point clouds applies the rotation $\vec{R}$ to all positions, while simultaneously permuting the positions and charges. Formally,
\begin{equation*}
	(\vec{R},\pi)_*\cloud
	\coloneq\bigl((\vec{R}\vec{x}_{\pi^{-1}(i)},
	\vec{q}_{\pi^{-1}(i)})\bigr)_{i\in[n]}.
\end{equation*}
A \emph{point-cloud isomorphism} from $\cloud$ to $\othercloud$ is a
pair $(\vec{R},\pi)\in G_n$ satisfying $\othercloud=(\vec{R},\pi)_*\cloud$. Two clouds are \emph{congruent}, written $\cloud\cong\othercloud$,
if such an isomorphism exists.

We note that in most practical settings, we are also interested in translation symmetry. In some settings, we are also interested in reflection symmetry. Our architecture and theory apply directly to rotations, permutations and, after centering the cloud, translations; with a straightforward adaptation to account for reflections, by averaging over $\pm\vec{X}$, they also apply to the full orthogonal group, where universality is preserved (Appendix~\ref{app:other_sym}).

\paragraph{Invariance, equivariance and geometric operations.}
A map $F$ is \emph{invariant} if $F(g_*\cloud)=F(\cloud)$, for every
$g\in G_n$. It is \emph{equivariant} if transforming the input induces
the prescribed transformation of its output.

For example, for $\vec{R}\in\SO(3)$, and vectors $\vec{u},\vec{v} \in \RR^3 $,
\[
	(\vec{R}\vec{u})\cdot(\vec{R}\vec{v})=\vec{u}\cdot\vec{v},
	\qquad
	(\vec{R}\vec{u})\times(\vec{R}\vec{v})
	=\vec{R}(\vec{u}\times\vec{v}).
\]
Consequently, inner products are invariant and cross products are
equivariant. The signed triple product
$(\vec{u}\times\vec{v})\cdot\vec{w}$ is invariant under proper
rotations and changes sign under an orthogonal transformation with
determinant $-1$. It therefore retains orientation information that
pairwise distances and inner products alone do not retain.

\paragraph{Completeness and universality.}
A family $\mathcal F$ of invariant maps is \emph{complete}, or
\emph{separates orbits}, if every noncongruent pair
$\cloud,\othercloud$ is distinguished by some map $F\in\mathcal F$.
The separating map may depend on the pair. A single invariant map
$F$ is \emph{globally separating} on $M\subseteq\cloudspace$ if
\[
	F(\cloud)=F(\othercloud)
	\quad\Longleftrightarrow\quad\cloud\cong\othercloud,
	\qquad\text{for all }\cloud,\othercloud\in M.
\]
Equivariance alone does not imply either separation property.
Finally, a family is \emph{universal} for continuous invariant functions
on a compact domain if it approximates every such function uniformly
to arbitrary accuracy. We establish family completeness first, then a
continuous globally separating neural representation, and finally
universality by adding a feed-forward neural network readout.

\section{A Complete Geometric Message-Passing Architecture for Point Clouds}\label{sec:architecture}
We define a $G_n$-invariant architecture via an all-to-all message-passing architecture on the attributed
point clouds introduced in \cref{sec:background}. We note that the architecture we introduce now is an abstract and minimal architecture for which completeness can be obtained. A concrete neural parameterization is given in
\cref{sec:realization}, including extensions to enable global separation and universality guarantees.

Writing $r_i \coloneq \|\vec{x}_i\|$, the architecture consists of four operations, i.e.
\begin{align}
	\vec{h}_i
	 & \coloneq\phi_0\Bigl(\vec{q}_i,r_i,
	\multiset{(\vec{q}_j,r_j,\vec{x}_i\cdot\vec{x}_j)\colon j\in[n]}\Bigr),
	\label{eq:h}                          \\[2pt]
	\vec{y}_{i,k}
	 & \coloneq\vec{x}_i+\sum_{j=1}^n
	\psi_k\bigl(\vec{h}_i,\vec{h}_j,\vec{x}_i\cdot\vec{x}_j\bigr)\vec{x}_j,
	\qquad k \in [2],
	\label{eq:y}                          \\[2pt]
	\vec{z}_i
	 & \coloneq\phi_1\Bigl(\vec{g}_i,
	\multiset{(\vec{h}_j,\vec{p}_{ij})\colon j\in[n]}\Bigr),
	\label{eq:z}                          \\[2pt]
	E(\cloud)
	 & \coloneq\phi_2\Bigl(
	\multiset{(\vec{h}_i,\vec{z}_i)\colon i\in[n]}\Bigr),
	\label{eq:E}
\end{align}
where the invariant descriptors $\vec{g}_i,\vec{p}_{ij}\in\R^3$ collect
the Gram entries and oriented measurements, respectively, i.e.,

\begin{align}
	\vec{g}_i
	 & \coloneq\bigl(\|\vec{y}_{i,1}\|^2,\|\vec{y}_{i,2}\|^2,
	\vec{y}_{i,1}\cdot\vec{y}_{i,2}\bigr),\label{eq:gram}     \\[2pt]
	\vec{p}_{ij}
	 & \coloneq\bigl(\vec{y}_{i,1}\cdot\vec{x}_j,
	\vec{y}_{i,2}\cdot\vec{x}_j,
	(\vec{y}_{i,1}\times\vec{y}_{i,2})\cdot\vec{x}_j\bigr). \label{eq:proj}
\end{align}

The functions $\phi_0$ and $\phi_1$ combine a vector input with a
multiset of vectors to produce a feature vector; $\phi_2$ maps a
multiset of vectors to a vector. The scalar-valued functions
$\psi_1$ and $\psi_2$ determine the message weights. All nodewise functions are shared across points, and the input multisets retain
repeated elements.

The intuition behind this construction is as follows. The first step, Equation~\eqref{eq:h}, encodes all available rotation-invariant information via a message-passing aggregation to obtain the rotation-invariant feature $\vec{h}_i $. The second step, Equation~\eqref{eq:y}, then produces two
equivariant vectors, $\vec{y}_{i,1}$ and $\vec{y}_{i,2}$. Together with their vector product $\vec{y}_{i,1}\times \vec{y}_{i,2}$, these three vectors (may) form a basis for $\RR^3$. The third step then records the properties of this basis via the Gram matrix, and the projection of the point cloud onto this basis, to obtain another rotation invariant feature $\vec{z}_i$. Finally, the last step aggregates all node invariant features to obtain a final global representation $E(\cloud)$.

We denote by $\funspace(d)$ the family of all maps $E$ obtained
from \eqref{eq:h}--\eqref{eq:E} by varying these functions and their
feature dimensions. The multiset functions accept multisets of any
finite size, so each $E\in\funspace(d)$ is defined on clouds with any
number $n$ of points.

\section{Completeness theorem}
We now state our main result, i.e., the message-passing architecture from~\cref{sec:architecture} is invariant and complete.

\begin{figure}[!htbp]
	\centering
	\input{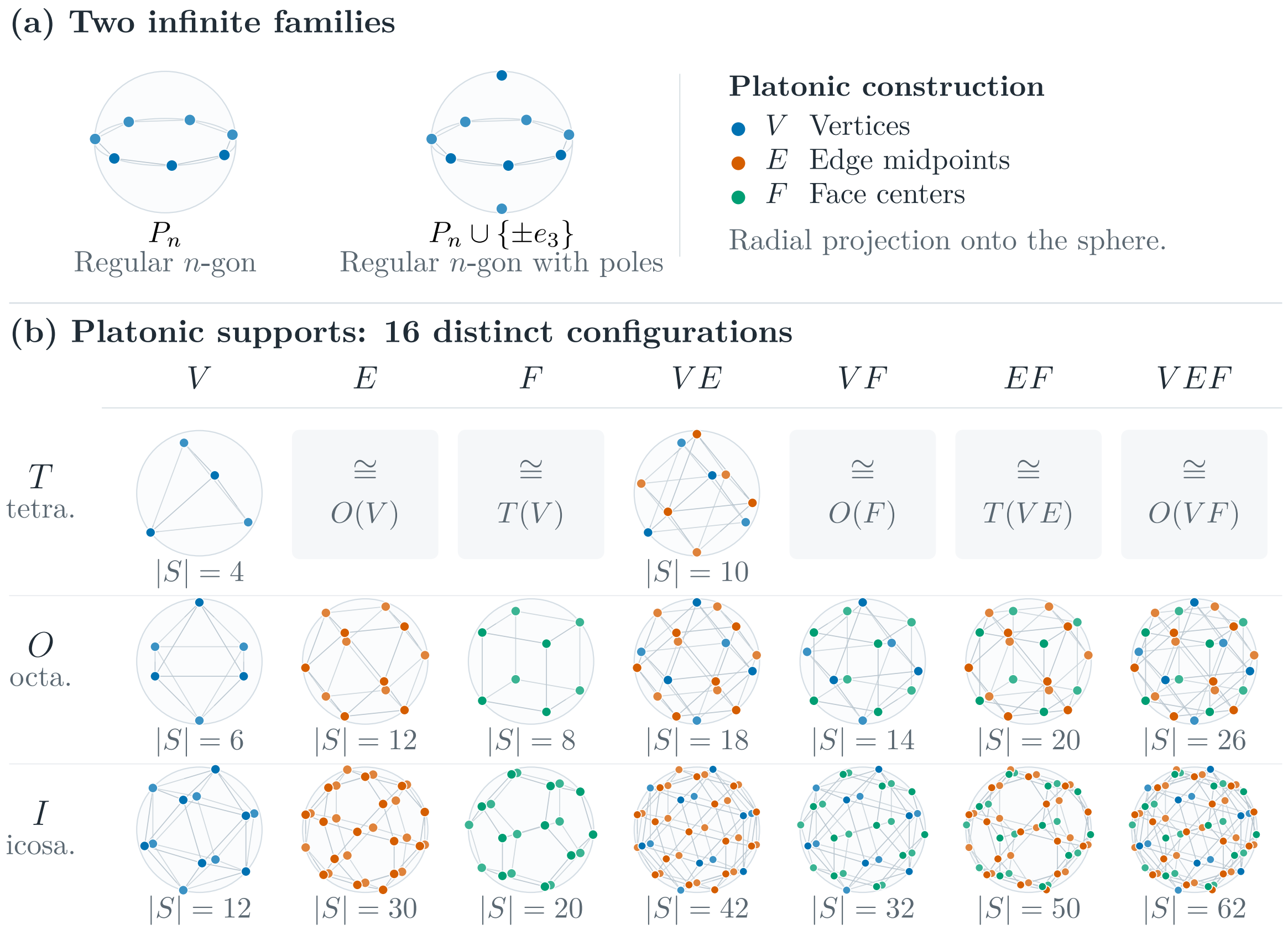}
	\caption{\emph{Leech's classification of the balanced subsets of $S^2$, up to
			congruence.} (a) Regular polygons on a great circle, with or without its
		two poles. (b) The 16 distinct
		Platonic balanced sets. Rows correspond to a tetrahedron $T$, octahedron $O$,
		or icosahedron $I$. Blue denotes vertices $V$, orange denotes edge midpoints $E$, and
		green denotes face centers $F$. All these points are projected radially onto the unit sphere; concatenation denotes union, e.g., $VE=V\cup E$. The 21 descriptions include five duplicates, identified by congruence labels in shaded
		cells.}
	\label{fig:leech-classification}
\end{figure}

\begin{restatable}[Charged-cloud completeness]{theorem}{chargedcloudcompleteness}\label{thm:main}
	For every fixed $d,n\geq1$, any function $E\in\funspace(d)$ is $\SO(3)\times S_n$ invariant on $\cloudspace$. Moreover, $\funspace(d)$ is complete, i.e., if $\cloud,\othercloud\in\cloudspace$ are two non-congruent point clouds, then there exists $E\in\funspace(d)$ such that
	$E(\cloud)\ne E(\othercloud)$.
\end{restatable}
\begin{proof}[Proof idea]
	The full proof appears in Appendix~\ref{app:main-proof}; here we outline its main steps.

	\par\medskip\noindent\emph{1. Equivariant linear independence is enough.}
	The first idea behind the proof is well known from frame-based architectures~\citep{luo2022orientations,du2022clofnet}. That is, if, for a given point-cloud $\cloud$, there exists some implementation of the architecture and some index $i$, such that the equivariant features $\vec{y}_{i,1}$ and $\vec{y}_{i,2}$ are linearly independent, then $\vec{y}_{i,1}, \vec{y}_{i,2},\vec{y}_{i,1}\times \vec{y}_{i,2} $ form an equivariant basis for $\RR^3$, and as a result, the network's invariant features which include the representation of $\cloud$ according to this basis, will determine $\cloud$ uniquely, up to congruence.

	\par\medskip\noindent\emph{2. Characterizing linear independence failure.} Our next step is to characterize when linear independence of $ \vec{y}_{i,1}$ and $\vec{y}_{i,2}$ cannot be attained. We first reduce to the case that $\cloud$ contains distinct points on the sphere. This is done by replacing the points of $\cloud$ on each ray by a single point on the sphere, whose new charge $\vec{q}_j$, which we call its \new{color}, is a vector encoding the radii and charges of the replaced points. Zero points are dealt with separately. Next, we show that the failure to construct an equivariant basis only occurs if $\cloud$ is \new{color-balanced}, which means that for every index $i$, and every fixed angular distance $t$ and color $\vec{q}$,

	\begin{equation*}
		\sum_{j \colon \vec{x}_i\cdot \vec{x}_j=t, \vec{q}_j=\vec{q}} \vec{x}_j\in\operatorname{span}\{\vec{x}_i\}.
	\end{equation*}
	In words, around each point $\vec{x}_i$, the points of any one color at any one distance sum to a vector on the axis through that point.

	\par\medskip\noindent\emph{3. The color-balanced case.} The final and most challenging step is showing completeness on the subset of color-balanced point clouds. This analysis is based on a characterization of \cite{leech1957equilibrium} for balanced sets. Leech shows that the class of balanced sets is restricted to three classes of symmetric point sets: (i) regular polygons in the $x$-$y$ plane, (ii) regular polygons in the $x$-$y$ plane together with the two poles $(0,0,\pm1)$, and (iii) sixteen balanced sets obtained from taking the vertices, edges, and/or faces of the Platonic solids; see \cref{fig:leech-classification}. The analysis first shows that the balanced set type is determined uniquely by the multiset of angular distances $\multiset{\vec{x}_i\cdot \vec{x}_j \colon i,j \in [n] }$, and this information can be easily retrieved from the invariant features of our network.  We then use this to show that in the color-balanced case we can guarantee equivalence up to a color-preserving congruence. This requires a case-by-case analysis of all the 16 Platonic cases, and this analysis is done computationally in exact arithmetic.
\end{proof}

\FloatBarrier

\section{Globally complete realization and universality}\label{sec:realization}

We now give a concrete realization of the architecture described in~\cref{sec:architecture}. We will prove that a single head of our architecture separates each fixed pair of non-congruent clouds, and that concatenating finitely many independently parameterized heads gives a single map that separates all orbits simultaneously. Composing this map with a feed-forward neural network then gives universal approximation of continuous invariant functions on compact sets.

The proofs of all theorems in this section are given in Appendix \ref{app:neural_proofs}. They are a rather standard adaptation of arguments developed in \cite{amir2023injective}.

\subsection{A single neural head}
We realize each multiset function by applying a one-layer neural
network to its elements and summing the outputs. For vector-multiset
pairs, we add a linear transformation of the vector component. The scalar-valued message functions $\psi_k$ have one hidden layer and a linear readout. A single neural head therefore takes the form
\begin{align}
	\vec{h}_i                & \coloneq \vec{C}_0(\vec{q}_i,r_i)+\sum_{j=1}^n\sigma\big(\vec{A}_0(\vec{q}_j,r_j,\vec{x}_i\cdot \vec{x}_j)+\vec{b}_0\big),
	\label{eq:hexplicit}                                                                                                                                                       \\
	\vec{y}_{i,k}            & \coloneq \vec{x}_i+\sum_{j=1}^n \vec{w}_k \cdot \,\sigma\big(\vec{W}_k(\vec{h}_i,\vec{h}_j,\vec{x}_i\cdot \vec{x}_j)+\vec{t}_k\big)\,\vec{x}_j,
	\qquad k \in [2],
	\label{eq:yexplicit}                                                                                                                                                       \\
	\vec{z}_i                & \coloneq \vec{C}_1\vec{g}_{i}
	+\sum_{j=1}^n\sigma\Big(\vec{A}_1\big(\vec{h}_j,\vec{p}_{i,j}\big)+\vec{b}_1\Big),
	\label{eq:zexplicit}                                                                                                                                                       \\
	E_{\vec{\theta}}(\cloud) & \coloneq \sum_{i=1}^n \sigma\big(\vec{A}_2(\vec{h}_i,\vec{z}_i)+\vec{b}_2\big),
	\label{eq:Eexplicit}
\end{align}
where $\vec{g}_i$ and $\vec{p}_{i,j} $ in \eqref{eq:zexplicit} are as defined in \eqref{eq:gram} and \eqref{eq:proj}, respectively.
Hyperparameters of this model are the activation $\sigma\colon\R\to\R$ and the dimensions  $\dd=(\dh,\dz,\dE,\dpsi)$, where the first three widths are
the dimensions of $\vec{h}_i,\vec{z}_i,E_{\vec{\theta}}$, respectively, and $\dpsi$ is the
hidden width of each scalar-valued function $\psi_k$.
The activation is applied componentwise to vector inputs. Concatenating the parameters, with matrices vectorized, gives

\[
	\vec{\theta} \coloneq (\vec{A}_0,\vec{b}_0,\vec{C}_0,\,\vec{W}_1,\vec{t}_1,\vec{w}_1,\,\vec{W}_2,\vec{t}_2,\vec{w}_2,\,
	\vec{A}_1,\vec{b}_1,\vec{C}_1,\,\vec{A}_2,\vec{b}_2),
\]
We denote the resulting Euclidean parameter space by
$\Theta \coloneq \Theta(d;\sigma,\dd)$.

\begin{restatable}[Pairwise separation]{theorem}{pairwiseseparation}\label{thm:neural-pairwise}
	Fix $n,d\geq1$, an analytic non-polynomial activation $\sigma$, and
	any positive integer widths $\dd$. For every fixed pair
	$\cloud,\othercloud\in \cloudspace$ with $\cloud\not\cong\othercloud$,
	\[
		E_{\vec{\theta}}(\cloud)\neq E_{\vec{\theta}}(\othercloud),
		\quad\text{for Lebesgue almost every }\vec{\theta}\in\Theta(d;\sigma,\dd).
	\]
\end{restatable}

\subsection{Independent heads and global separation}
Our next goal is to attain global separation, namely, an architecture for which a fixed choice of parameters separates all non-congruent pairs simultaneously. This is attained by taking several heads of the previous pairwise separating architecture. For an integer $H\geq1$, take $H$ copies of
\eqref{eq:hexplicit}--\eqref{eq:Eexplicit}, each with its own full
parameter vector $\vec{\theta}^{(a)}\in\Theta$. Thus each head computes
its own vectors $\vec{h}_i^{(a)},\vec{y}_{i,k}^{(a)},\vec{z}_i^{(a)}$; parameters are not shared
between heads. Concatenate their outputs to obtain
\begin{equation}\label{eq:multihead}
	\mathcal E^{(H)}_{\vec{\vartheta}}(\cloud)
	\coloneq \bigl(E_{\vec{\theta}^{(1)}}(\cloud),\ldots,
	E_{\vec{\theta}^{(H)}}(\cloud)\bigr)\in\R^{H\cdot \dE},
\end{equation}
where $\vec{\vartheta}\coloneq(\vec{\theta}^{(1)},\ldots,\vec{\theta}^{(H)})\in\Theta^H$. We can then prove global separation provided that the number of heads is at least $2\delta+1$, where $\delta$ is the dimension of the input point cloud domain. If we are interested in the full domain $\cloudspace$, its dimension is $\delta=n(d+3) $, and so we will require $2nd+6n+1 $ heads. However, in many cases, it may be reasonable to employ the \say{manifold} assumption, and assume that the point clouds we are interested in for some task reside in some domain $M\subseteq \cloudspace $ whose intrinsic dimension is $\delta \ll n(d+3) $. In this case, the number of heads will depend on the intrinsic dimension $\delta$ only. This intrinsic result requires us to assume that $M$ is a \emph{$\sigma$-subanalytic} set in the sense of~\cite{amir2023injective}, that is, a countable union of subanalytic sets (the prefix $\sigma$ indicates a countable union, as in $\sigma$-algebra or $\sigma$-compact).

\begin{restatable}[Global separation]{theorem}{globalseparation}\label{thm:neural-global}
	Let $M\subset \cloudspace$ be a nonempty $\sigma$-subanalytic set of
	dimension $\delta$. Fix an analytic non-polynomial activation
	$\sigma$ and any vector of positive integer widths $\dd\in \mathbb{N}^4$. Denote $\Theta=\Theta(d;\sigma,\dd)$. If
	$H\geq2\delta+1$,
	then for Lebesgue almost every $\vec{\vartheta}\in\Theta^H$, the function
	$\mathcal E^{(H)}_{\vec{\vartheta}}\colon\cloudspace\to\R^{H\cdot \dE}$ is globally separating on $M$. In other words, for such a single parameter choice $\vec{\vartheta}$,
	\[
		\mathcal E^{(H)}_{\vec{\vartheta}}(\cloud)
		=\mathcal E^{(H)}_{\vec{\vartheta}}(\othercloud)
		\quad\Longleftrightarrow\quad \cloud\cong\othercloud,
		\qquad\text{for all }\cloud,\othercloud\in M.
	\]
\end{restatable}

\subsection{Universal approximation of invariant functions}
Finally, we can obtain universal approximation of invariant functions with our architecture by concatenating it with a feed-forward neural network. This corollary uses the separation-to-factorization principle of~\cite{dym2025low}.

\newcommand{\invariantapproximationbody}{%
	Under the hypotheses of Theorem~\ref{thm:neural-global}, for almost
	every $\vec{\vartheta}\in\Theta^H$ the following holds. For every
	compact set $K\subset M$, and every continuous $\SO(3)\times S_n $ invariant function
	$f\colon K\to\R$, and every $\varepsilon>0$, there is a feed-forward neural network
	$\rho\colon\R^{H\cdot \dE}\to\R$ such that
	\[
		\sup_{\cloud\in K}
		\bigl|f(\cloud)-\rho\bigl(
		\mathcal E^{(H)}_{\vec{\vartheta}}(\cloud)\bigr)\bigr|
		<\varepsilon.
	\]}
\begin{corollary}[Invariant universal approximation]\label{cor:neural-universality}
	\invariantapproximationbody
\end{corollary}

\section{Verifying completeness for existing architectures}

To showcase the usefulness of our completeness results, we now show that it extends to an $\SO(3)$-equivariant specialization of the \new{steerable graph neural network} (SEGNN) of
\citet{brandstetter2022segnn}, which we denote by SEGNN(3). This architecture maintains invariant scalar features and equivariant vector features. Its message and update functions
combine these features with geometric inputs, called
\emph{steering attributes}, using tensor products and nonlinearities. We use the complete graph, including self-edges, and choose
the steering attributes
\[
	\vec{a}_{ij}=(1,\vec{x}_j),
	\qquad
	\vec{a}_i=(1,\vec{x}_i).
\]
The initial features consist of the scalar channels
$(1,\vec{q}_i,r_i)$ and the vector channel $\vec{x}_i$.
Let $\mathcal S_{n,d}^{(4)}$ denote the resulting family with
four message-passing layers, arbitrary finite widths, only
scalar and vector channels, and an invariant neural readout.
The following theorem shows that this family is complete.
Its proof approximates a separating realization of
\eqref{eq:h}--\eqref{eq:E} sufficiently accurately to preserve
separation. We give the full architectural specification,
simulation construction, and proof in Appendix~\ref{app:segnn}.

\begin{theorem}[Completeness of SEGNN(3)]
	\label{thm:segnn-completeness}
	For every fixed $n,d\geq1$, the family $\mathcal S_{n,d}^{(4)}$ is $G_n$-invariant and complete.
	That is, every network in this family is invariant, and for every pair of non-congruent clouds
	$\cloud,\othercloud\in\cloudspace$, there exists
	$N\in\mathcal S_{n,d}^{(4)}$ such that
	\[
		N(\cloud)\neq N(\othercloud).
	\]
	Four message-passing layers using only scalar and vector channels
	suffice. The widths and parameters may depend on the pair.
\end{theorem}

\section{Experimental results}

\begin{table}[htb!]
	\centering
	\small
	\setlength{\tabcolsep}{8pt}
	\resizebox{\textwidth}{!}{
		\renewcommand{\arraystretch}{1.12}
		\begin{tabular}{@{}lccc@{}}
			\toprule
			                        & \multicolumn{2}{c}{\textbf{\citet{li2025geongnn}}}
			                        & \textbf{\citet{incompleteness1}}                                           \\
			\cmidrule(lr){2-3}\cmidrule(l){4-4}
			\textbf{Model}
			                        & \textbf{10 isolated pairs}
			                        & \textbf{7 combined pairs}
			                        & \textbf{4 pairs}                                                           \\
			\midrule
			\multicolumn{4}{@{}l}{\textbf{Random parameters} (100 draws)}                                        \\
			\addlinespace[3pt]
			Distance-only control   & $0/1000$                                           & $0/700$   & $0/400$   \\
			\emph{Ours}             & $1000/1000$                                        & $700/700$ & $400/400$ \\
			\midrule
			\multicolumn{4}{@{}l}{\textbf{After training} (10 seeds)}                                            \\
			\addlinespace[3pt]
			GeoNGNN (released code) & $100/100$                                          & $70/70$   & $40/40$   \\
			Distance-only control   & $0/100$
			                        & $1/70$\rlap{\textsuperscript{$\dagger$}}
			                        & $0/40$                                                                     \\
			\emph{Ours}             & $100/100$                                          & $70/70$   & $40/40$   \\
			\bottomrule
		\end{tabular}}
	\caption{\emph{Results on synthetic datasets.}
		Entries report successful runs over total runs:
		separated pair-draw combinations without training (top)
		and correct pair-seed runs after training (bottom).
		GeoNGNN uses its released training schedule; the other models use ours.
		\textsuperscript{$\dagger$}\,Both logits are of order $10^{-16}$,
		so this run is decided by round-off.}
	\label{tab:results}
\end{table}

Here, we verify that the architecture described in~\cref{sec:architecture} can be trained via gradient descent to distinguish pairs of non-congruent point clouds that non-complete architectures fail to distinguish. Concretely, we address the following questions.
\begin{description}[leftmargin=2.6em,labelwidth=2.1em,itemsep=0pt,topsep=3pt]
	\item[Q1] Can the architecture distinguish these pairs of point clouds at random initialization?
	\item[Q2] Can it distinguish these pairs after training via gradient descent?
	\item[Q3] How do the layer widths affect the ability to distinguish these pairs?
\end{description}
\paragraph{Datasets.} We consider all  point cloud pairs from~\citet{li2025geongnn} and four additional pairs generated using the construction of~\citet[Equations~2--3]{incompleteness1}. These are pairs of non-congruent point clouds that provably cannot be separated by distance-based methods defined in \eqref{eq:dist}. The first set comprises 10 isolated pairs and 7 combined pairs, with point cloud sizes ranging from 6 to 40 points. For the second set, we use $P \in \{1,2,3,4\}$ repeat units and fix the parameters at $p=4$, $c_y=0.5$, $c_z=1$, $w_y=0.3$, $w_z=0.7$, $v_x=0.8$, and $v_y=0.4$. The resulting point clouds contain 6, 12, 18, and 24 points, respectively.

\begin{table}[htb!]
	\centering
	\small
	\setlength{\tabcolsep}{8pt}
	\resizebox{\textwidth}{!}{
		\renewcommand{\arraystretch}{1.12}
		\begin{tabular}{@{}lccc@{}}
			\toprule
			                          & \multicolumn{2}{c}{\textbf{\citet{li2025geongnn}}}
			                          & \textbf{\citet{incompleteness1}}                                       \\
			\cmidrule(lr){2-3}\cmidrule(l){4-4}
			\textbf{Width}
			                          & \textbf{10 isolated pairs}
			                          & \textbf{7 combined pairs}
			                          & \textbf{4 pairs}                                                       \\
			\midrule
			$4$                       & $75/100$                                           & $29/70$ & $25/40$ \\
			$8$                       & $96/100$                                           & $69/70$ & $38/40$ \\
			$16$                      & $98/100$                                           & $67/70$ & $40/40$ \\
			$32$ (\cref{tab:results}) & $100/100$                                          & $70/70$ & $40/40$ \\
			$64$                      & $100/100$                                          & $69/70$ & $40/40$ \\
			\bottomrule
		\end{tabular}}
	\caption{\emph{Effect of the width on our model.}
		Each row sets the width of the hidden layers of the feed-forward networks to a given value between 4 and 64.
		Entries report successful runs over total runs after training ($10$ seeds), with a single head and no additional readout network.
		At random parameters, every width separates every pair in all $100$ draws.}
	\label{tab:hyperparameters}
\end{table}

\paragraph{Model configuration and experimental setup.}
Our goal is to examine whether a given invariant model can separate the point cloud pairs in our dataset. We consider two settings: the first is the random parameter setting, where we take the models with random parameters, and count a pair as separated if its two outputs differ by more than $1000$ times the numerical noise, which we measure on rotated, translated, and permuted copies of the point clouds. In the second setting, we model the separation problem as a binary classification task, and train the models using standard binary classification protocol. Each random parameter (and training) experiment is repeated with 100 (respectively 10) random seeds.

In the experiment, we considered three invariant models: Our model, the released GeoNGNN implementation~\citep{li2025geongnn}, and a message-passing network that uses only pairwise distances as a negative control. We ran our model and the negative control in double precision on centered point clouds. The expected results are that the distance-based methods will fail to separate any examples, as these pairs are constructed for this purpose. In contrast, our method and GeoNGNN are both provably complete, with respect to $\SO(3)\times S_n$ and the Euclidean group $\mathrm{E}(3)\times S_n$, respectively, and so are expected to separate all inputs. The advantage of our method over GeoNGNN is that it maintains features only for individual nodes, rather than for every node pair, which are expensive to compute.

We implemented the architecture in~\cref{sec:architecture} in PyTorch~\citep{Pas+2019}. The model uses feed-forward networks with two hidden layers of width $32$ and the analytic smooth leaky ReLU activation $\sigma(x)=0.2\,x+\frac{0.8}{b}\log\bigl(1+e^{bx}\bigr)$, with $b=100$. Each of $\phi_0$, $\phi_1$, and $\phi_2$ applies one such network before summing over its multiset and another after summation. A single network produces both $\psi_1$ and $\psi_2$. The model uses a single head, whose output serves as the classification logit.

We trained the model with AdamW~\citep{Los+2019} and AMSGrad for up to $2000$ epochs, using an initial learning rate of $10^{-3}$, zero weight decay, and gradient-norm clipping at $10$. We reduced the learning rate by a factor of $0.8$ after $100$ epochs without improvement. Following~\citet{li2025geongnn}, we used cross-entropy loss and counted a run as successful if the parameters with the lowest loss classified both point clouds correctly. We stopped training early once the loss reached at most $0.01$ and both point clouds were classified correctly, and ran the experiments with $10$ random seeds. We note that generally our model requires more epochs than GeoNGNN to attain a high accuracy result.

\paragraph{Experimental results.}
\Cref{tab:results} reports the number of successful runs on each dataset, in the random parameter and trained setting. As expected, our model separates every tested pair under all $100$ random parameter draws, and both our model and GeoNGNN achieve perfect separation after training across all $10$ seeds.
By contrast, the distance-only control fails to separate any pair at random parameters and fails after training except for a single run decided by floating-point round-off.
These results show that our model can distinguish these examples without learning suitable parameters, and that this ability is retained after training, answering \textbf{Q1} and \textbf{Q2}.

We next study the influence of width on separation, \textbf{Q3}. We vary the widths between $4$ and $64$, and rerun our experiments. At random initialization, every sampled parameter setting separates every pair at all tested widths from $4$ to $64$. However, the
separation can be brittle, i.e., at width $4$, the output difference for one pair is as small as $7.7\cdot10^{-12}$, although still well above our threshold of $\,1000$ times the numerical noise. This agrees with
previous observations for graph neural networks with analytic
activations, where the predicted separation can yield feature
differences as small as $10^{-7}$~\citep{amir2023injective}
and $10^{-12}$~\citep{pmlr-v235-hordan24a}. Thus, these results are coherent with \cref{thm:neural-pairwise}, which guarantees pairwise separation for arbitrary positive feature widths (for the realization of Section~\ref{sec:realization}; the trained model uses deeper networks)
and a single head, but also suggest that practical separation requires larger widths. This conclusion is strengthened by the results after training, which are reported in \cref{tab:hyperparameters}. Here we see an overall trend of increased accuracy as widths increase.

In summary, our experiments corroborate our theoretical findings by showing that our architecture can distinguish between hard-to-separate pairs.

\section{Conclusion and future directions}
The main result of this paper is that a simple Euclidean equivariant model for point clouds is provably complete. Future directions include establishing completeness of other existing architectures by showing that they can simulate our construction, and evaluating our architecture on real-world datasets. In the long term, we would like to show that Euclidean message passing is sufficiently expressive for point cloud tasks, and can attain comparable performance with state-of-the-art architectures such as MACE and EquiformerV3, with a significantly reduced computational budget.

\paragraph{AI-usage statement.} The proof of Theorem~\ref{thm:main} was developed through interaction with Astra 6. Claude and ChatGPT were also used to assist with writing and code generation. All results were verified by the authors, who take
full responsibility for the content of this paper.

\paragraph{Acknowledgments.} CM is partially funded by an Israel ISF-DFG Grant (587125021) and the Werner Siemens Foundation within the WSS Research Center \say{catalaix}. ND and TA are supported by the Israel Science Foundation
grant~272/23 and DFG-ISF grant~3597/26.

\newpage
\bibliography{tex/bibliography}
\newpage
\appendix

\section{Proof of the completeness theorem}\label{app:main-proof}

We now prove our main result,~\cref{thm:main}. In this appendix, we prove the theorem, assuming a lemma on color-balanced sets (Lemma~\ref{lem:colored}). We then prove the lemma in Appendix \ref{sec:colored-proof}, and describe the exact arithmetic used to prove it in Appendix \ref{app:exact}.

\chargedcloudcompleteness*

\begin{proof}[Proof of Theorem~\ref{thm:main}]
	We first prove invariance.
	The scalar-channel feature $\vec{h}_i$ depends only on charges, radii, and inner products, so it is rotation invariant, while both $\vec{y}_{i,k}$ transform
	as vectors. For any rotation $\vec{R}\in\SO(3)$,
	$(\vec{R}\vec{a})\times(\vec{R}\vec{b})=\vec{R}(\vec{a}\times \vec{b})$, so all arguments of the readout functions
	are invariant. Charges remain unchanged by rotations, and permuting the points
	permutes the features accordingly and leaves every multiset unchanged. This proves
	invariance. Under an orthogonal transformation with determinant $-1$,
	such as a reflection, the signed triple product
	$(\vec{y}_{i,1}\times\vec{y}_{i,2})\cdot\vec{x}_j$ changes sign. This is
	what allows the architecture to distinguish noncongruent mirror images.

	For separation, fix $n\geq1$ and a pair of point clouds $\cloud,\othercloud\in\cloudspace$, and assume that $E(\cloud)=E(\othercloud) $, for all $E\in \funspace(d)$. We need to show that point clouds $\cloud$ and $\othercloud$ are related by a permutation and proper rotation. Throughout the proof we denote the features obtained for point cloud $\cloud$ by $\vec{h}_i,\vec{y}_{i,k},\vec{z}_i$, and so on, and those obtained for $\othercloud$ by the same symbols with a prime. Moreover, we take the functions $\phi_0,\phi_1,\phi_2$ to be injective throughout. This is possible because only their values on the finitely many inputs arising from $\cloud$ and $\othercloud$ matter. The message functions $\psi_1,\psi_2$ are chosen separately in each step; because they may depend on $\vec{h}_i$, they are chosen after $\phi_0$, and $\phi_1,\phi_2$ are chosen after them, injective on the resulting inputs.

	For the fixed pair of point clouds $\cloud,\othercloud$, a \new{node feature}, denoted by $a$, is a pair of families $(a_i)_{i\in I}$ and $(a_i')_{i\in I'}$ of values assigned to the points of $\cloud$ indexed by $I$ and to the points of $\othercloud$ indexed by $I'$, where either $I=I'=[n]$, i.e., the node feature is defined on all points, or $I=\{i\in[n]\colon r_i>0\}$ and $I'=\{i\in[n]\colon r_i'>0\}$, i.e., the node feature is defined only on the points with positive radius. Let $b$ be a node feature with index sets $I$ and $I'$, and let $a$ be a node feature defined at least on the same points, i.e., whose index sets contain $I$ and $I'$, respectively. We say that $a_i$ \new{determines} $b_i$ if there exists some function $F$ such that $b_i=F(a_i)$ for all $i\in I$ and $b_i'=F(a_i')$, for all $i\in I'$. Note that $F$ may be chosen after $\cloud$ and $\othercloud$ are fixed, so it may depend on them. For example, since $\phi_0$ is injective, $\vec{h}_i$ determines $\vec{q}_i$ and $r_i$: there exists some $F$ that satisfies $(\vec{q}_i,r_i)=F(\vec{h}_i)$ and $(\vec{q}_i',r_i')=F(\vec{h}_i')$ for all $i$. The proof of separation proceeds in the three steps outlined after \cref{thm:main}.

	\par\medskip\noindent\emph{1. Equivariant linear independence is enough.}
	The first idea behind the proof is well known from frame-based architectures~\citep{luo2022orientations,du2022clofnet}. That is, if, for some choice of functions $\psi_1,\psi_2$ and some $i \in [n]$, the equivariant features $\vec{y}_{i,1}$ and $\vec{y}_{i,2}$ are linearly independent and $E(\cloud)=E(\othercloud)$ for all $E\in\funspace(d)$, then $\cloud\cong\othercloud$. To see this, recall that the functions $\phi_0,\phi_1,\phi_2$ are injective. Since $\vec{y}_{i,1},\vec{y}_{i,2}$ are linearly independent, the vectors
	\[
		\vec{f}_1 \coloneq \vec{y}_{i,1},\qquad \vec{f}_2 \coloneq \vec{y}_{i,2},\qquad \vec{f}_3 \coloneq \vec{y}_{i,1}\times\vec{y}_{i,2}
	\]
	form a positively oriented basis of $\R^3$, which we call the \new{frame} at point $i$. The feature $\vec{z}_i$ records this frame and the cloud relative to it. Indeed, since the function $\phi_1$ is injective, the vector $\vec{z}_i$ determines the Gram vector $\vec{g}_i$, which fixes all inner products among $\vec{f}_1,\vec{f}_2,\vec{f}_3$, and the multiset $\multiset{(\vec{h}_j,\vec{p}_{ij})\colon j\in[n]}$. Here $\vec{p}_{ij}$ lists the inner products of $\vec{x}_j$ with the frame, which determine $\vec{x}_j$, and $\vec{h}_j$ determines $\vec{q}_j$ since $\phi_0$ is injective.

	We now compare the two clouds. Since $\phi_2$ is injective, $E(\cloud)=E(\othercloud)$ gives $\multiset{(\vec{h}_j,\vec{z}_j)\colon j\in[n]}=\multiset{(\vec{h}_j',\vec{z}_j')\colon j\in[n]}$, so after reindexing $\othercloud$ we have $\vec{z}_i'=\vec{z}_i$. Hence $\vec{g}_i'=\vec{g}_i$, so $\vec{y}_{i,1}',\vec{y}_{i,2}'$ are also linearly independent, and their frame $\vec{f}_1',\vec{f}_2',\vec{f}_3'$ has the same Gram matrix and orientation as $\vec{f}_1,\vec{f}_2,\vec{f}_3$. Thus, some rotation $\vec{R}\in\SO(3)$ satisfies $\vec{R}\vec{f}_m=\vec{f}_m'$ for $m\in[3]$. Moreover, $\vec{z}_i'=\vec{z}_i$ gives a permutation $\sigma$ of $[n]$ with $\vec{h}_{\sigma(j)}'=\vec{h}_j$ and $\vec{p}_{i\sigma(j)}'=\vec{p}_{ij}$, for all $j$. Since $\vec{R}$ preserves inner products, the second equality says that $\vec{x}_{\sigma(j)}'$ and $\vec{R}\vec{x}_j$ have the same inner products with the basis $\vec{f}_1',\vec{f}_2',\vec{f}_3'$, so $\vec{x}_{\sigma(j)}'=\vec{R}\vec{x}_j$. The first equality gives $\vec{q}_{\sigma(j)}'=\vec{q}_j$. Hence the point cloud $\othercloud$ is obtained from $\cloud$ by the rotation $\vec{R}$ and a permutation, i.e., $\cloud\cong\othercloud$.

	\par\medskip\noindent\emph{2. Characterizing linear independence failure.}
	Our next step is to characterize when linear independence of $ \vec{y}_{i,1}$ and $\vec{y}_{i,2}$ cannot be attained. We will then deal with this case separately in step 3.

	To attain this characterization, we introduce some notation.
	Recall that $\vec{u}_i=\vec{x}_i/r_i$ denotes the direction of $\vec{x}_i$ when $r_i>0$. Let $S_\cloud$ be the set of distinct directions of the nonzero points, which we call the \new{support} of $\cloud$, and let $C_\cloud$ be the coloring of $S_\cloud$ that assigns to each direction $\vec{u}$ the multiset $C_\cloud(\vec{u})$ of radii and charges of the points on its ray:
	\begin{equation}
		S_{\cloud}=\{\vec{u}_i\colon i\in[n],\ r_i>0\},\qquad
		C_{\cloud}(\vec{u})=\multiset{(r_i,\vec{q}_i)\colon i\in[n],\ r_i>0,\ \vec{x}_i=r_i \vec{u}}.
		\label{eq:rayprofile}
	\end{equation}
	Here, $C_\cloud(\vec{u})$ ranges over all indices $i$, so point multiplicity is accounted for, i.e., each point on the ray contributes its own pair $(r_i,\vec{q}_i)$, also when several points share a position, whether their charges agree or differ.
	We will be interested in the notion of color-balanced sets.
	\begin{definition}[Color balance]\label{def:cb}
		A finite set $S\subset S^2$ with a coloring $C\colon S\to\mathcal C$ is
		\emph{color-balanced} if, for every $\vec{u}\in S$, $t\in[-1,1]$, and
		$c\in\mathcal C$,
		\begin{equation}
			\sum_{\substack{\vec{v}\in S\colon \vec{u}\cdot \vec{v}=t\\ C(\vec{v})=c}}\vec{v}
			\in\operatorname{span}\{\vec{u}\}.
			\tag{CB}\label{eq:CB}
		\end{equation}
	\end{definition}
	Equivalently, every union of color classes of $(S,C)$ is balanced in the sense of Definition~\ref{def:b} (adding \eqref{eq:CB} over the colors in the union gives one direction; conversely, at a point $\vec{u}\in S$ of color $c'$, the sum over color $c$ is the difference of the sums over the unions of the classes $c$ and $c'$ and of the class $c'$ alone); physically, if the points of $S$ are particles confined to the sphere, they are in equilibrium under every pairwise force law that depends on the distance and on the colors of the two particles, i.e., the net force on each particle is normal to the sphere. For a single color, this is the notion of a \emph{balanced} set of \citet{leech1957equilibrium}, whose Section~5 also treats the colored version.

	We first show that the feature $\vec{h}_i$ determines $\vec{q}_i$, $r_i$ and $C_\cloud(\vec{u}_i)$.
	Since the function $\phi_0$ is injective, $\vec{h}_i$ determines the inputs of $\phi_0$ in \eqref{eq:h}, namely $\vec{q}_i$, $r_i$ and the multiset $\multiset{(\vec{q}_j,r_j,\vec{x}_i\cdot\vec{x}_j)\colon j\in[n]}$; this gives $\vec{q}_i$ and $r_i$ directly. The color $C_\cloud(\vec{u}_i)$ takes one more step. For a feature $\vec{h}=\phi_0(\vec{q}_0,r_0,\mathcal{M})$ with $r_0>0$, let
	\[
		F(\vec{h})\coloneq\multiset{(r,\vec{q})\colon(\vec{q},r,s)\in\mathcal{M},\ r>0,\ s=r_0r};
	\]
	this is well defined since the function $\phi_0$ is injective. A point $j$ lies on the ray through $\vec{u}_i$ if and only if $r_j>0$ and $\vec{x}_i\cdot\vec{x}_j=r_ir_j$ (equality in Cauchy--Schwarz), so $F(\vec{h}_i)=C_\cloud(\vec{u}_i)$ whenever $r_i>0$, and likewise $F(\vec{h}_i')=C_{\othercloud}(\vec{u}_i')$ whenever $r_i'>0$.

	We claim that, if $(S_{\cloud},C_{\cloud})$ is \emph{not} color-balanced, then the functions can be chosen so that $\vec{y}_{i,1},\vec{y}_{i,2}$ are linearly independent for some $i$, and hence, by step~1, $\cloud$ and $\othercloud$ are congruent. To see this, assume that \eqref{eq:CB} does not hold for some $\vec{u}\in S_\cloud$, $t\in[-1,1]$, and color $c\in C_\cloud(S_\cloud)$. To attain two linearly independent vectors, choose some $i$ such that $\vec{x}_i$ is nonzero and $\vec{u}_i=\vec{u} $. Choose $\psi_1=0$ so that $\vec{y}_{i,1}=\vec{x}_i $. By assumption, this vector is nonzero. Now, define $\psi_2=\psi_2(\vec{h}_i,\vec{h}_j,\vec{x}_i\cdot \vec{x}_j)$ to be $r_j^{-1}$ if (a) $r_i>0$ and $r_j>0$, (b) $\vec{u}_i\cdot\vec{u}_j=t$, and (c) $C_\cloud(\vec{u}_j)=c$, and zero otherwise. All three conditions can be read from $(\vec{h}_i,\vec{h}_j,\vec{x}_i\cdot\vec{x}_j)$, since $\vec{u}_i\cdot\vec{u}_j=(r_ir_j)^{-1}\vec{x}_i\cdot\vec{x}_j$. As a result, we obtain that
	\begin{align*}
		\vec{y}_{i,2}-\vec{x}_i
		 & =\sum_{\substack{j\colon r_j>0,\ \vec{u}\cdot\vec{u}_j=t         \\ C_\cloud(\vec{u}_j)=c}} r_j^{-1}\vec{x}_j
		=\sum_{\substack{j\colon r_j>0,\ \vec{u}\cdot\vec{u}_j=t            \\ C_\cloud(\vec{u}_j)=c}} \vec{u}_j\\
		 & =\sum_{\substack{\vec{v}\in S_\cloud\colon \vec{u}\cdot\vec{v}=t \\ C_\cloud(\vec{v})=c}} \bigl|\{j\colon r_j>0,\ \vec{u}_j=\vec{v}\}\bigr|\,\vec{v}
		=|c|\sum_{\substack{\vec{v}\in S_\cloud\colon \vec{u}\cdot\vec{v}=t \\ C_\cloud(\vec{v})=c}}\vec{v}.
	\end{align*}
	The first equality follows from \eqref{eq:y} and the choice of the function $\psi_2$, using $\vec{u}_i=\vec{u}$; the second from $\vec{x}_j=r_j\vec{u}_j$; the third by grouping the indices $j$ by their direction $\vec{v}=\vec{u}_j$; and the last because the ray through each $\vec{v}$ of color $c$ contains exactly $|c|$ points, where $|c|$ is the size of the multiset $c$. Since \eqref{eq:CB} fails for $\vec{u}$, $t$ and $c$,
	\[
		\sum_{\substack{\vec{v}\in S_\cloud\colon \vec{u}\cdot\vec{v}=t\\ C_\cloud(\vec{v})=c}}\vec{v}\notin\operatorname{span}\{\vec{u}\}.
	\]
	Since $|c|>0$, it follows that $\vec{y}_{i,2}-\vec{x}_i\notin\operatorname{span}\{\vec{u}\}$, and since $\vec{x}_i=r_i\vec{u}\in\operatorname{span}\{\vec{u}\}$ by the choice of $i$, also $\vec{y}_{i,2}\notin\operatorname{span}\{\vec{u}\}$. Thus $\vec{y}_{i,2}$ and $\vec{y}_{i,1}=r_i\vec{u} $ are linearly independent. Exchanging the roles of $\cloud$ and $\othercloud$, the same conclusion holds if $(S_{\othercloud},C_{\othercloud})$ is not color-balanced.

	\par\medskip\noindent\emph{3. The color-balanced case.}
	It remains to show that, if both $(S_{\cloud},C_{\cloud})$ and $(S_{\othercloud},C_{\othercloud})$ are color-balanced and $E(\cloud)=E(\othercloud)$ for all $E\in\funspace(d)$, then $\cloud\cong\othercloud$. The main tool in this argument is the following lemma.
	\gdef\coloredbalancedsetsbody{%
		Let $S,S'\subset S^2$ be finite, possibly empty, sets with colorings
		$C\colon S\to\mathcal C$ and $C'\colon S'\to\mathcal C$ such that $(S,C)$
		and $(S',C')$ are color-balanced. Suppose that their ordered-pair angle
		histograms and their color histograms agree:
		\begin{align}
			\multiset{\vec{u}\cdot \vec{v}\colon (\vec{u},\vec{v})\in S\times S}
			 & =\multiset{\vec{u}'\cdot \vec{v}'\colon(\vec{u}',\vec{v}')\in S'\times S'},
			\stmtlabel{eq:angle-hist}                                                      \\
			\multiset{C(\vec{u})\colon\vec{u}\in S}
			 & =\multiset{C'(\vec{u}')\colon\vec{u}'\in S'}.
			\stmtlabel{eq:color-hist}
		\end{align}
		Then some rotation $\vec{R}\in\SO(3)$ maps $S$ onto $S'$ while preserving all colors,
		i.e., $\vec{R}S=S'$ and $C'(\vec{R}\vec{u})=C(\vec{u})$, for every $\vec{u}\in S$.}
	\begin{lemma}[Histograms determine color-balanced sets]\label{lem:colored}
		\coloredbalancedsetsbody
	\end{lemma}
	In the multisets in \eqref{eq:angle-hist}, each ordered pair in the set $S\times S$ or $S'\times S'$ is counted once. Similarly, in \eqref{eq:color-hist}, each element of the set $S$ or $S'$ is counted once.

	We prove the lemma in Appendix~\ref{sec:colored-proof}; here we explain how it completes the proof. First, we show that, regardless of color balance, if $E(\cloud)=E(\othercloud)$ for all $E\in\funspace(d)$, then the histogram equalities \eqref{eq:angle-hist}--\eqref{eq:color-hist} hold with $S=S_\cloud$, $C=C_\cloud$, $S'=S_{\othercloud}$ and $C'=C_{\othercloud}$, and so does the following equality of the multisets of charges of the points at the origin:
	\begin{equation}\label{eq:zero}
		\multiset{\vec{q}_i\colon r_i=0}=\multiset{\vec{q}_i'\colon r_i'=0}.
	\end{equation}
	Then, using that $(S_\cloud,C_\cloud)$ and $(S_{\othercloud},C_{\othercloud})$ are color-balanced, we apply the lemma to show that $\cloud\cong\othercloud$.

	We begin by showing that the three equalities \eqref{eq:angle-hist}--\eqref{eq:zero} hold. For any choice of $\psi_1,\psi_2$, the injectivity of $\phi_2$ and the assumption $E(\cloud)=E(\othercloud)$ imply
	\[
		\multiset{(\vec{h}_i,\vec{z}_i)\colon i\in[n]}=\multiset{(\vec{h}_i',\vec{z}_i')\colon i\in[n]}.
	\]
	This in turn implies that, for every function $F$ defined on these pairs,
	\begin{equation}\label{eq:F-all}
		\multiset{F(\vec{h}_i,\vec{z}_i)\colon i\in[n]}=\multiset{F(\vec{h}_i',\vec{z}_i')\colon i\in[n]}.
	\end{equation}
	Since $\vec{h}_i$ determines $r_i$, this equality also holds separately for the indices with zero and with nonzero radius:
	\begin{subequations}
		\begin{align}
			\multiset{F(\vec{h}_i,\vec{z}_i)\colon i\in[n],\ r_i=0} & =\multiset{F(\vec{h}_i',\vec{z}_i')\colon i\in[n],\ r_i'=0},\label{eq:split-zero}    \\
			\multiset{F(\vec{h}_i,\vec{z}_i)\colon i\in[n],\ r_i>0} & =\multiset{F(\vec{h}_i',\vec{z}_i')\colon i\in[n],\ r_i'>0}.\label{eq:split-nonzero}
		\end{align}
	\end{subequations}
	To see this, apply \eqref{eq:F-all} to the function $\tilde F$ that agrees with $F$ on the pairs with $r_i=0$ and maps all other pairs to a dummy value $\ast$ outside the range of $F$; removing all copies of $\ast$ from both sides then gives \eqref{eq:split-zero}. The same argument with the roles of the conditions $r_i=0$ and $r_i>0$ exchanged gives \eqref{eq:split-nonzero}.
	We obtain \eqref{eq:zero}, \eqref{eq:color-hist} and \eqref{eq:angle-hist}, in this order, by suitable choices of $F$.

	For \eqref{eq:zero}, since $\phi_0$ is injective, there is a function $F$ with $F(\vec{h}_i,\vec{z}_i)=\vec{q}_i$ and $F(\vec{h}_i',\vec{z}_i')=\vec{q}_i'$ for all $i$; for this $F$, \eqref{eq:split-zero} is exactly \eqref{eq:zero}.

	For \eqref{eq:color-hist}, let $F$ be the function from step~2, so that $F(\vec{h}_i)=C_\cloud(\vec{u}_i)$ whenever $r_i>0$ and $F(\vec{h}_i')=C_{\othercloud}(\vec{u}_i')$ whenever $r_i'>0$. Applying \eqref{eq:split-nonzero} to $(\vec{h},\vec{z})\mapsto F(\vec{h})$ gives
	\[\multiset{C_{\cloud}(\vec{u}_i)\colon i\in[n],\ r_i>0} =\multiset{C_{\othercloud}(\vec{u}_i')\colon i\in[n],\ r_i'>0}.\]
	For every $c\in C_\cloud(S_\cloud)$, the ray through each $\vec{u}\in S_\cloud$ of color $c=C_\cloud(\vec{u})=\multiset{(r_i,\vec{q}_i)\colon i\in[n],\ r_i>0,\ \vec{x}_i=r_i\vec{u}}$ contains exactly $|c|=|\{i\in[n]\colon r_i>0,\ \vec{x}_i=r_i\vec{u}\}|$ points of $\cloud$, so the multiplicity of $c$ on the left-hand side is $|c|$ times its multiplicity in $\multiset{C_\cloud(\vec{u})\colon\vec{u}\in S_\cloud}$. The same holds on the right-hand side for every $c\in C_{\othercloud}(S_{\othercloud})$, with $\othercloud$ in place of $\cloud$. Since the two sides are equal, they contain the same colors, $C_\cloud(S_\cloud)=C_{\othercloud}(S_{\othercloud})$, each with the same multiplicity. Dividing this multiplicity by $|c|$, which depends only on $c$, shows that each $c$ has the same multiplicity in $\multiset{C_\cloud(\vec{u})\colon\vec{u}\in S_\cloud}$ and in $\multiset{C_{\othercloud}(\vec{u}')\colon\vec{u}'\in S_{\othercloud}}$, and thus \eqref{eq:color-hist} holds.

	For \eqref{eq:angle-hist}, choose $\psi_1=0$, so that for all $i\in[n]$, $\vec{y}_{i,1}=\vec{x}_i$ and thus the first entry of $\vec{p}_{ij}$ is $\vec{y}_{i,1}\cdot\vec{x}_j=\vec{x}_i\cdot\vec{x}_j$; the choice of $\psi_2$ does not matter.

	Since the function $\phi_1$ is injective, $\vec{z}_i$ determines the multiset
	$\multiset{(\vec{h}_j,\vec{y}_{i,1}\cdot\vec{x}_j)\colon j\in[n]}$,
	which equals
	\begin{equation}\label{eq:zi-multiset}
		\multiset{(\vec{h}_j,\vec{x}_i\cdot\vec{x}_j)\colon j\in[n]}.
	\end{equation}
	Moreover, since $\phi_0$ is injective, $\vec{h}_i$ determines $r_i$, and by step~2, $\vec{h}_i$ also determines $C_\cloud(\vec{u}_i)$.

	If no $i$ has $r_i>0$, then by \eqref{eq:split-nonzero} with a constant $F$, no $i$ has $r_i'>0$ either, so $S_\cloud=S_{\othercloud}=\emptyset$, and thus \eqref{eq:angle-hist} holds trivially, with both of its sides being the empty multiset. Hence, we may assume that not all the $r_i$ are zero. Now let $i$ be any index in $[n]$ with $r_i>0$, and fix it.

	Since each $\vec{h}_j$ determines $r_j$ (by the above, with $j$ in place of $i$), keeping only the elements of \eqref{eq:zi-multiset} with $r_j>0$ shows that $\vec{z}_i$ also determines
	\[
		\multiset{(\vec{h}_j,\vec{x}_i\cdot\vec{x}_j)\colon j\in[n],\ r_j>0}.
	\]
	For each of these elements, $r_j>0$, so $C_\cloud(\vec{u}_j)$ is defined and determined by $\vec{h}_j$. Thus $\vec{z}_i$ determines
	\[
		\multiset{(r_j,\,C_\cloud(\vec{u}_j),\,\vec{x}_i\cdot\vec{x}_j)\colon j\in[n],\ r_j>0}.
	\]
	Finally, $\vec{h}_i$ determines $r_i$, and $\vec{u}_i\cdot\vec{u}_j=(r_ir_j)^{-1}\,\vec{x}_i\cdot\vec{x}_j$. Hence, replacing each element $(r_j,C_\cloud(\vec{u}_j),\vec{x}_i\cdot\vec{x}_j)$ above by $(C_\cloud(\vec{u}_j),\vec{u}_i\cdot\vec{u}_j)$, we see that the pair $(\vec{h}_i,\vec{z}_i)$ determines
	\begin{equation}\label{eq:color-angle-multiset}
		\multiset{(C_\cloud(\vec{u}_j),\,\vec{u}_i\cdot\vec{u}_j)\colon j\in[n],\ r_j>0}.
	\end{equation}
	For $\vec{u}\in S_\cloud$, let
	\[
		A_\cloud(\vec{u})\coloneq\multiset{\vec{u}\cdot\vec{v}\colon\vec{v}\in S_\cloud}
	\]
	be the \new{angle multiset} of $\vec{u}$, i.e., the multiset of inner products of $\vec{u}$ with the directions in $S_\cloud$. As in the derivation of \eqref{eq:color-hist}, each direction $\vec{v}\in S_\cloud$ contributes $|C_\cloud(\vec{v})|$ copies of $(C_\cloud(\vec{v}),\vec{u}_i\cdot\vec{v})$ to the multiset \eqref{eq:color-angle-multiset}, one for each point on its ray. Dividing the multiplicity of each pair $(c,t)$ by $|c|$ and then dropping $c$ therefore gives $A_\cloud(\vec{u}_i)$. Since $i$ was an arbitrary index with $r_i>0$, the pair $(\vec{h}_i,\vec{z}_i)$ thus determines $(C_\cloud(\vec{u}_i),A_\cloud(\vec{u}_i))$, that is, there is a function $F$ with $F(\vec{h}_i,\vec{z}_i)=(C_\cloud(\vec{u}_i),A_\cloud(\vec{u}_i))$ for all $i$ with $r_i>0$ and $F(\vec{h}_i',\vec{z}_i')=(C_{\othercloud}(\vec{u}_i'),A_{\othercloud}(\vec{u}_i'))$, for all $i$ with $r_i'>0$. Recall that \eqref{eq:F-all} holds for every function $F$ and every choice of functions $\psi_1,\psi_2$; as shown above, it follows that \eqref{eq:split-nonzero} holds for every $F$ as well, i.e.,
	\[
		\multiset{F(\vec{h}_i,\vec{z}_i)\colon i\in[n],\ r_i>0}=\multiset{F(\vec{h}_i',\vec{z}_i')\colon i\in[n],\ r_i'>0}.
	\]
	For the function $F$ just constructed, this equality reads
	\[
		\multiset{(C_\cloud(\vec{u}_i),A_\cloud(\vec{u}_i))\colon i\in[n],\ r_i>0}
		=\multiset{(C_{\othercloud}(\vec{u}_i'),A_{\othercloud}(\vec{u}_i'))\colon i\in[n],\ r_i'>0}.
	\]
	As in the derivation of \eqref{eq:color-hist}, dividing the multiplicity of each pair $(c,A)$ by $|c|$ in the above multisets and then dropping the entry $c$ gives
	\[
		\multiset{A_\cloud(\vec{u})\colon\vec{u}\in S_\cloud}=\multiset{A_{\othercloud}(\vec{u}')\colon\vec{u}'\in S_{\othercloud}}.
	\]

	Merging the multisets $A_\cloud(\vec{u})$ on the left-hand side (adding up multiplicities) gives the left-hand side of \eqref{eq:angle-hist}, and doing the same for the right-hand side above gives the right-hand side of \eqref{eq:angle-hist}. Since the two sides are equal, this proves \eqref{eq:angle-hist}.

	Finally, we apply the lemma with $S=S_\cloud$, $C=C_\cloud$, $S'=S_{\othercloud}$ and $C'=C_{\othercloud}$. Its hypotheses hold, i.e., $C_\cloud$ and $C_{\othercloud}$ both take values in the set $\mathcal C$ of finite multisets of pairs in $(0,\infty)\times\R^d$, $(S_\cloud,C_\cloud)$ and $(S_{\othercloud},C_{\othercloud})$ are color-balanced, and \eqref{eq:angle-hist}--\eqref{eq:color-hist} hold. Hence there is a rotation $\vec{R}\in\SO(3)$ such that $\vec{R}S_\cloud=S_{\othercloud}$ and $C_{\othercloud}(\vec{R}\vec{u})=C_\cloud(\vec{u})$, for every $\vec{u}\in S_\cloud$.

	It remains to show that $\vec{R}$, together with some permutation of the points, maps $\cloud$ to $\othercloud$. Let $\vec{u}\in S_\cloud$. By the definition \eqref{eq:rayprofile} of the colors, the equality $C_\cloud(\vec{u})=C_{\othercloud}(\vec{R}\vec{u})$ reads
	\[
		\multiset{(r_i,\vec{q}_i)\colon i\in[n],\ r_i>0,\ \vec{x}_i=r_i\vec{u}}=\multiset{(r_i',\vec{q}_i')\colon i\in[n],\ r_i'>0,\ \vec{x}_i'=r_i'\vec{R}\vec{u}}.
	\]
	Applying the map $(r,\vec{q})\mapsto(r\,\vec{R}\vec{u},\vec{q})$ to every element on both sides gives
	\[
		\multiset{(r_i\vec{R}\vec{u},\vec{q}_i)\colon i\in[n],\ r_i>0,\ \vec{x}_i=r_i\vec{u}}=\multiset{(r_i'\vec{R}\vec{u},\vec{q}_i')\colon i\in[n],\ r_i'>0,\ \vec{x}_i'=r_i'\vec{R}\vec{u}}.
	\]
	Since $r_i\vec{R}\vec{u}=\vec{R}(r_i\vec{u})=\vec{R}\vec{x}_i$ whenever $\vec{x}_i=r_i\vec{u}$, and $r_i'\vec{R}\vec{u}=\vec{x}_i'$ whenever $\vec{x}_i'=r_i'\vec{R}\vec{u}$, this equality reads
	\[
		\multiset{(\vec{R}\vec{x}_i,\vec{q}_i)\colon i\in[n],\ r_i>0,\ \vec{x}_i=r_i\vec{u}}=\multiset{(\vec{x}_i',\vec{q}_i')\colon i\in[n],\ r_i'>0,\ \vec{x}_i'=r_i'\vec{R}\vec{u}}.
	\]
	This equality holds for every $\vec{u}\in S_\cloud$, so the unions of both sides over all $\vec{u}\in S_\cloud$, taken with multiplicities added up, are equal as well. In the union of the left-hand sides, each $i$ with $r_i>0$ appears exactly once, namely for $\vec{u}=\vec{u}_i$. In the union of the right-hand sides, each $i$ with $r_i'>0$ also appears exactly once, namely for the $\vec{u}$ with $\vec{R}\vec{u}=\vec{u}_i'$, which exists and is unique because $\vec{R}$ maps $S_\cloud$ bijectively onto $S_{\othercloud}$. Hence, the following multisets are equal:
	\[
		\multiset{(\vec{R}\vec{x}_i,\vec{q}_i)\colon i\in[n],\ r_i>0}=\multiset{(\vec{x}_i',\vec{q}_i')\colon i\in[n],\ r_i'>0}.
	\]
	Together with \eqref{eq:zero} and $\vec{R}\vec{0}=\vec{0}$, which take care of the points at the origin, we obtain
	\[
		\multiset{(\vec{R}\vec{x}_i,\vec{q}_i)\colon i\in[n]}=\multiset{(\vec{x}_i',\vec{q}_i')\colon i\in[n]},
	\]
	that is, $\othercloud$ is obtained from $\cloud$ by the rotation $\vec{R}$ and a permutation of the points. Hence $\cloud\cong\othercloud$.
\end{proof}

\section{Proof of the lemma on color-balanced sets}
\label{sec:colored-proof}

The goal of this section is to prove Lemma~\ref{lem:colored}.

\begin{restatedlemmacolored}[Histograms determine color-balanced sets]
	\begingroup\renewcommand{\stmtlabel}[1]{\tag{\ref{#1}}}\coloredbalancedsetsbody\endgroup
\end{restatedlemmacolored}

Our proof of the lemma combines Leech's classification of balanced sets \citep[Section~4]{leech1957equilibrium} with a new congruence argument. Leech showed that every uncolored balanced set belongs, up to an orthogonal transformation, to one of the following classes: (a) sets of equally spaced points on a great circle, possibly together with its two axial poles; or (b) unions of the vertices, edge midpoints and face centers of a Platonic solid, projected to the sphere. This result is stated formally in Theorem~\ref{thm:leech} below. In Section~5 of the same paper, he then considered the colored case, framed as multiple sets of particles that remain in equilibrium for every choice of force laws, where the force between two particles may depend on their distance and on the sets they belong to; these sets correspond to our color classes. He also listed the possible partitions of a balanced set into such sets, that is, in our terms, its color-balanced colorings. What is new here is the congruence result and its proof: we show that these sets and their colorings are determined, up to rotation, by their angle and color histograms. For uncolored sets, we use his classification as stated (Theorem~\ref{thm:leech}). He obtains the colored case directly from the uncolored classification; we re-derive this colored part of his result in the form our proof needs.

The proof consists of two stages. First, in Appendix~\ref{sec:uncolored}, we show that the angle histogram determines a balanced set up to rotation (Proposition~\ref{prop:uncolored}). Second, in Appendix~\ref{sec:colored-general}, we extend this to colorings. By the first stage, the \emph{supports} $S$ and $S'$ of the two color-balanced sets in Lemma~\ref{lem:colored}, that is, the sets with their colors ignored, are related by a rotation. Moving the colors of $S'$ back to $S$ along this rotation, we obtain a second color-balanced coloring of $S$ with the same color histogram as $C'$, so it suffices to compare two colorings of the same set $S$. For each balanced set $S$, we determine the constraints that color balance imposes on a coloring of $S$, and use them to show that any two color-balanced colorings of the same set $S$ with equal color histograms are related by a rotation that maps $S$ onto itself. Since $S$ is balanced, it belongs, up to an orthogonal transformation, to one of Leech's classes (a) and (b), that is, to the first two families or to the third family of Theorem~\ref{thm:leech}. For class (a), we give a direct proof; for class (b), we give a computer-assisted proof in exact arithmetic (Appendix~\ref{app:exact}). Throughout, every claim about the finitely many Platonic balanced sets is either proved in the text or marked as verified in Appendix~\ref{app:exact}. An \emph{angle} between two unit vectors $\vec{u},\vec{v}$ means their inner product $t=\vec{u}\cdot\vec{v}$, and the \emph{angle histogram} of $S$ is the multiset \eqref{eq:angle-hist}. We first define balance:
\begin{definition}[Balance]\label{def:b}
	A finite set $S\subset S^2$ is \emph{balanced} if, for every $\vec{u}\in S$ and $t\in[-1,1]$,
	\begin{equation}
		\sum_{\substack{\vec{v}\in S\colon\ \vec{u}\cdot \vec{v}=t}}\vec{v}
		\in\operatorname{span}\{\vec{u}\}.
		\tag{B}\label{eq:B}
	\end{equation}
\end{definition}
Note that if $(S,C)$ is color-balanced, then $S$ is balanced, by summing \eqref{eq:CB} over all colors $c$. Moreover, $S$ is balanced if and only if it is color-balanced with respect to the constant coloring $C\equiv 0$.

\subsection{Uncolored balanced sets}\label{sec:uncolored}
Our proof for uncolored balanced sets relies on Leech's classification of balanced sets \cite{leech1957equilibrium} (see also \cite{cohn2010balanced}; the planar case is treated in \cite{pierson2022balanced}). To state it, we first define the Platonic balanced sets. These are constructed from the \new{Platonic solids}, which are the five convex regular polyhedra: the tetrahedron, the cube, the octahedron, the dodecahedron and the icosahedron. The faces of a Platonic solid are congruent regular polygons, and the same number of faces meet at every vertex. A \new{Platonic balanced set} is constructed from a Platonic solid in three steps:
first, center the solid at the origin and scale it so that its vertices lie on the unit sphere;
second, project its edge midpoints and its face centers radially onto the unit sphere;
third, take the union of one, two, or all three of the following sets: the vertices ($V$), the projected edge midpoints ($E$), and the projected face centers ($F$).
See \cref{fig:leech-classification}(b).
\begin{theorem}[\citet{leech1957equilibrium}]\label{thm:leech}
	Every nonempty finite balanced set $S\subset S^2$ is the image under some $\vec{Q}\in\OO(3)$ of a set in one of the following three families:
	\begin{enumerate}
		\item \textbf{regular polygons:} $m\geq1$ equally spaced points on the unit circle in the $XY$ plane (including the degenerate cases $m=1,2$);
		\item \textbf{regular polygons with poles:} a regular polygon with $m-2\geq3$ points in the $XY$ plane, together with the north and south poles;
		\item \textbf{Platonic balanced sets.}
	\end{enumerate}
\end{theorem}
In the second family, the cases $m-2\le 2$ are excluded: for $m-2=0$ and $m-2=2$, the set is congruent to the regular polygon with $m=2$ and $m=4$ points respectively, and for $m-2=1$ it is not balanced, since at either pole, the only point at inner product $0$ is the single point on the ring, which violates the balance condition.

The \new{dual} of a Platonic solid is the Platonic solid whose vertices are the projections of the face centers of the original one onto the sphere: the cube and the octahedron are dual to each other, as are the dodecahedron and the icosahedron, and the tetrahedron is dual to itself. A solid and its dual give rise to the same Platonic balanced sets, with the roles of vertices and face centers exchanged; their edge midpoints point in the same directions. It therefore suffices to consider the tetrahedron, the octahedron and the icosahedron; these duality statements are verified in Appendix~\ref{sec:solids-check}. This a priori gives 21 options, as for each of the three solids, we can consider 7 types of unions: $\{V,E,F,VE,EF,VF,VEF \}$. We write $T(\cdot)$, $O(\cdot)$ and $I(\cdot)$ for the sets obtained from the tetrahedron, the octahedron and the icosahedron, respectively; for example, $O(VF)$ is the union of the vertices and the projected face centers of the octahedron. However, some of these descriptions give congruent sets, that is, sets related by a rotation. The 21 descriptions fall into 16 congruence classes: five classes contain two descriptions each, namely $T(V)\cong T(F)$, $T(VE)\cong T(EF)$, $T(E)\cong O(V)$, $T(VF)\cong O(F)$ and $T(VEF)\cong O(VF)$, and the remaining eleven contain one each. Indeed, $T(V)\cong T(F)$ and $T(VE)\cong T(EF)$ via a quarter-turn about a coordinate axis, which maps the vertices of a tetrahedron onto their antipodes, which are its projected face centers, and maps its projected edge midpoints, which point along the coordinate axes, onto themselves; $T(E)=O(V)$, since the edge midpoints of a tetrahedron point along the three coordinate axes; and $T(VF)=O(F)$ and $T(VEF)=O(VF)$, since the vertices and projected face centers of a tetrahedron together form a cube, whose vertices are the projected face centers of the octahedron. These congruences are also verified in Appendix~\ref{sec:solids-check}. In total, there are 16 different possible Platonic balanced sets, and they are described in Table \ref{tab:hosts}. The explicit coordinates are given in
Appendix~\ref{app:exact}, and all 21 descriptions and their congruences are listed in Table~\ref{tab:21}.

Since orthogonal transformations preserve balance, Theorem~\ref{thm:leech} describes the balanced sets only up to $\OO(3)$, whereas Lemma~\ref{lem:colored} asks for a rotation in $\SO(3)$: congruence in this paper means a proper rotation, since the architecture distinguishes mirror images. Fortunately, for the sets listed in Theorem~\ref{thm:leech}, the gap closes: whenever such a $\vec{Q}$ exists, a rotation in $\SO(3)$ mapping one set onto the other exists as well. The reason is that each set $S$ above has an \new{improper symmetry}, that is, some $\vec{J}\in\OO(3)$ with $\det(\vec{J})=-1$ and $\vec{J}S=S$: for a polygon, with or without poles, a reflection in a plane containing the polygon; for a Platonic balanced set, a reflection symmetry of the solid, which maps its vertices, edge midpoints and face centers to themselves. Hence, if $\vec{Q}\in\OO(3)\setminus\SO(3)$, then $\vec{Q}\vec{J}\in\SO(3)$ also maps $S$ onto $\vec{Q}S$. This fails for general point sets: a chiral set and its mirror image are not related by any rotation.

We first show, in Proposition~\ref{prop:uncolored}, that the sets listed in Theorem~\ref{thm:leech} are determined, up to rotation, by their angle histograms: first the family, using the second-moment matrix, and then the set within its family. Together with Theorem~\ref{thm:leech}, this gives the uncolored case of Lemma~\ref{lem:colored} (Corollary~\ref{cor:uncolored}).
\begin{proposition}[The angle histogram determines a listed set]
	\label{prop:uncolored}
	Let $S,S'\subset S^2$ each be the image, under some rotation in $\SO(3)$, of one of the sets listed in Theorem~\ref{thm:leech}. If their angle histograms agree,
	\[
		\multiset{\vec{u}\cdot \vec{v}\colon(\vec{u},\vec{v})\in S\times S}
		=\multiset{\vec{u}'\cdot \vec{v}'\colon(\vec{u}',\vec{v}')\in S'\times S'},
	\]
	then $\vec{R}S=S'$ for some $\vec{R}\in\SO(3)$.
\end{proposition}
\begin{proof}
	Let $m=|S|$. Since the angle histogram has $m^2$ entries, also $|S'|=m$. The cases $m\le2$ are immediate: for $m=2$, the only listed set is the regular polygon with two points, that is, a pair of antipodal points, and any two pairs of antipodal points are related by a rotation. Now suppose $m\geq3$.

	Define the $3\times3$ second-moment matrix
	\[
		\vec{M}(S)\coloneq\sum_{\vec{u}\in S}\vec{u}\vec{u}^{\top}.
	\]
	The angle histogram determines its squared Frobenius norm: by bilinearity of the Frobenius inner product $\langle\cdot,\cdot\rangle_F$ and the identity $\langle\vec{u}\vec{u}^{\top},\vec{v}\vec{v}^{\top}\rangle_F=(\vec{u}\cdot\vec{v})^2$,
	\[
		\|\vec{M}(S)\|_F^2=\Bigl\langle\sum_{\vec{u}\in S}\vec{u}\vec{u}^{\top},\sum_{\vec{v}\in S}\vec{v}\vec{v}^{\top}\Bigr\rangle_F
		=\sum_{(\vec{u},\vec{v})\in S\times S}\langle\vec{u}\vec{u}^{\top},\vec{v}\vec{v}^{\top}\rangle_F
		=\sum_{(\vec{u},\vec{v})\in S\times S}(\vec{u}\cdot\vec{v})^2,
	\]
	which is the sum of the squares of the entries of the angle histogram.

	The eigenvalues of $\vec{M}(S)$ in each family are as follows.
	\begin{itemize}
		\item \textbf{Regular polygons} ($m\geq3$): $(m/2,m/2,0)$. Indeed, write the vertices as $(\cos\theta_k,\sin\theta_k,0)$ with $\theta_k=2\pi k/m$, $k\in[m]$. The third row and column of $\vec{M}(S)$ vanish, and its upper-left $2\times2$ block is
		      \[
			      \sum_{k\in[m]}\begin{pmatrix}\cos^2\theta_k&\cos\theta_k\sin\theta_k\\ \cos\theta_k\sin\theta_k&\sin^2\theta_k\end{pmatrix}
			      =\frac{m}{2}\vec{I}_2+\frac12\sum_{k\in[m]}\begin{pmatrix}\cos2\theta_k&\sin2\theta_k\\ \sin2\theta_k&-\cos2\theta_k\end{pmatrix}
			      =\frac{m}{2}\vec{I}_2,
		      \]
		      since $\sum_{k\in[m]}e^{2i\theta_k}=0$ for $m\geq3$.
		\item \textbf{Regular polygons with poles:} $((m-2)/2,(m-2)/2,2)$, since the poles $\pm\vec{e}_3$ add $2\vec{e}_3\vec{e}_3^{\top}$ to the second-moment matrix of the regular $(m-2)$-gon.
		\item \textbf{Platonic balanced sets:} $(m/3,m/3,m/3)$. Indeed, every symmetry $\vec{Q}\in\OO(3)$ of $S$ satisfies $\vec{Q}\vec{M}(S)\vec{Q}^{\top}=\vec{M}(\vec{Q}S)=\vec{M}(S)$, so each eigenspace of the symmetric matrix $\vec{M}(S)$ is invariant under the symmetry group of $S$. Since this group acts irreducibly on $\R^3$ (it contains rotations of order at least $3$ about two non-parallel axes, so it leaves no line invariant, and hence, taking orthogonal complements, no plane), $\vec{M}(S)$ has a single eigenspace, that is, $\vec{M}(S)=\lambda\vec{I}_3$ for some $\lambda\in\R$. Taking traces gives $3\lambda=\sum_{\vec{u}\in S}\|\vec{u}\|^2=m$.
	\end{itemize}
	Since $\vec{M}(S)$ is symmetric, $\|\vec{M}(S)\|_F^2$ is the sum of the squares of its eigenvalues. Hence, for a given $m$, $\|\vec{M}(S)\|_F^2$ takes the following values:
	\[
		\begin{array}{ll}
			\text{regular polygons:}            & m^2/2,       \\
			\text{regular polygons with poles:} & (m-2)^2/2+4, \\
			\text{Platonic balanced sets:}      & m^2/3.
		\end{array}
	\]
	The angle histogram determines $m$, as it has $m^2$ entries, and it determines $\|\vec{M}(S)\|_F^2$. We compare the three values pairwise for a given $m$:
	\begin{itemize}
		\item $m^2/2\neq m^2/3$ for every $m\geq1$.
		\item $m^2/2=(m-2)^2/2+4$ if and only if $m=3$, which is excluded, since regular polygons with poles have $m\geq5$.
		\item $(m-2)^2/2+4=m^2/3$ if and only if $(m-6)^2=0$, that is, $m=6$. In this case, the regular polygon with poles is the square with two poles, which is precisely the octahedron, a Platonic balanced set. So both families describe the same set.
	\end{itemize}
	Hence the angle histogram determines the family of $S$, except at $m=6$, where the two families that share a value describe the same set.

	It remains to show that within each family, the cardinality $m$ and the angle histogram determine $S$ up to rotation. In the first two families, $m$ alone suffices: for each $m$ there is exactly one regular polygon with $m$ points and, for $m\geq5$, exactly one regular polygon with poles with $m$ points, up to rotation. In the Platonic family, Table~\ref{tab:hosts} lists the 16 Platonic balanced sets up to rotation. Their sizes $m$ are pairwise distinct, except for two pairs: $O(E)$ and $I(V)$, both with $m=12$, and $O(EF)$ and $I(F)$, both with $m=20$. The sets in each pair are distinguished by their minimum squared distances, which can be inferred from the angle histogram:
	\[
		\begin{array}{c|c}
			\text{Support} & \min_{\vec{u}\ne \vec{v}}\|\vec{u}-\vec{v}\|^2 \\\hline
			O(E)           & 1                                              \\
			I(V)           & 2-2/\sqrt5\approx1.1056                        \\
			O(EF)          & 2-2\sqrt{2/3}\approx0.3670                     \\
			I(F)           & 2-2\sqrt5/3\approx0.5093.
		\end{array}
	\]
	These values are verified exactly in
	Appendix~\ref{app:exact}; equivalently, they are two minus twice the
	largest off-diagonal inner product. Thus the data determine the support
	up to a (special) orthogonal transformation.
\end{proof}
\begin{corollary}[Lemma~\ref{lem:colored} for uncolored sets]
	\label{cor:uncolored}
	Let $S,S'\subset S^2$ be finite, possibly empty, balanced sets whose angle histograms agree. Then $\vec{R}S=S'$ for some $\vec{R}\in\SO(3)$.
\end{corollary}
\begin{proof}
	If $S=\emptyset$, then the angle histograms agree only if $S'=\emptyset$, and the claim is trivial. Otherwise, by Theorem~\ref{thm:leech} and the improper symmetries discussed above, each of $S$ and $S'$ is the image under a rotation in $\SO(3)$ of one of the sets listed in Theorem~\ref{thm:leech}, and Proposition~\ref{prop:uncolored} applies.
\end{proof}

\begin{table}[tbp]
	\centering
	\small
	\begin{tabular}{r l l}
		\toprule
		$m$           & Support  & Sizes of its color atoms \\
		\midrule
		4             & $T(V)$   & $4$                      \\
		6             & $O(V)$   & $2+2+2$                  \\
		8             & $O(F)$   & $4+4$                    \\
		10            & $T(VE)$  & $4+6$                    \\
		$\mathbf{12}$ & $O(E)$   & $12$                     \\
		$\mathbf{12}$ & $I(V)$   & $12$                     \\
		14            & $O(VF)$  & $4+4+6$                  \\
		18            & $O(VE)$  & $12+6$                   \\
		$\mathbf{20}$ & $O(EF)$  & $8+12$                   \\
		$\mathbf{20}$ & $I(F)$   & $20$                     \\
		26            & $O(VEF)$ & $8+12+6$                 \\
		30            & $I(E)$   & $30$                     \\
		32            & $I(VF)$  & $12+20$                  \\
		42            & $I(VE)$  & $12+30$                  \\
		50            & $I(EF)$  & $30+20$                  \\
		62            & $I(VEF)$ & $12+30+20$               \\
		\bottomrule
	\end{tabular}
	\caption{\emph{The 16 distinct Platonic balanced sets.} Color atoms are defined in Appendix~\ref{sec:colored-general}; any assignment of colors to whole color atoms is color-balanced. The color atoms of $O(V)$ are
		antipodal pairs; those of $O(F)$ are its two inscribed tetrahedra.
		Bold: the two values of $m$ shared by two sets, which the proof of Proposition~\ref{prop:uncolored} distinguishes by minimum distances.}\label{tab:hosts}
\end{table}

\subsection{Proof for color-balanced sets}\label{sec:colored-general}
We now extend Corollary~\ref{cor:uncolored} to general colorings, which proves Lemma~\ref{lem:colored}. Since color-balanced sets are balanced, Corollary~\ref{cor:uncolored} applies to their supports. Thus, if $(S,C)$ and $(S',C')$ are color-balanced sets satisfying the hypotheses of Lemma~\ref{lem:colored}, then $S$ and $S'$ are related by a rotation in $\SO(3)$. It remains to show that this rotation can be chosen to preserve the colors.

By Corollary~\ref{cor:uncolored} there is $\vec{R}_0\in\SO(3)$ with $\vec{R}_0S=S'$. The coloring $C'\circ\vec{R}_0$ of $S$ is again color-balanced and has the same color histogram as $C'$, so we may assume $S'=S$. It then suffices to find $\vec{R}\in\SO(3)$ with $\vec{R}S=S$ and $C'(\vec{R}\vec{u})=C(\vec{u})$ for all $\vec{u}\in S$.

Call $\vec{u},\vec{v}\in S$ \new{inseparable} if $C(\vec{u})=C(\vec{v})$ for every color-balanced coloring $C$ of $S$. This is an equivalence relation, and its equivalence classes are the \new{color atoms} of $S$. Thus every color-balanced coloring of $S$ is constant on each color atom, and the color atoms are the blocks of the coarsest partition of $S$ with this property. We show below that, conversely, every coloring of $S$ that is constant on each color atom is color-balanced. This gives a second, equivalent description: the color atoms of $S$ are the color classes of the finest color-balanced coloring of $S$, the one that gives each color atom its own color, and every color-balanced coloring of $S$ is obtained from it by merging colors. For example, in a regular hexagon the two neighbors of a vertex $\vec{u}$ are the only points $\vec{v}$ with $\vec{u}\cdot\vec{v}=1/2$, and neither of them alone lies in $\operatorname{span}\{\vec{u}\}$, so by \eqref{eq:CB} they share a color in every color-balanced coloring; the color atoms are the two inscribed equilateral triangles. The name \emph{color atoms} refers to the indivisibility of the classes and is unrelated to the atoms of a molecule: a color atom is never split by a color-balanced coloring, and every color class of a color-balanced coloring is a union of whole color atoms, much as the atoms of a Boolean algebra are its minimal nonzero elements.

Let $B_1,\dots,B_k$ be the color atoms of $S$. Suppose that:
(i) for all $I,J\subseteq[k]$ with $\sum_{i\in I}|B_i|=\sum_{j\in J}|B_j|$, the multisets $\multiset{|B_i|\colon i\in I}$ and $\multiset{|B_j|\colon j\in J}$ are equal; that is, the number of points in a union of color atoms determines how many color atoms of each size it contains; and
(ii) for every permutation $\pi$ of $[k]$ with $|B_{\pi(i)}|=|B_i|$ for all $i\in[k]$, there exists $\vec{R}\in\SO(3)$ with $\vec{R}S=S$ and $\vec{R}B_i=B_{\pi(i)}$ for all $i\in[k]$.
For example, in $O(VF)$ the color atoms have sizes $4,4,6$ (Table~\ref{tab:hosts}), so a union of color atoms with $8$ points consists of the two tetrahedra, and one with $10$ points consists of a tetrahedron and the octahedron; property (i) would fail for sizes $2,2,4$, since $4$ points could form either $2+2$ or $4$.
Then the claim follows. Indeed, for each color $c$, write $C^{-1}(c)=\bigcup_{i\in I_c}B_i$ and $C'^{-1}(c)=\bigcup_{j\in J_c}B_j$. These two sets have the same cardinality by \eqref{eq:color-hist}, so by (i) the multisets $\multiset{|B_i|\colon i\in I_c}$ and $\multiset{|B_j|\colon j\in J_c}$ are equal. Since the sets $I_c$ partition $[k]$, and so do the sets $J_c$, there is a permutation $\pi$ of $[k]$ with $|B_{\pi(i)}|=|B_i|$ for all $i\in[k]$ and $\pi(I_c)=J_c$ for every color $c$, and the rotation $\vec{R}$ given by (ii) satisfies $C'(\vec{R}\vec{u})=C(\vec{u})$ for all $\vec{u}\in S$. This argument uses only that every color-balanced coloring is constant on each color atom, so it applies verbatim to any partition of $S$ into sets with this property. It therefore remains to exhibit, for each balanced set, such a partition satisfying (i) and (ii).

The color atoms are found with the following lemma. For a root $\vec{u}\in S$ and $t$ with $|t|<1$, the \new{shell} of $\vec{u}$ at $t$ is the set of $\vec{v}\in S$ with $\vec{u}\cdot\vec{v}=t$, and the \new{tangent vector} of such a $\vec{v}$ is $\vec{w}_{\vec{v}}=\vec{v}-t\vec{u}\in\vec{u}^\perp$. The tangent vectors of a shell are nonzero, pairwise distinct, and all of length $\sqrt{1-t^2}$. A sum of points of the shell lies in $\operatorname{span}\{\vec{u}\}$ exactly when the sum of their tangent vectors is zero. So \eqref{eq:B} says that the tangent vectors of each shell sum to zero, and \eqref{eq:CB} says that, within each shell, the tangent vectors of each color sum to zero.

\begin{lemma}[Small-shell constraints]\label{lem:shellrules}
	Let $(S,C)$ be color-balanced, $\vec{u}\in S$ and $|t|<1$, and consider the shell of $\vec{u}$ at $t$.
	\begin{enumerate}
		\item If the shell has two or three points, they all have the same color.
		\item If the shell has four points and $\vec{w}_{\vec{v}}=-\vec{w}_{\vec{v}'}$ for two of them, then $C(\vec{v})=C(\vec{v}')$.
		\item If the tangent vectors of the shell form a regular pentagon, all its points have the same color.
	\end{enumerate}
\end{lemma}
\begin{proof}
	By \eqref{eq:CB}, within the shell, the tangent vectors of each color sum to zero; since $(S,C)$ is color-balanced, $S$ is balanced, so the tangent vectors of the whole shell sum to zero as well. A singleton nonzero tangent vector cannot sum to zero. A proper
	nonempty zero-sum subset of a two- or three-point zero-sum shell would
	have a singleton complement or would itself be a singleton, proving
	the first assertion. In a four-point shell, a proper nonempty zero-sum
	subset can only be a pair, and such a pair must consist of opposite
	vectors. Now let $\vec{w}_{\vec{v}}=-\vec{w}_{\vec{v}'}$. If $C(\vec{v})\neq C(\vec{v}')$, the color class of $\vec{v}$ in the shell would be such a pair $\{\vec{v},\vec{x}\}$ with $\vec{w}_{\vec{x}}=-\vec{w}_{\vec{v}}=\vec{w}_{\vec{v}'}$, hence $\vec{x}=\vec{v}'$, since the tangent vectors are pairwise distinct; a contradiction.
	In a regular pentagon there is no zero-sum singleton or pair, since
	there are no opposite vertices. Taking complements rules out zero-sum
	subsets of size three or four, proving the last assertion.
\end{proof}

\paragraph{Polygon families.}
Using this lemma, we first prove the claim for polygon families (with or without poles). As in Proposition~\ref{prop:uncolored}, $m$ denotes the cardinality of the balanced set; for a polygon with poles, the equatorial ring itself therefore consists of $m-2$ points.

If $m\le1$, there is nothing to prove. If $m=2$, then $S=\{\vec{u},-\vec{u}\}$ has only shells with $t=\pm1$, so every coloring is color-balanced, and the color atoms are the two singletons. Then (i) holds trivially, and (ii) holds because a half-turn about an axis orthogonal to $\vec{u}$ exchanges the two points.

Consider a regular $m$-gon in a great circle, with cyclically indexed
vertices $\vec{u}_j$, $j\in\mathbb Z/m\mathbb Z$, and $m\geq3$. The two
ring neighbors $\vec{u}_{j-1},\vec{u}_{j+1}$ form a shell with two points. Thus their
colors agree, giving
\begin{equation}
	C(\vec{u}_{j-1})=C(\vec{u}_{j+1})\quad\text{for all }j.
	\label{eq:periodtwo}
\end{equation}
Hence every color-balanced coloring of an odd polygon is constant, and that of an even polygon is constant on each of the two alternating classes. Conversely, every coloring that is constant on these classes is color-balanced: at a root $\vec{u}_j$, every shell with $|t|<1$ consists of pairs $\{\vec{u}_{j-k},\vec{u}_{j+k}\}$, which lie in the same class and have opposite tangent vectors (the shells with $t=\pm1$ impose no condition). So these classes are the color atoms. For them, (i) holds, and (ii) holds because the rotation by $2\pi/m$ about the axis of the polygon exchanges the two classes.

Now consider the ring together with the north and south poles, so that the ring has $m-2\geq3$ points and the total cardinality is $m$. Except when $m-2=4$ (i.e.\ $m=6$), the nearest
ring-neighbor shell has inner product $\cos(2\pi/(m-2))\ne0$, so the poles
do not enter it and \eqref{eq:periodtwo}, applied to the ring of $m-2$ points, still holds. The exceptional $m-2=4$ case is the octahedron; we will deal with it later. At any ring root,
the poles belong to its zero-inner-product shell. If $4\nmid (m-2)$, that
shell consists only of the poles. If $4\mid (m-2)$, it also contains the
two ring vertices at angular offsets $\pm\pi/2$; the tangent vectors
form two opposite pairs. In both cases Lemma~\ref{lem:shellrules}
forces the two poles to have the same color.

Hence every color-balanced coloring is constant on the pair of poles and on the ring if $m-2$ is odd, or on each of its two alternating classes if $m-2$ is even. Conversely, every coloring that is constant on these sets is color-balanced. At a ring root $\vec{u}_j$, every shell with $|t|<1$ consists of pairs $\{\vec{u}_{j-k},\vec{u}_{j+k}\}$ of ring points in the same set and, at $t=0$, of the pair of poles; the tangent vectors of each such pair are opposite. At a pole, the only shell with $|t|<1$ is the ring, at $t=0$, and each of the sets above that it contains is a regular polygon with at least two points, centered on the polar axis, so its tangent vectors sum to zero. So these sets are the color atoms. The color-atom sizes are $m-2,2$ or $(m-2)/2,(m-2)/2,2$; since $m-2\neq2$, and $(m-2)/2\neq2$ because the octahedral case $m-2=4$ was set aside above, (i) holds. Property (ii) holds because the rotation by $2\pi/(m-2)$ about the polar axis exchanges the two ring classes and fixes the poles.

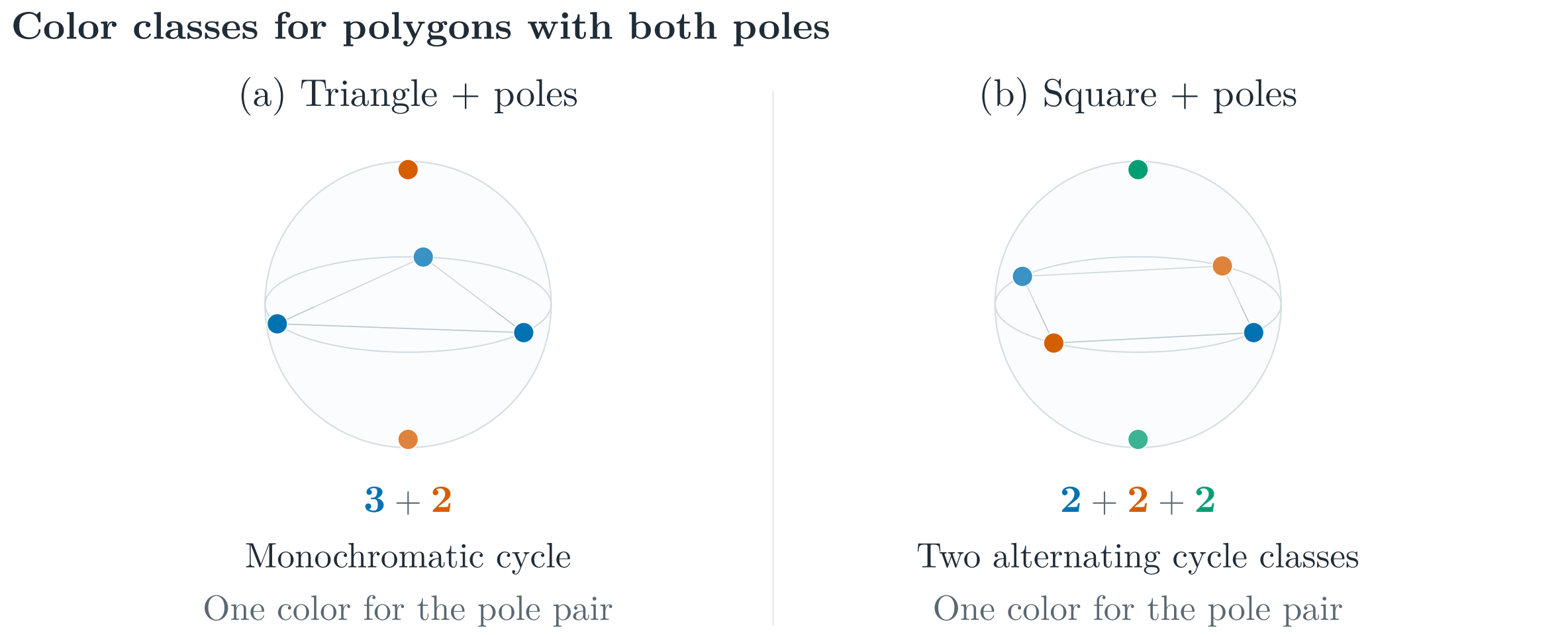
\begin{figure}[!htbp]
	\centering
	\input{figures/leech-polygon-poles}
	\caption{\emph{Admissible color classes for a color-balanced regular polygon with both poles.}
		(a) For a triangle, the three cycle points have one color and the two
		poles have one color, giving color-atom sizes $3+2$.
		(b) For a square, opposite cycle points share a color, giving two
		alternating cycle classes and a pole-pair class of sizes $2+2+2$.
		This support is the octahedron. Colors distinguish the finest classes;
		any classes may receive the same color. Thus the diagrams also
		describe all admissible colorings obtained by merging classes.}
	\label{fig:leech-polygon-poles}
\end{figure}

\Cref{fig:leech-polygon-poles} illustrates the color classes for rings
with three or four points and both poles. The four-point case is the
octahedral exception discussed below.

\FloatBarrier

\paragraph{The finite Platonic verification.}

For each of the 16 Platonic balanced sets, we apply the implications of
Lemma~\ref{lem:shellrules} at every root and every shell. Join two
points whenever one implication forces their colors to agree, and
take the connected components of this equality graph. The exact
calculation described in Appendix~\ref{app:exact} proves the following
finite assertion.

\begin{proposition}[Exact finite classification]\label{prop:finite}
	For every Platonic balanced set, the equality-graph components are
	the color atoms listed in Table~\ref{tab:hosts}. Every color atom is itself
	balanced at every root of the full support: for every color atom $B$, root
	$\vec{u}$, and angle $t$,
	\begin{equation}
		\sum_{\vec{v}\in B\colon\,\vec{u}\cdot \vec{v}=t}\vec{v} \in \operatorname{span}\{\vec{u}\}.
		\label{eq:atombalance}
	\end{equation}
	Consequently, a coloring satisfies \eqref{eq:CB} if and only if it is
	constant on each listed color atom.
\end{proposition}

\begin{proof}
	Necessity follows from the equality graph and
	Lemma~\ref{lem:shellrules}. Equation~\eqref{eq:atombalance} proves
	sufficiency for every coloring of the color atoms, since a color class is a
	union of whole color atoms. Appendix~\ref{app:exact} specifies the exact field,
	coordinates, every type of inference, and the output certificate.
\end{proof}
Only the first assertion, that every color-balanced coloring is constant on each component, is used in the proof of Lemma~\ref{lem:colored} (see the argument after properties (i) and (ii)); sufficiency shows that the components are exactly the color atoms.

The color atoms in Table~\ref{tab:hosts} satisfy (i), as can be checked row by row. Property (ii) is nontrivial only where color atoms have equal sizes, which happens in three cases: the octahedron vertices $O(V)$, with $2+2+2$, the octahedron faces $O(F)$ (the cube), with $4+4 $, and the union of vertices and faces $O(VF)$, with $4+4+6$.  Proper rotations realize
all permutations of the three unoriented octahedral axes and interchange
the two sets of cube vertices, while preserving the octahedron when it is
present. For example, a quarter-turn about a coordinate axis interchanges the two sets of cube vertices; see \cref{fig:leech-color-atoms}. The required rotations are verified explicitly in Appendix~\ref{app:exact}. This concludes the proof of Lemma~\ref{lem:colored}.\qedsymbol

\paragraph{Remark (mirror images).} Lemma~\ref{lem:colored} provides a rotation, and its proof only ever uses rotations, since Corollary~\ref{cor:uncolored} already relates the supports by a rotation. Still, as for uncolored sets (see the paragraph before Proposition~\ref{prop:uncolored}), reflections give nothing new for color-balanced sets: if $(S,C)$ is color-balanced and $\vec{Q}\in\OO(3)$ has $\det(\vec{Q})=-1$, then its mirror image $(\vec{Q}S,\,C\circ\vec{Q}^{-1})$ is also the image of $(S,C)$ under a rotation. The reason is that every balanced set $S$ has an improper symmetry $\vec{J}$, that is, $\vec{J}\in\OO(3)$ with $\det(\vec{J})=-1$ and $\vec{J}S=S$, that maps each color atom of $S$ onto itself: for a polygon, with or without poles, a reflection in a plane containing the polygon, which fixes every point of the polygon and exchanges the two poles, which form one color atom; for $O(V)$, whose color atoms are its three antipodal pairs, the sign change of one coordinate; for $O(F)$ and $O(VF)$, whose color atoms include the two inscribed tetrahedra $\{(\epsilon_1,\epsilon_2,\epsilon_3)/\sqrt3\colon\epsilon_1\epsilon_2\epsilon_3=\pm1\}$, the exchange of two coordinates; and for every other Platonic balanced set, any reflection symmetry of the solid, since there each color atom is the set of vertices, of projected edge midpoints or of projected face centers (Table~\ref{tab:hosts}). Every color-balanced coloring $C$ of $S$ is constant on each color atom, so $C\circ\vec{J}=C$, and the rotation $\vec{R}=\vec{Q}\vec{J}\in\SO(3)$ satisfies $\vec{R}S=\vec{Q}S$ and $C\circ\vec{R}^{-1}=C\circ\vec{J}^{-1}\circ\vec{Q}^{-1}=C\circ\vec{Q}^{-1}$. For each of the 21 descriptions of the Platonic balanced sets, the existence of such an improper symmetry is also verified in Appendix~\ref{app:exact}.

\begin{figure}[!htbp]
	\centering
	\input{figures/leech-color-atoms}
	\caption{\emph{Color atoms in the three Platonic balanced sets with equal-sized
			color atoms.} Colors here distinguish color atoms. (a) The octahedron $O(V)$ splits
		into three antipodal pairs of sizes $2+2+2$; proper rotations realize
		every permutation of the pairs. (b) The cube $O(F)$ splits into two
		inscribed tetrahedra of sizes $4+4$. (c) In $O(VF)$ the two tetrahedra
		and the octahedral vertices give color-atom sizes $4+4+6$. A quarter-turn
		about a coordinate axis interchanges the tetrahedra in (b) and (c)
		and preserves the octahedron in (c), as verified in Appendix~\ref{app:exact}. Any assignment of colors to
		whole color atoms is color-balanced, including assignments that merge
		color atoms into one color class.}
	\label{fig:leech-color-atoms}
\end{figure}
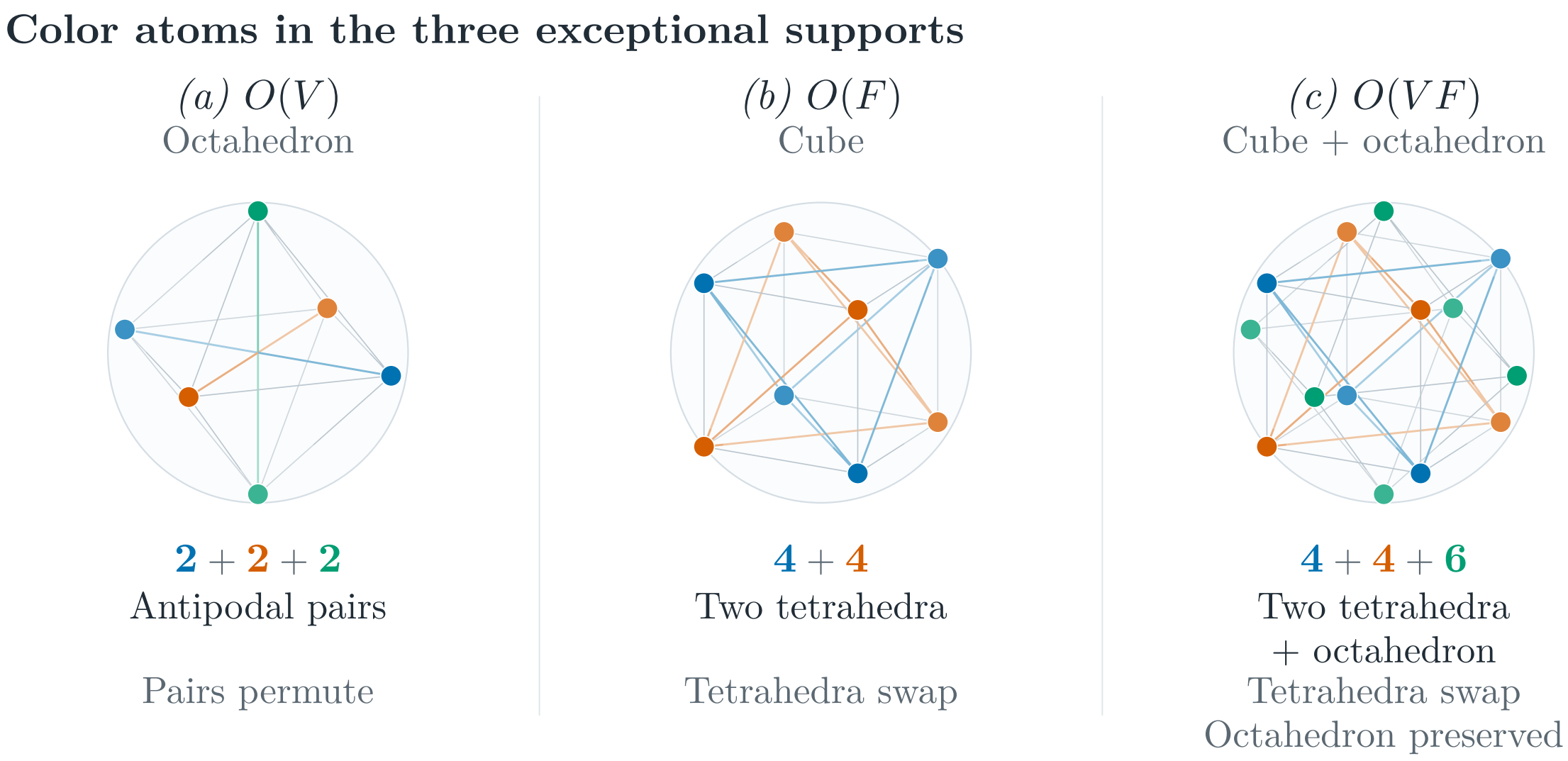

\FloatBarrier
\section{Exact arithmetic verification of the Platonic color atoms}
\label{app:exact}

The verification involves only the supports and which of their points must share a color; colors enter only through this, so the verification is independent of the set of colors and applies to charged clouds as well. The program \texttt{verify\_\allowbreak colored\_\allowbreak balance.py}, the certificate \texttt{colored\_\allowbreak balance\_\allowbreak certificate.json} it produces, and a companion notebook \texttt{balanced\_\allowbreak sets.ipynb}, which runs the same checks step by step with explanations, are provided as ancillary files with the arXiv version of this paper.
The program needs Python 3.9 or later and uses only the standard library. It constructs the
Platonic coordinates, forms their exact shells, records the necessary
color-equality implications, verifies the resulting color-atom partitions,
and checks the required color-atom symmetries and subset sums. It also verifies that the coordinates are those of the Platonic solids, the duality between the solids, the congruences listed in Table~\ref{tab:21}, and the minimum distances that separate the 16 sets of Table~\ref{tab:hosts}. From the
directory containing the script, run
\begin{quote}\ttfamily\small
	python3 verify\_colored\_balance.py\ \textbackslash\\
	\hspace*{1em}-{}-output colored\_balance\_certificate.json
\end{quote}
The certificate written by the program records the finite data behind Proposition~\ref{prop:finite}; it is meant to be inspected alongside the program, not as a substitute for running it.

\subsection{Number field and equality}

Set
\[
	\varphi=\frac{1+\sqrt5}{2},\qquad
	a=\sqrt{\varphi+2},\qquad b=\sqrt3,\qquad c=\sqrt2,
	\qquad K=\Q(\varphi,a,b,c).
\]
The positive real embedding is understood. Computations use the basis
\begin{equation}
	\varphi^e a^f b^g c^h,\qquad e,f,g,h\in\{0,1\},
	\label{eq:fieldbasis}
\end{equation}
with rational coefficients, reducing products by
\[
	\varphi^2=\varphi+1,\qquad a^2=\varphi+2,
	\qquad b^2=3,\qquad c^2=2.
\]
These 16 monomials span $K$ over $\Q$. They are also linearly independent: note that
$a=2\cos(\pi/10)$ and $\Q(a)=\Q(\zeta_{20})^+$ is cyclic quartic,
with unique quadratic subfield $\Q(\sqrt5)=\Q(\varphi)$.
The quadratic subfields of $\Q(\sqrt2,\sqrt3)$ are
$\Q(\sqrt2)$, $\Q(\sqrt3)$, and $\Q(\sqrt6)$. The intersection of the two fields is a subfield of $\Q(a)$, hence $\Q$, $\Q(\sqrt5)$ or $\Q(a)$; it is not $\Q(a)$, which is cyclic of degree $4$ while $\Q(\sqrt2,\sqrt3)$ is not cyclic, and it is not $\Q(\sqrt5)$, which is not a subfield of $\Q(\sqrt2,\sqrt3)$. So the intersection is $\Q$, and since both fields are Galois over $\Q$, their compositum $K$ has degree
$4\cdot4=16$. Hence the monomials \eqref{eq:fieldbasis} form a basis of $K$,
and coefficientwise equality is exact equality in $K$.

No floating-point equality or tolerance is used. Useful exact inverses
for constructing the coordinates are
\[
	\varphi^{-1}=\varphi-1,\qquad
	a^{-1}=\frac{a(3-\varphi)}5,\qquad
	b^{-1}=b/3,\qquad c^{-1}=c/2.
\]
The program checks these relations and inverses, and the commutativity and associativity of the reduction on all $16^3$ triples of basis elements.

\subsection{Coordinates and the 21 cases}

For the tetrahedral case, take
\begin{align*}
	V_T & =\{(\epsilon_1,\epsilon_2,\epsilon_3)/b\colon
	\epsilon_i\in\{-1,1\},\ \epsilon_1\epsilon_2\epsilon_3=1\},           \\
	E_T & =\{\pm \vec{e}_1,\pm \vec{e}_2,\pm \vec{e}_3\},\qquad F_T=-V_T.
\end{align*}
For the octahedral case, take
\begin{align*}
	V_O & =\{\pm \vec{e}_1,\pm \vec{e}_2,\pm \vec{e}_3\}, \\
	E_O & =\{(\sigma \vec{e}_i+\tau \vec{e}_j)/c\colon
	1\leq i<j\leq3,\ \sigma,\tau\in\{-1,1\}\},            \\
	F_O & =\{(\epsilon_1,\epsilon_2,\epsilon_3)/b\colon
	\epsilon_i\in\{-1,1\}\}.
\end{align*}
Thus $V_O$ is the octahedron, $E_O$ its normalized edge-midpoint set
(the cuboctahedron), and $F_O$ the normalized face-center set (the
dual cube).

For the icosahedral case let
\[
	W=\{(0,\pm1,\pm\varphi),\ (\pm\varphi,0,\pm1),
	\ (\pm1,\pm\varphi,0)\},
\]
with independent signs. There are 12 vectors in $W$. Pairs $\vec{w},\vec{w}'$ with
$\vec{w}\cdot \vec{w}'=\varphi$ form the 30 edges, and triples with all three
pairwise inner products equal to $\varphi$ form the 20 triangular
faces. Define
\[
	V_I=\{\vec{w}/a\colon\vec{w}\in W\},\qquad
	E_I=\{(\vec{w}+\vec{w}')/(2\varphi)\colon\vec{w}\cdot \vec{w}'=\varphi\},
\]
\[
	F_I=\{(\vec{w}+\vec{w}'+\vec{w}'')/(b\varphi^2)\colon
	\{\vec{w},\vec{w}',\vec{w}''\}\text{ is a face}\}.
\]
All these vectors have norm one, as checked exactly. The code checks
their cardinalities and distinctness as well, and that $W$ has exactly 30 such pairs and 20 such triples. For each type $T,O,I$,
it considers all seven nonempty unions of the three indicated pieces.

For precision, Table~\ref{tab:21} lists all cases, including the
geometric duplicates eliminated from Table~\ref{tab:hosts}. Its labels
use the standard octahedron convention above. The certificate uses a dual-cube convention for its internal $O$ case IDs: its
vertex and face orbits are interchanged, while its edge orbit is
unchanged. The certificate IDs are included in the table to make this
translation explicit. A mask is the sum of $1$, $2$, and $4$ for the
certificate's internal $V$, $E$, and $F$ pieces, respectively.

\begin{table}[htbp]
	\centering
	\small
	\begin{tabular}{c c r l c}
		\toprule
		Displayed set & Certificate ID & $m$ & Color-atom sizes & Representative in Table~\ref{tab:hosts} \\
		\midrule
		$T(V)$        & T1             & 4   & $4$              & $T(V)$                                  \\
		$T(E)$        & T2             & 6   & $2,2,2$          & $\cong O(V)$                            \\
		$T(VE)$       & T3             & 10  & $4,6$            & $T(VE)$                                 \\
		$T(F)$        & T4             & 4   & $4$              & $\cong T(V)$                            \\
		$T(VF)$       & T5             & 8   & $4,4$            & $\cong O(F)$                            \\
		$T(EF)$       & T6             & 10  & $6,4$            & $\cong T(VE)$                           \\
		$T(VEF)$      & T7             & 14  & $4,6,4$          & $\cong O(VF)$                           \\
		$O(F)$        & O1             & 8   & $4,4$            & $O(F)$                                  \\
		$O(E)$        & O2             & 12  & $12$             & $O(E)$                                  \\
		$O(EF)$       & O3             & 20  & $8,12$           & $O(EF)$                                 \\
		$O(V)$        & O4             & 6   & $2,2,2$          & $O(V)$                                  \\
		$O(VF)$       & O5             & 14  & $4,4,6$          & $O(VF)$                                 \\
		$O(VE)$       & O6             & 18  & $12,6$           & $O(VE)$                                 \\
		$O(VEF)$      & O7             & 26  & $8,12,6$         & $O(VEF)$                                \\
		$I(V)$        & I1             & 12  & $12$             & $I(V)$                                  \\
		$I(E)$        & I2             & 30  & $30$             & $I(E)$                                  \\
		$I(VE)$       & I3             & 42  & $12,30$          & $I(VE)$                                 \\
		$I(F)$        & I4             & 20  & $20$             & $I(F)$                                  \\
		$I(VF)$       & I5             & 32  & $12,20$          & $I(VF)$                                 \\
		$I(EF)$       & I6             & 50  & $30,20$          & $I(EF)$                                 \\
		$I(VEF)$      & I7             & 62  & $12,30,20$       & $I(VEF)$                                \\
		\bottomrule
	\end{tabular}
	\caption{\emph{Every finite case verified by the program.} Within a row's
		list of color-atom sizes, the order only reflects the vertex enumeration. The
		last column gives the corresponding distinct support in
		Table~\ref{tab:hosts}; rows marked $\cong$ are congruent to the indicated set (verified in Appendix~\ref{sec:solids-check}).}
	\label{tab:21}
\end{table}

\subsection{The solids, duality and congruences}\label{sec:solids-check}

Besides the coordinates above, the program builds the five Platonic solids from their standard vertex coordinates alone: the tetrahedron $\{(\epsilon_1,\epsilon_2,\epsilon_3)\colon\epsilon_i\in\{-1,1\},\ \epsilon_1\epsilon_2\epsilon_3=1\}$, the cube $\{-1,1\}^3$, the octahedron $\{\pm\vec{e}_1,\pm\vec{e}_2,\pm\vec{e}_3\}$, the icosahedron $W$, and the dodecahedron, whose vertices are those of the cube together with the cyclic permutations of $(0,\pm\varphi,\pm\varphi^{-1})$. All of these coordinates lie in $\Q(\sqrt5)$, so signs are decided exactly. For each solid, the edges are the pairs of vertices at maximal inner product, and the faces are the vertex sets of the supporting planes through three vertices; the program checks the numbers of vertices, edges and faces, that each face has the right number of vertices and edges, and that each edge lies in exactly two faces. It then verifies the following.
\begin{enumerate}
	\item \emph{Identification.} The sets $V_T,E_T,F_T$, $V_O,E_O,F_O$ and $V_I,E_I,F_I$ above consist exactly of the unit vectors in the directions of the vertices, edge midpoints and face centers of the tetrahedron, the octahedron and the icosahedron, respectively.
	\item \emph{Duality} (Appendix~\ref{sec:uncolored}). The vertices of the cube and of the dodecahedron point in the directions of the face centers of the octahedron and of the icosahedron, respectively, and vice versa, and dual solids have the same edge-midpoint directions. The face centers of the tetrahedron point in the directions of the antipodes of its vertices, which form the image of the tetrahedron under a quarter-turn about a coordinate axis. Hence the tetrahedron, the octahedron and the icosahedron give all Platonic balanced sets.
	\item \emph{Congruences} (Appendix~\ref{sec:uncolored}, Table~\ref{tab:21}). The quarter-turn about the $z$-axis maps $T(V)$ onto $T(F)$ and $T(VE)$ onto $T(EF)$, and $T(E)=O(V)$, $T(VF)=O(F)$ and $T(VEF)=O(VF)$, so the 21 descriptions give the 16 sets of Table~\ref{tab:hosts}. These have pairwise distinct sizes, except for the two pairs separated by their minimum distances (Appendix~\ref{sec:finite-procedure}), so they are pairwise non-congruent.
\end{enumerate}

\subsection{Finite proof procedure and certificate}\label{sec:finite-procedure}

For each support $S=\{\vec{v}_1,\ldots,\vec{v}_m\}$ the exact verification performs
the following steps.
\begin{enumerate}
	\item Compute $G_{ij}=\vec{v}_i\cdot \vec{v}_j$ in $K$ and group each row into
	      shells of equal exact inner product. Verify unit norms and distinctness.
	\item For every root $i$ and shell value $t$, verify ordinary balance
	      by summing the tangent vectors $\vec{v}_j-t\vec{v}_i$. The shells $t=\pm1$
	      have zero tangent vectors and impose no color equalities. In every other shell, verify that the
	      tangent vectors are pairwise distinct and all have length $\sqrt{1-t^2}$, as used in Lemma~\ref{lem:shellrules}.
	\item Apply the three rules of Lemma~\ref{lem:shellrules}. For the rule on four-point shells,
	      verify that the four tangent vectors form two opposite pairs. For the
	      pentagon rule, regularity is certified by checking that the pairwise
	      inner products of its tangent vectors belong to
	      \[
		      (1-t^2)\left\{\frac{\varphi-1}{2},-\frac{\varphi}{2}\right\},
	      \]
	      the values $ (1-t^2)\cos(2\pi/5)$ and
	      $(1-t^2)\cos(4\pi/5)$. Maintain the resulting connected components
	      and record a spanning forest of color-equality implications, with
	      their witnessing roots and shells.
	\item Compare the components with the expected color atoms in
	      Table~\ref{tab:21}, and verify \eqref{eq:atombalance} for every color atom,
	      every root of the support, and every shell. This checks sufficiency, not merely
	      the listed necessary implications. Also verify that every color atom with two points is a pair of
	      antipodal points and every color atom with four points is a regular tetrahedron (pairwise inner
	      products $-1/3$), as described in \cref{fig:leech-color-atoms}.
	\item Check the subset-sum property of color-atom sizes. Verify explicit
	      proper coordinate symmetries inducing every required permutation of
	      equal-sized color atoms and an improper coordinate symmetry preserving
	      every color atom individually. Orthogonality, determinant, and the induced
	      vertex and color-atom permutations are verified exactly. Finally, verify that the quarter-turn
	      about the $z$-axis interchanges the two tetrahedral color atoms of $O(F)$ and $O(VF)$ and preserves
	      the octahedron (\cref{fig:leech-color-atoms}).
\end{enumerate}
In addition, for the four supports $O(E)$, $I(V)$, $O(EF)$ and $I(F)$ of Proposition~\ref{prop:uncolored}, the program verifies the minimum squared distances stated in its proof: the claimed value occurs among the squared distances $2-2\,\vec{u}\cdot\vec{v}$, $\vec{u}\neq\vec{v}$, and every other one is at least as large, the signs of the differences, which lie in $\Q(\sqrt5)$ or $\Q(\sqrt6)$, being decided exactly.
These steps prove Proposition~\ref{prop:finite} and the finite symmetry
claims used in Lemma~\ref{lem:colored}; together with Appendix~\ref{sec:solids-check}, they also verify the finite claims about the Platonic balanced sets made in Appendix~\ref{sec:uncolored} and in the proof of Proposition~\ref{prop:uncolored}. The certificate records the
finite supports, their color atoms and equality witnesses, the checked
symmetry and counting data, the congruences, the quarter-turn of \cref{fig:leech-color-atoms}, and the minimum squared distances. The infinite polygon families are not
delegated to finite computation; their proof is the analytic argument
in Appendix~\ref{sec:colored-general}.

\section{Proofs for the neural realization}\label{app:neural_proofs}
We first restate Theorem \ref{thm:neural-pairwise}.
\pairwiseseparation*

Before we prove the theorem, we will prove a lemma showing that the multiset and vector-multiset functions we use are injective on finite collections of inputs.

\begin{restatable}[Finite neural separation]{lemma}{finiteneuralseparation}\label{lem:finite-neural}
	Let $\sigma\colon\R\to\R$ be analytic and non-polynomial. For any finite
	collection of distinct finite multisets with elements from $\R^b$, the scalar map
	\[
		S\longmapsto\sum_{\vec{u}\in S}\sigma(\vec{\alpha}\cdot \vec{u}+\beta)
	\]
	is injective on that collection for almost every $(\vec{\alpha},\beta)$.
	Likewise, for any finite collection of distinct pairs
	$(\vec{v},S)$, where $\vec{v}\in\R^a$ and $S$ is a finite multiset of elements from $\R^b$,
	\[
		(\vec{v},S)\longmapsto \vec{\gamma}\cdot \vec{v}+\sum_{\vec{u}\in S}\sigma(\vec{\alpha}\cdot \vec{u}+\beta)
	\]
	is injective on that collection for almost every $(\vec{\alpha},\beta,\vec{\gamma})$.
	Both conclusions also hold for any positive output dimension (that is, when $\vec{\alpha}, \vec{\gamma}$ are matrices instead of vectors, and $\beta$ is a vector instead of a scalar).
\end{restatable}

\begin{proof}
	By the discriminatory property of continuous non-polynomial
	activations~\cite[Proposition~3.2]{amir2023injective}, two distinct
	finite multisets have different sums for some choice of $(\vec{\alpha},\beta)$.
	Their difference is therefore a nonzero analytic function of these
	parameters, whose zero set has Lebesgue measure zero \citep{mityagin2015zerosetrealanalytic}. Taking the
	finite union of these exceptional sets proves the first claim.
	For two distinct pairs, if their vector components differ, varying
	$\vec{\gamma}$ shows that the output difference is not identically zero. If
	their vector components agree, apply the first claim to the distinct
	multisets. The same zero-set argument proves the second claim.
	For vector outputs, one coordinate suffices, and its parameters can
	be chosen independently of the remaining coordinates.
\end{proof}

\begin{proof}[Proof of Theorem~\ref{thm:neural-pairwise}]
	Fix the two non-congruent clouds. We first construct one separating
	parameter choice. The analytic zero-set argument will then give the
	almost-everywhere conclusion.

	For a node in either cloud, let
	\[
		D_i=\left((\vec{q}_i,r_i),\,
		\multiset{(\vec{q}_j,r_j,\vec{x}_i\cdot \vec{x}_j)\colon j\in[n]}\right)
	\]
	be its first-layer input. By Lemma~\ref{lem:finite-neural}, choose
	$\phi_0$ so that $\vec{h}_i$ distinguishes all distinct $D_i$'s occurring
	in the two clouds. Thus $\vec{h}_i$ determines $\vec{q}_i,r_i$ and, if $r_i>0$,
	the color $C_\cloud(\vec{u}_i)$ of its ray, from \eqref{eq:rayprofile}:
	the particles on that ray are exactly those with
	$r_j>0$ and $\vec{x}_i\cdot \vec{x}_j=r_ir_j$.

	For any subsequent choice of $\psi_1,\psi_2$, write
	\begin{align*}
		\vec{g}_i & =(\|\vec{y}_{i,1}\|^2,\|\vec{y}_{i,2}\|^2,
		\vec{y}_{i,1}\cdot \vec{y}_{i,2}),                                                           \\
		M_i       & =\multiset{(\vec{h}_j,\vec{y}_{i,1}\cdot \vec{x}_j,\vec{y}_{i,2}\cdot \vec{x}_j,
		(\vec{y}_{i,1}\times \vec{y}_{i,2})\cdot \vec{x}_j)\colon j\in[n]},                          \\
		T_i       & =(\vec{g}_i,M_i).
	\end{align*}
	It suffices to choose the two $\psi$'s so that the multisets
	$\multiset{(\vec{h}_i,T_i)\colon i\in[n]}$ differ between the clouds.
	Indeed, Lemma~\ref{lem:finite-neural} then lets us choose $\phi_1$
	injective on the finitely many $T_i$'s and $\phi_2$ separating the
	resulting two multisets of $(\vec{h}_i,\vec{z}_i)$.

	\emph{Case 1: at least one cloud is not color-balanced.}
	Interchange the clouds if necessary so that this is $\cloud$.
	The construction in the proof of Theorem~\ref{thm:main} shows that in this case the functions $\psi_1,\psi_2$ can be chosen so that, for some $i$, the vectors $\vec{y}_{i,1},\vec{y}_{i,2}$ are linearly independent. It is sufficient to show that this is also the case when $\psi_1,\psi_2$ are the one-hidden-layer networks of the neural head, as in \eqref{eq:yexplicit}. If we can show this, then the argument in Theorem~\ref{thm:main} shows that the multisets $\multiset{(\vec{h}_i,T_i)\colon i\in[n]}$ of both clouds cannot agree, since that would imply congruence of the clouds.

	In the proof of Theorem~\ref{thm:main}, we choose $\psi_1=0$ so that $\vec{y}_{i,1}=\vec{x}_i$. We can make this choice in \eqref{eq:yexplicit} as well, by choosing $\vec{w}_1=0$. The proof then shows that there is a scalar function $f$ on the finitely many inputs
	$\vec{s}_j=(\vec{h}_i,\vec{h}_j,\vec{x}_i\cdot \vec{x}_j)$ such that
	\begin{equation}\label{eq:ineq}
		L(f)=\vec{x}_i\times\sum_{j=1}^n f(\vec{s}_j)\vec{x}_j\neq 0
	\end{equation}

	We now apply the universal approximation for
	shallow networks with continuous non-polynomial
	activations~\citep{pinkus1999approximation}: The finite set of distinct
	$\vec{s}_j$ is compact, and $f$ is continuous on it. Thus, it can be approximated uniformly on this set by shallow neural networks with the activation $\sigma$. This implies that the inequality \eqref{eq:ineq} holds when $f$ is replaced with a function of the form
	\[
		\widetilde f(\vec{s})=\sum_{\ell=1}^m c_\ell
		\sigma(\vec{a}_\ell^\top \vec{s}+b_\ell)
	\]
	and so we obtain $L(\widetilde f)\neq 0 $. Finally, $\widetilde f $ is a linear combination of functions of the form $\vec{s}\mapsto  \sigma(\vec{a}_\ell^\top \vec{s}+b_\ell) $, and $L$ is linear. Hence, there exists some $\ell$ for which
	\[L\left(  \sigma(\vec{a}_\ell^\top \vec{s}+b_\ell)  \right)\neq 0 .\]
	This gives us a realization of the architecture with $d[\psi]=1$ for which $\vec{y}_{i,1}\times \vec{y}_{i,2}\neq\vec{0}$. Larger $d[\psi]$ can be handled by appending zeros to the parameters of $\psi_1,\psi_2$.

	\emph{Case 2: both clouds are color-balanced.}
	Set $\psi_1=\psi_2=0$, so $\vec{y}_{i,1}=\vec{y}_{i,2}=\vec{x}_i$. The pair
	$(\vec{h}_i,T_i)$ then determines
	\[
		\left(\vec{h}_i,\,\multiset{(\vec{h}_j,\vec{x}_i\cdot \vec{x}_j)\colon j\in[n]}\right).
	\]
	We claim that equality of the multisets of these records would
	imply congruence. For $r_i>0$, let
	$m_i=|C_\cloud(\vec{u}_i)|$, the number of particles on its ray.
	This number, the color $C_\cloud(\vec{u}_i)$, and the radius $r_i$ are determined by $\vec{h}_i$.
	Consequently, the records determine the two counting measures
	\begin{align}
		\sum_{i\colon r_i>0}\frac{1}{m_i}\delta_{C_\cloud(\vec{u}_i)}
		 & =\sum_{\vec{u}\in S_\cloud}\delta_{C_\cloud(\vec{u})},\label{eq:neural-color-count} \\
		\sum_{\substack{i,j\colon  r_i>0,\,r_j>0}}
		\frac{1}{m_im_j}\delta_{(\vec{x}_i\cdot \vec{x}_j)/(r_ir_j)}
		 & =\sum_{(\vec{u},\vec{v})\in S_\cloud\times S_\cloud}
		\delta_{\vec{u}\cdot \vec{v}}.
		\label{eq:neural-angle-count}
	\end{align}
	Here $\delta_a$ denotes a unit point mass at the color or scalar $a$.
	The factors $1/m_i$ and $1/(m_im_j)$ count each ray and each ordered
	pair of rays exactly once, even with repeated positions or several
	particles on a ray. The records also determine
	$\multiset{\vec{q}_i\colon r_i=0}$. Equality of records between the two clouds
	would therefore give the two histograms in Lemma~\ref{lem:colored}
	and the same charges at the origin. That lemma supplies a color-preserving proper
	rotation; including the particles at
	the origin gives $\cloud\cong\othercloud$, a contradiction.
	This proves the required distinction in this case as well.

	Choose $\phi_1$ and $\phi_2$ as above to obtain a separating
	parameter $\vec{\theta}_*$. For the fixed pair of clouds, the function
	\[
		g(\vec{\theta})=\|E_{\vec{\theta}}(\cloud)-E_{\vec{\theta}}(\othercloud)\|^2
	\]
	is real analytic on the Euclidean space $\Theta$: the input radii
	are constants, and all subsequent operations are polynomial or
	analytic. Since $g(\vec{\theta}_*)>0$, its zero set has Lebesgue measure
	zero. This zero set is exactly the set of parameters failing to
	separate the fixed pair.
\end{proof}

\globalseparation*

\begin{proof}
	Continuity follows from the defining operations, and invariance follows
	from Theorem~\ref{thm:main}. For separation, apply the full finite
	witness theorem of Amir et al.~\cite[Theorem~A.2]{amir2023injective}.
	In its terminology, a function is $\sigma$-subanalytic when its
	graph is. This class contains
	semialgebraic and real analytic functions and is closed under finite
	compositions and arithmetic operations.

	Define, on $M\times M\times\Theta$,
	\[
		F((\cloud,\othercloud);\vec{\theta})
		=\|E_{\vec{\theta}}(\cloud)-E_{\vec{\theta}}(\othercloud)\|^2.
	\]
	This is jointly $\sigma$-subanalytic: the input norm maps are
	semialgebraic, and the remaining operations are polynomial or
	real analytic. For every fixed pair of clouds, $F$ is analytic
	in $\vec{\theta}$, and $\Theta$ is open and connected. Finally,
	Theorem~\ref{thm:neural-pairwise} and invariance show that its
	common zero set, over all $\vec{\theta}\in\Theta$, is precisely
	\[
		\mathcal N=\{(\cloud,\othercloud)\in M\times M\colon
		\cloud\cong\othercloud\}.
	\]
	The pair domain $M\times M$ is $\sigma$-subanalytic of dimension
	$2\delta$. The finite witness theorem therefore says that, for
	almost every choice of $2\delta+1$ parameters,
	this common zero set is already determined by those parameters.
	By \eqref{eq:multihead}, this is precisely the asserted global
	separation for $H=2\delta+1$. For larger $H$, the first $2\delta+1$ heads
	already suffice, giving the same almost-everywhere conclusion.
\end{proof}

We next restate and prove our universality result:

\begin{restatedcoruniversality}[Invariant universal approximation]
	\invariantapproximationbody
\end{restatedcoruniversality}

\begin{proof}
	Choose $\vec{\vartheta}$ in the full-measure set from
	Theorem~\ref{thm:neural-global}, and abbreviate the corresponding
	map by $\mathcal E$. This map is continuous and separates orbits, and so by \cite[Proposition~2.3]{dym2025low} there is a continuous function
	\[
		\widetilde f\colon \R^{H\cdot \dE}\longrightarrow\R
		\qquad\text{with}\qquad f=\widetilde f\circ\mathcal E.
	\]
	 Universal approximation with the
	continuous non-polynomial activation
	$\sigma$~\citep{pinkus1999approximation} gives a feed-forward neural network
	$\rho(\vec{v})=\vec{B}\sigma(\vec{A}\vec{v}+\vec{b})+\vec{c}$ approximating $\widetilde f$ within
	$\varepsilon$ on the compact set $\mathcal E(K)$. Composing with
	$\mathcal E$ proves the stated bound, and the composition is
	invariant because $\mathcal E$ is invariant.
\end{proof}

\section{Extension to other symmetries}\label{app:other_sym}
Our architecture and theory are $\SO(3)\times S_n $ invariant. In some settings, we are interested in additional symmetries: translations and/or reflections. To add translation symmetry, we can simply apply centralization as a pre-processing step: this means mapping each point $\vec{x}_i$ to $\vec{x}_i-n^{-1}\sum_{j=1}^n \vec{x}_j $. Our model composed with the centralization operator will have the same completeness and universality guarantees, with respect to the group of rotations, permutations \emph{and translations}. For a formal argument, see Section 2.6 in \citep{dym2025low}.

Similarly, reflection symmetry can be obtained by an averaging procedure: consider the linear operator
\[P[f](\Vec{X},\Vec{Q})=\frac{1}{2}\left(f(\Vec{X},\Vec{Q})+f(-\Vec{X},\Vec{Q}) \right),\]
where $\Vec{X}$ and $\Vec{Q} $ are the matrices of positions and charges, respectively. The operator $P$  projects $\SO(3) $ invariant functions into the space of $\mathrm{O}(3) $ invariant functions, and is fixed on $\mathrm{O}(3) $ invariant functions. Accordingly, if $f_\theta$ is some $\SO(3)$ invariant model, then $P[f_\theta]$ is a $\mathrm{O}(3) $ invariant model. If   $f_\theta$ is permutation- and translation-invariant, this invariance will be preserved. Finally, if $f_\theta$ is an $\epsilon$ approximation of a $\mathrm{O}(3) $ invariant function on some $\mathrm{O}(3) $ invariant domain, then $P[f_\theta]$ will be an $\epsilon$ approximation of $f=P[f]$ on the same domain. Thus, an analog of Corollary \ref{cor:neural-universality} can be obtained for this symmetrized model.

\section{Completeness of SEGNN(3)}\label{app:segnn}

Here, we show the completeness for the SEGNN(3) architecture.

That is, we define the family $\mathcal S_{n,d}^{(4)}$ and subsequently prove
\cref{thm:segnn-completeness}. We first describe its message-passing
layers, then show that four such layers approximate any continuous
realization of \eqref{eq:h}--\eqref{eq:E}. Applying this simulation
to a separating neural head gives completeness.

\subsection{The SEGNN(3) architecture}\label{app:segnn-architecture}

We use the steerable message and update functions of
\citet{brandstetter2022segnn}, specialized to $\SO(3)$ and the
steering attributes defined below. The original architecture also
takes care of reflection parities to obtain $\OO(3)$ equivariance.
Here, equivariance is required only under proper rotations.

\paragraph{Features and steering attributes.}
A \new{type-$0$ channel} is a scalar that stays fixed under rotations.
A \new{type-$1$ channel} is a vector $\vec{u}\in\R^3$ that transforms
as $\vec{u}\mapsto\vec{R}\vec{u}$, for rotations $\vec{R}\in\SO(3)$.
A node representation $\vec{f}_i^{(t)}$ concatenates a pre-defined
number of channels of each type. Assuming $c_0$ scalar channels and
$c_1$ vector channels, the node representation $\vec{f}_i^{(t)} \in \R^{c_0}\oplus(\R^3)^{c_1}$ and has
$D_f=c_0+3c_1$ components. Here, $\vec{h}_i$ denotes the
invariant feature in \eqref{eq:h}.

Following~\cref{sec:architecture}, for $n > 0$, we use all ordered pairs $(i,j)\in[n]^2$, including self-edges.
For the message from $j$ to $i$ and the update at $i$, respectively, we
choose the \new{steering attributes}
\begin{equation}\label{eq:segnn-attributes}
	\vec{a}_{ij}\coloneq(1,\vec{x}_j),
	\qquad
	\vec{a}_i\coloneq(1,\vec{x}_i).
\end{equation}
These attributes are computed from the input and reused in every
layer. Here, their scalar component stays fixed, and their vector
component rotates with the cloud, and also respects permutations
of the point indices. We use unnormalized positions, so the
attributes are defined at the origin.
As discussed in \cref{app:other_sym}, we can center before evaluating both the network
and the targets, giving a translation-invariant version.

The initial representation contains scalar channels
$(1,\vec{q}_i,r_i)$ and the vector channel $\vec{x}_i$, where
$r_i=\|\vec{x}_i\|$ is computed before message passing. Note that we could do an additional message-layer to incorporate the norm computation into our architecture. Additional scalar and vector channels are initialized to zero
to store quantities computed in later layers.

\paragraph{Message passing and readout.}
For $t\in[4]$, a SEGNN(3) message-passing layer has the form
\begin{equation}\label{eq:segnn-round}
	\begin{aligned}
		\vec{m}_{ij}^{(t)}
		 & \coloneq\mathsf{AGG}_t\!\left(
		\vec{f}_i^{(t-1)},\vec{f}_j^{(t-1)},\delta_{ij};
		\vec{a}_{ij}\right),              \\
		\vec{f}_i^{(t)}
		 & \coloneq\vec{f}_i^{(t-1)}+
		\mathsf{UPD}_t\!\left(
		\vec{f}_i^{(t-1)},\sum_{j=1}^n\vec{m}_{ij}^{(t)};
		\vec{a}_i\right),
	\end{aligned}
\end{equation}
where $\delta_{ij}\coloneq\|\vec{x}_j-\vec{x}_i\|^2$ is the
invariant edge input used in SEGNN. Our construction ignores it
by setting its input coefficients to zero. Messages
consist of the steering attributes and scalar and vector channels, where the sum is taken channel- or component-wise. The residual addition follows Appendix~C.2 of \citet{brandstetter2022segnn}.

Let $\vec{s}_i$ collect the scalar channels of $\vec{f}_i^{(4)}$.
The invariant readout is
\begin{equation}\label{eq:segnn-readout}
	N(\cloud)=\rho\!\left(\sum_{i=1}^n\eta(\vec{s}_i)\right),
\end{equation}
where $\eta$ and $\rho$ are rotation-invariant feed-forward neural networks,
each with one hidden Swish layer and an affine output layer.

\paragraph{Steerable internal layers.}
The functions $\mathsf{AGG}_t$ and $\mathsf{UPD}_t$ are \new{steerable
	feed-forward neural networks}. The first internal layer of
$\mathsf{AGG}_t$ receives the concatenated channels of
$\vec{f}_i^{(t-1)}$ and $\vec{f}_j^{(t-1)}$, together with
$\delta_{ij}$. For $\mathsf{UPD}_t$, it receives the channels of
$\vec{f}_i^{(t-1)}$ and the summed messages. Subsequent internal
layers receive the outputs of the preceding \new{internal layer}.

Let  $s_b\in\R$ and $\vec{u}_c\in\R^3$ be the scalar and
vector input channels of an internal layer. Its steering attribute
is $(1,\vec{a})$, with $\vec{a}=\vec{x}_j$ in a message function
and $\vec{a}=\vec{x}_i$ in an update function. Before activation,
the layer computes
\begin{align}
	S_r
	 & \coloneq\beta_r+\sum_b A_{rb}s_b
	+\sum_c B_{rc}(\vec{u}_c\cdot\vec{a})\in\R,
	\label{eq:segnn-scalar}             \\
	\vec{U}_r
	 & \coloneq\sum_c C_{rc}\vec{u}_c
	+\sum_b D_{rb}s_b\vec{a}
	+\sum_c F_{rc}(\vec{u}_c\times\vec{a})\in\R^3,
	\label{eq:segnn-vector}
\end{align}
where coefficients are learned scalars, which are shared across points and edges. The constant scalar component of
the attribute permits the products $s_b\,1=s_b$ and
$1\,\vec{u}_c=\vec{u}_c$, and hence ordinary linear combinations
within each type.

The dot and cross products are the scalar and vector components
of the Clebsch--Gordan tensor product of two type-$1$ channels.
Note that any constant normalization factors can be absorbed into the
learned coefficients. It holds that
\[
	(\vec{R}\vec{u})\cdot(\vec{R}\vec{a})=\vec{u}\cdot\vec{a},
	\qquad
	(\vec{R}\vec{u})\times(\vec{R}\vec{a})
	=\vec{R}(\vec{u}\times\vec{a}),
	\quad \vec{R}\in\SO(3).
\]
Thus, \eqref{eq:segnn-scalar} produces invariant scalars and
\eqref{eq:segnn-vector} produces equivariant vectors.

\paragraph{Activations and vector gates.}
A gated internal layer applies
\[
	\operatorname{Swish}(s)\coloneq s\operatorname{sigmoid}(s),
	\qquad
	\operatorname{sigmoid}(s)\coloneq(1+e^{-s})^{-1}
\]
to each scalar feature output. For each vector output $\vec{U}_r$,
it computes an additional scalar $L_r$ by
\eqref{eq:segnn-scalar}, with its own coefficients, and passes on
$\operatorname{sigmoid}(L_r)\vec{U}_r$. All three coordinates
are multiplied by the same invariant scalar, so the resulting
vector remains equivariant. The values $L_r$ are used to compute
these multipliers and are not retained as separate feature channels.
Consequently, a layer with $m_0$ scalar and $m_1$ vector feature
outputs computes $m_0+4m_1$ coordinates before activation and
gating, and passes on $m_0+3m_1$ feature coordinates.
Swish is the scalar activation in Appendix~C.2 of
\citet{brandstetter2022segnn}; the sigmoid vector gates follow
the authors' implementation.\footnote{See
	\href{https://github.com/RobDHess/Steerable-E3-GNN/blob/main/models/segnn/o3_building_blocks.py}
	{\texttt{O3TensorProductSwishGate}}.
	Constant normalization factors in the activations can be absorbed
	into adjacent learned coefficients.}

Each message function has two gated internal layers. Each update
function has one gated internal layer followed by an affine
steerable layer of the form
\cref{eq:segnn-scalar,eq:segnn-vector}, before the residual addition.
We allow arbitrary finite channel counts and do not use feature
normalization. Together with \cref{eq:segnn-attributes,eq:segnn-readout},
these choices specify $\mathcal S_{n,d}^{(4)}$.
Observe that every operation is rotation equivariant, and the shared functions
and sums preserve point permutations. The scalar readout therefore
makes every $N\in\mathcal S_{n,d}^{(4)}$ exactly $G_n$-invariant.

\subsection{Simulation and proof of completeness}\label{app:segnn-simulation}

Now, the following results show that we can approximate our complete architectures arbitrarily closely.

\begin{proposition}[Uniform simulation]\label[proposition]{prop:segnn-simulation}
	Fix $n,d\geq1$ and a realization $E\colon\cloudspace\to\R^p$ of
	\eqref{eq:h}--\eqref{eq:E} in which
	$\phi_0,\phi_1,\phi_2,\psi_1,\psi_2$ are continuous.
	For every compact $K\subseteq\cloudspace$ and every
	$\varepsilon>0$, there exists $N\in\mathcal S_{n,d}^{(4)}$ such that
	\[
		\sup_{\cloud\in K}\|N(\cloud)-E(\cloud)\|<\varepsilon.
	\]
	The features $\vec{h}_i$, $(\vec{y}_{i,1},\vec{y}_{i,2})$, and
	$\vec{z}_i$ can also be approximated uniformly, to any prescribed
	accuracy, in designated channels after message-passing layers
	one, two, and four, respectively.
\end{proposition}

\begin{proof}
	For fixed multiset size $n$, a continuous function
	of a vector $\vec{u}$ and a multiset $S$ admits approximations
	of the Deep Sets form
	\begin{equation}\label{eq:segnn-deepsets}
		\phi(\vec{u},S)\simeq
		\rho\!\left(\vec{u},\sum_{\vec{z}\in S}\eta(\vec{z})\right),
	\end{equation}
	where $\eta,\rho$ are rotation-invariant networks with one hidden Swish layer
	and an affine output layer. This follows from \citet{zaheer2017deepsets,amir2023injective,dym2025low}:
	for a suitable $\eta$, the map $F(\vec{u},S)=\bigl(\vec{u},\sum_{\vec{z}\in S}\eta(\vec{z})\bigr)$ is continuous
	and injective on the relevant compact set, so it has a continuous inverse $F^{-1}$ on its image, and
	$\phi=g\circ F$ with the continuous function $g=\phi\circ F^{-1}$.
	Here $\simeq$ means uniform approximation to arbitrary accuracy
	on the relevant compact set.

	For a steering attribute $(1,\vec{a})$, set
	$\vec{\xi}=((s_b)_b,(\vec{u}_c\cdot\vec{a})_c)$.
	Equation~\eqref{eq:segnn-scalar} can form any affine function of
	$\vec{\xi}$. Swish followed by an affine combination in the
	second internal layer therefore gives an ordinary one-hidden-layer
	rotation-invariant network on $\vec{\xi}$. It approximates any continuous
	function of these invariant inputs~\citep{pinkus1999approximation}. Likewise, the first internal
	layer can form copies of a vector $\vec{b}$, such as $\vec{a}$
	or $\vec{u}_c\times\vec{a}$, and multiply them by different
	sigmoid gates. Their linear combination in the second layer gives
	\[
		\sum_{a=1}^A c_a\,
		\operatorname{sigmoid}(\vec{\gamma}_a\cdot\vec{\xi}+\beta_a)
		\vec{b}
		\simeq w(\vec{\xi})\vec{b}
	\]
	for any continuous scalar function $w$, by scalar universal
	approximation. The error is uniform when $\vec{b}$ is bounded.

	For a scalar value $s$ computed before that activation, use two
	outputs $\operatorname{Swish}(s)$ and $\operatorname{Swish}(-s)$.
	Their difference is $s$, so subtracting the corresponding summed
	channels in the first affine operation of the update recovers
	the desired sum. For a vector message, set the final gate input
	to zero. The gate then multiplies by $1/2$; doubling the vector
	before this multiplication gives the desired message.
	Residual updates preserve previously computed channels by adding
	zero to them, and store new quantities in channels initialized
	to zero. We now describe the four layers using their target
	intermediate values, accounting for approximation errors below.

	\par\medskip\noindent\emph{1. Invariant features.}
	The message function $\mathsf{AGG}_1$ forms
	$s_{ij}\coloneq\vec{x}_i\cdot\vec{x}_j$ by
	\eqref{eq:segnn-scalar} and sends
	$\eta_0(\vec{q}_j,r_j,s_{ij})$ using the scalar construction above.
	The update function $\mathsf{UPD}_1$ combines the summed messages
	with $\vec{q}_i,r_i$, giving
	\[
		\vec{h}_i\simeq\rho_0\!\left(
		\vec{q}_i,r_i,\sum_{j=1}^n\eta_0(\vec{q}_j,r_j,s_{ij})\right).
	\]
	By \eqref{eq:segnn-deepsets}, this approximates \eqref{eq:h}.

	\par\medskip\noindent\emph{2. Equivariant vectors.}
	For $k\in[2]$, write
	$w_{ij,k}\coloneq\psi_k(\vec{h}_i,\vec{h}_j,s_{ij})$.
	The message function $\mathsf{AGG}_2$ recomputes $s_{ij}$ and
	uses the gated vector construction with $\vec{b}=\vec{x}_j$
	to approximate $w_{ij,k}\vec{x}_j$.
	The update function $\mathsf{UPD}_2$ adds $\vec{x}_i$ to each
	summed vector message, giving the targets
	\[
		\vec{y}_{i,k}=\vec{x}_i+\sum_{j=1}^n w_{ij,k}\vec{x}_j,
	\]
	as in \eqref{eq:y}.

	\par\medskip\noindent\emph{3. Cross product and Gram entries.}
	To keep the steering attributes fixed, expand the second factor
	of each product using \eqref{eq:y}. For
	$(k,k')\in\{(1,1),(1,2),(2,2)\}$, bilinearity gives
	\begin{align}
		\vec{c}_i
		 & =\vec{y}_{i,1}\times\vec{x}_i+
		\sum_{j=1}^n w_{ij,2}
		(\vec{y}_{i,1}\times\vec{x}_j)
		=\vec{y}_{i,1}\times\vec{y}_{i,2},
		\label{eq:segnn-cross}            \\
		G_{i,kk'}
		 & =\vec{y}_{i,k}\cdot\vec{x}_i+
		\sum_{j=1}^n w_{ij,k'}
		(\vec{y}_{i,k}\cdot\vec{x}_j)
		=\vec{y}_{i,k}\cdot\vec{y}_{i,k'}.
		\label{eq:segnn-gram}
	\end{align}
	The message function $\mathsf{AGG}_3$ recomputes the weights
	from the retained features and approximates each summand.
	The scalar summands are continuous functions of available
	scalar channels and dot products. The vector summands use the
	gated vector construction with
	$\vec{b}=\vec{y}_{i,1}\times\vec{x}_j$.
	The update function $\mathsf{UPD}_3$ forms the local terms using
	its steering vector $\vec{x}_i$ and adds the summed messages.
	It stores $\vec{c}_i$ and the three scalar entries
	$\vec{g}_i=(G_{i,11},G_{i,22},G_{i,12})$.

	\par\medskip\noindent\emph{4. Oriented measurements and readout.}
	The message function $\mathsf{AGG}_4$ forms
	\[
		\vec{p}_{ij}=\bigl(
		\vec{y}_{i,1}\cdot\vec{x}_j,
		\vec{y}_{i,2}\cdot\vec{x}_j,
		\vec{c}_i\cdot\vec{x}_j\bigr)
	\]
	by \eqref{eq:segnn-scalar}, and sends
	$\eta_1(\vec{h}_j,\vec{p}_{ij})$.
	The update function $\mathsf{UPD}_4$ and the scalar readout give
	\begin{align*}
		\vec{z}_i & \simeq\rho_1\!\left(
		\vec{g}_i,\sum_{j=1}^n\eta_1(\vec{h}_j,\vec{p}_{ij})\right), \\
		N(\cloud) & \simeq\rho_2\!\left(
		\sum_{i=1}^n\eta_2(\vec{h}_i,\vec{z}_i)\right),
	\end{align*}
	approximating \cref{eq:z,eq:E}. Each message keeps
	$\vec{h}_j$ and $\vec{p}_{ij}$ together, as required by
	the multiset in \eqref{eq:z}.

    The result then follows from the compactness and continuity assumptions via standard arguments.
\end{proof}

\begin{proof}[Proof of \cref{thm:segnn-completeness}]
	Every element of $\mathcal S_{n,d}^{(4)}$ is $G_n$-invariant by
	\cref{app:segnn-architecture}. Fix two non-congruent clouds
	$\cloud,\othercloud\in\cloudspace$.
	By \cref{thm:neural-pairwise}, there is a continuous neural head
	$E_{\vec{\theta}}$ of \eqref{eq:hexplicit}--\eqref{eq:Eexplicit}
	such that
	\[
		\Delta\coloneq
		\|E_{\vec{\theta}}(\cloud)-E_{\vec{\theta}}(\othercloud)\|>0.
	\]
	This head is a continuous realization of
	\eqref{eq:h}--\eqref{eq:E}.
	Apply \cref{prop:segnn-simulation} on
	$K=\{\cloud,\othercloud\}$ with error less than $\Delta/3$.
	The resulting $N\in\mathcal S_{n,d}^{(4)}$ satisfies
	\[
		\|N(\cloud)-N(\othercloud)\|
		>\Delta-2\Delta/3>0.
	\]
	Thus the family separates every non-congruent pair. The widths
	and parameters may depend on that pair, while four message-passing
	layers and types $0,1$ suffice throughout.
\end{proof}

\end{document}

%% file: figures/leech-polygon-poles.tex
\begingroup
\pgfmathsetlengthmacro{\leechunit}{\linewidth/540}
\begin{tikzpicture}[x=\leechunit,y=\leechunit,
  every node/.style={inner sep=0pt,outer sep=0pt},line cap=butt,line join=round]
\path[draw=none,fill={rgb,1:red,1;green,1;blue,1}] (0,0) rectangle (540,235);
\node[anchor=base west,font={\normalsize\bfseries},text={rgb,1:red,0.12549;green,0.17255;blue,0.21176}] at (9,219) {Color classes for polygons with both poles};
\node[anchor=base,font={\normalsize},text={rgb,1:red,0.12549;green,0.17255;blue,0.21176}] at (145,196) {(a) Triangle + poles};
\path[draw={rgb,1:red,0.83529;green,0.87059;blue,0.89804},line width={0.5*\leechunit},fill={rgb,1:red,0.98431;green,0.98824;blue,0.99216}] (145,128) circle[radius=49];
\draw[draw={rgb,1:red,0.81961;green,0.85882;blue,0.88627},line width={0.45*\leechunit}] (188.46275,120.45413) -- (186.55346,119.34018) -- (184.38798,118.27963) -- (181.97966,117.279) -- (179.34335,116.34447) -- (176.4953,115.48181) -- (173.45307,114.69632) -- (170.23542,113.99285) -- (166.86218,113.37574) -- (163.35416,112.8488) -- (159.73297,112.41527) -- (156.02096,112.07782) -- (152.24099,111.83854) -- (148.41638,111.6989) -- (144.57071,111.65976) -- (140.72769,111.72136) -- (136.911,111.88333) -- (133.14419,112.14466) -- (129.45047,112.50375) -- (125.85262,112.95837) -- (122.37282,113.50573) -- (119.03252,114.14246) -- (115.85233,114.86462) -- (112.85183,115.66776) -- (110.04954,116.54694) -- (107.46274,117.49672) -- (105.10736,118.51127) -- (102.99793,119.58431) -- (101.14746,120.70924) -- (99.56736,121.87913) -- (98.26736,123.08674) -- (97.25549,124.32465) -- (96.53797,125.58522) -- (96.11924,126.86068) -- (96.00188,128.14316) -- (96.18661,129.42476) -- (96.67228,130.69758) -- (97.45592,131.95376) -- (98.53268,133.18557) -- (99.89592,134.3854) -- (101.53725,135.54587) -- (103.44654,136.65982) -- (105.61202,137.72037) -- (108.02034,138.721) -- (110.65665,139.65553) -- (113.5047,140.51819) -- (116.54693,141.30368) -- (119.76458,142.00715) -- (123.13782,142.62426) -- (126.64584,143.1512) -- (130.26703,143.58473) -- (133.97904,143.92218) -- (137.75901,144.16146) -- (141.58362,144.3011) -- (145.42929,144.34024) -- (149.27231,144.27864) -- (153.089,144.11667) -- (156.85581,143.85534) -- (160.54953,143.49625) -- (164.14738,143.04163) -- (167.62718,142.49427) -- (170.96748,141.85754) -- (174.14767,141.13538) -- (177.14817,140.33224) -- (179.95046,139.45306) -- (182.53726,138.50328) -- (184.89264,137.48873) -- (187.00207,136.41569) -- (188.85254,135.29076) -- (190.43264,134.12087) -- (191.73264,132.91326) -- (192.74451,131.67535) -- (193.46203,130.41478) -- (193.88076,129.13932) -- (193.99812,127.85684) -- (193.81339,126.57524) -- (193.32772,125.30242) -- (192.54408,124.04624) -- (191.46732,122.81443) -- (190.10408,121.6146) -- (188.46275,120.45413);
\draw[draw={rgb,1:red,0.81333;green,0.84627;blue,0.87098},line width={0.38*\leechunit}]
  (150.20383,144.24846) -- (112.70308,127.09612)
  (153.06962,142.09216) -- (150.20383,144.24846)
  (155.93541,139.93586) -- (153.06962,142.09216)
  (158.80121,137.77956) -- (155.93541,139.93586)
  (161.667,135.62326) -- (158.80121,137.77956)
  (164.53279,133.46696) -- (161.667,135.62326)
  (167.39859,131.31066) -- (164.53279,133.46696)
  (170.26438,129.15436) -- (167.39859,131.31066)
  (173.13017,126.99806) -- (170.26438,129.15436);
\draw[draw={rgb,1:red,0.73333;green,0.78039;blue,0.81569},line width={0.38*\leechunit}]
  (112.70308,127.09612) -- (104.36957,123.2845) -- (100.20282,121.37868) -- (177.5608,118.62335) -- (184.59335,118.37286) -- (173.13017,126.99806);
\path[draw={rgb,1:red,1;green,1;blue,1},line width={0.4*\leechunit},fill={rgb,1:red,0.23;green,0.57424;blue,0.76749}] (150.20383,144.24846) circle[radius=3.5];
\path[draw={rgb,1:red,1;green,1;blue,1},line width={0.4*\leechunit},fill={rgb,1:red,0.87318;green,0.51384;blue,0.23}] (145,81.80502) circle[radius=3.5];
\path[draw={rgb,1:red,1;green,1;blue,1},line width={0.4*\leechunit},fill={rgb,1:red,0.83529;green,0.36863;blue,0}] (145,174.19498) circle[radius=3.5];
\path[draw={rgb,1:red,1;green,1;blue,1},line width={0.4*\leechunit},fill={rgb,1:red,0;green,0.44706;blue,0.69804}] (100.20282,121.37868) circle[radius=3.5];
\path[draw={rgb,1:red,1;green,1;blue,1},line width={0.4*\leechunit},fill={rgb,1:red,0;green,0.44706;blue,0.69804}] (184.59335,118.37286) circle[radius=3.5];
\node[anchor=base,font={\normalsize\bfseries},text={rgb,1:red,0;green,0.44706;blue,0.69804}] at (133.5,57) {3};
\node[anchor=base,font={\small},text={rgb,1:red,0.35686;green,0.40784;blue,0.44706}] at (145,57) {+};
\node[anchor=base,font={\normalsize\bfseries},text={rgb,1:red,0.83529;green,0.36863;blue,0}] at (156.5,57) {2};
\node[anchor=base,font={\small},text={rgb,1:red,0.12549;green,0.17255;blue,0.21176}] at (145,38) {Monochromatic cycle};
\node[anchor=base,font={\small},text={rgb,1:red,0.35686;green,0.40784;blue,0.44706}] at (145,20) {One color for the pole pair};
\node[anchor=base,font={\normalsize},text={rgb,1:red,0.12549;green,0.17255;blue,0.21176}] at (395,196) {(b) Square + poles};
\path[draw={rgb,1:red,0.83529;green,0.87059;blue,0.89804},line width={0.5*\leechunit},fill={rgb,1:red,0.98431;green,0.98824;blue,0.99216}] (395,128) circle[radius=49];
\draw[draw={rgb,1:red,0.81961;green,0.85882;blue,0.88627},line width={0.45*\leechunit}] (438.46275,120.45413) -- (436.55346,119.34018) -- (434.38798,118.27963) -- (431.97966,117.279) -- (429.34335,116.34447) -- (426.4953,115.48181) -- (423.45307,114.69632) -- (420.23542,113.99285) -- (416.86218,113.37574) -- (413.35416,112.8488) -- (409.73297,112.41527) -- (406.02096,112.07782) -- (402.24099,111.83854) -- (398.41638,111.6989) -- (394.57071,111.65976) -- (390.72769,111.72136) -- (386.911,111.88333) -- (383.14419,112.14466) -- (379.45047,112.50375) -- (375.85262,112.95837) -- (372.37282,113.50573) -- (369.03252,114.14246) -- (365.85233,114.86462) -- (362.85183,115.66776) -- (360.04954,116.54694) -- (357.46274,117.49672) -- (355.10736,118.51127) -- (352.99793,119.58431) -- (351.14746,120.70924) -- (349.56736,121.87913) -- (348.26736,123.08674) -- (347.25549,124.32465) -- (346.53797,125.58522) -- (346.11924,126.86068) -- (346.00188,128.14316) -- (346.18661,129.42476) -- (346.67228,130.69758) -- (347.45592,131.95376) -- (348.53268,133.18557) -- (349.89592,134.3854) -- (351.53725,135.54587) -- (353.44654,136.65982) -- (355.61202,137.72037) -- (358.02034,138.721) -- (360.65665,139.65553) -- (363.5047,140.51819) -- (366.54693,141.30368) -- (369.76458,142.00715) -- (373.13782,142.62426) -- (376.64584,143.1512) -- (380.26703,143.58473) -- (383.97904,143.92218) -- (387.75901,144.16146) -- (391.58362,144.3011) -- (395.42929,144.34024) -- (399.27231,144.27864) -- (403.089,144.11667) -- (406.85581,143.85534) -- (410.54953,143.49625) -- (414.14738,143.04163) -- (417.62718,142.49427) -- (420.96748,141.85754) -- (424.14767,141.13538) -- (427.14817,140.33224) -- (429.95046,139.45306) -- (432.53726,138.50328) -- (434.89264,137.48873) -- (437.00207,136.41569) -- (438.85254,135.29076) -- (440.43264,134.12087) -- (441.73264,132.91326) -- (442.74451,131.67535) -- (443.46203,130.41478) -- (443.88076,129.13932) -- (443.99812,127.85684) -- (443.81339,126.57524) -- (443.32772,125.30242) -- (442.54408,124.04624) -- (441.46732,122.81443) -- (440.10408,121.6146) -- (438.46275,120.45413);
\draw[draw={rgb,1:red,0.81333;green,0.84627;blue,0.87098},line width={0.38*\leechunit}]
  (423.86809,141.20387) -- (401.04761,140.01163) -- (355.40665,137.62714) -- (359.87551,128.11422)
  (424.76186,139.30129) -- (423.86809,141.20387)
  (425.65564,137.3987) -- (424.76186,139.30129)
  (426.54941,135.49612) -- (425.65564,137.3987)
  (427.44318,133.59353) -- (426.54941,135.49612)
  (428.33695,131.69095) -- (427.44318,133.59353)
  (429.23072,129.78837) -- (428.33695,131.69095)
  (430.12449,127.88578) -- (429.23072,129.78837);
\draw[draw={rgb,1:red,0.73333;green,0.78039;blue,0.81569},line width={0.38*\leechunit}]
  (359.87551,128.11422) -- (366.13191,114.79613) -- (388.95239,115.98837) -- (434.59335,118.37286) -- (430.12449,127.88578);
\path[draw={rgb,1:red,1;green,1;blue,1},line width={0.4*\leechunit},fill={rgb,1:red,0.87318;green,0.51384;blue,0.23}] (423.86809,141.20387) circle[radius=3.5];
\path[draw={rgb,1:red,1;green,1;blue,1},line width={0.4*\leechunit},fill={rgb,1:red,0.23;green,0.57424;blue,0.76749}] (355.40665,137.62714) circle[radius=3.5];
\path[draw={rgb,1:red,1;green,1;blue,1},line width={0.4*\leechunit},fill={rgb,1:red,0.23;green,0.7071;blue,0.57725}] (395,81.80502) circle[radius=3.5];
\path[draw={rgb,1:red,1;green,1;blue,1},line width={0.4*\leechunit},fill={rgb,1:red,0;green,0.61961;blue,0.45098}] (395,174.19498) circle[radius=3.5];
\path[draw={rgb,1:red,1;green,1;blue,1},line width={0.4*\leechunit},fill={rgb,1:red,0;green,0.44706;blue,0.69804}] (434.59335,118.37286) circle[radius=3.5];
\path[draw={rgb,1:red,1;green,1;blue,1},line width={0.4*\leechunit},fill={rgb,1:red,0.83529;green,0.36863;blue,0}] (366.13191,114.79613) circle[radius=3.5];
\node[anchor=base,font={\normalsize\bfseries},text={rgb,1:red,0;green,0.44706;blue,0.69804}] at (372,57) {2};
\node[anchor=base,font={\small},text={rgb,1:red,0.35686;green,0.40784;blue,0.44706}] at (383.5,57) {+};
\node[anchor=base,font={\normalsize\bfseries},text={rgb,1:red,0.83529;green,0.36863;blue,0}] at (395,57) {2};
\node[anchor=base,font={\small},text={rgb,1:red,0.35686;green,0.40784;blue,0.44706}] at (406.5,57) {+};
\node[anchor=base,font={\normalsize\bfseries},text={rgb,1:red,0;green,0.61961;blue,0.45098}] at (418,57) {2};
\node[anchor=base,font={\small},text={rgb,1:red,0.12549;green,0.17255;blue,0.21176}] at (395,38) {Two alternating cycle classes};
\node[anchor=base,font={\small},text={rgb,1:red,0.35686;green,0.40784;blue,0.44706}] at (395,20) {One color for the pole pair};
\draw[draw={rgb,1:red,0.88235;green,0.90196;blue,0.91765},line width={0.45*\leechunit}]
  (270,18) -- (270,201);
\pgfresetboundingbox
\path[use as bounding box] (0,0) rectangle (540,235);
\end{tikzpicture}
\endgroup

%% file: figures/leech-color-atoms.tex
\begingroup
\pgfmathsetlengthmacro{\leechunit}{\linewidth/540}
\begin{tikzpicture}[x=\leechunit,y=\leechunit,
  every node/.style={inner sep=0pt,outer sep=0pt},line cap=butt,line join=round]
\path[draw=none,fill={rgb,1:red,1;green,1;blue,1}] (0,0) rectangle (540,246);
\node[anchor=base west,font={\normalsize\bfseries},text={rgb,1:red,0.12549;green,0.17255;blue,0.21176}] at (9,232) {Color atoms in the three exceptional supports};
\node[anchor=base,font={\normalsize\itshape},text={rgb,1:red,0.12549;green,0.17255;blue,0.21176}] at (90,211) {(a)  $O(V)$};
\node[anchor=base,font={\small},text={rgb,1:red,0.35686;green,0.40784;blue,0.44706}] at (90,197) {Octahedron};
\path[draw={rgb,1:red,0.83529;green,0.87059;blue,0.89804},line width={0.5*\leechunit},fill={rgb,1:red,0.98431;green,0.98824;blue,0.99216}] (90,133) circle[radius=48];
\draw[draw={rgb,1:red,0.81333;green,0.84627;blue,0.87098},line width={0.38*\leechunit}]
  (112.1654,147.19847) -- (101.37521,146.06403) -- (85.18992,144.36239) -- (74.39973,143.22795) -- (58.21444,141.52631) -- (47.42424,140.39187)
  (112.1654,147.19847) -- (97.38847,107.56467) -- (90,87.74778)
  (112.1654,147.19847) -- (97.38847,167.90097) -- (95.54135,170.48878)
  (112.1654,147.19847) -- (125.7723,132.80491)
  (47.42424,140.39187) -- (86.45202,92.13478) -- (90,87.74778)
  (47.42424,140.39187) -- (72.2601,162.47708)
  (47.42424,140.39187) -- (54.2277,133.19509)
  (90,87.74778) -- (107.7399,103.52292)
  (90,87.74778) -- (84.45865,95.51122);
\draw[draw={rgb,1:red,0.94235;green,0.77902;blue,0.65},line width={0.65*\leechunit}]
  (112.1654,147.19847) -- (97.38847,137.73282) -- (90,133);
\draw[draw={rgb,1:red,0.65;green,0.80647;blue,0.89431},line width={0.65*\leechunit}]
  (47.42424,140.39187) -- (90,133);
\draw[draw={rgb,1:red,0.65;green,0.86686;blue,0.80784},line width={0.65*\leechunit}]
  (90,87.74778) -- (90,133);
\draw[draw={rgb,1:red,0.73333;green,0.78039;blue,0.81569},line width={0.38*\leechunit}]
  (84.45865,95.51122) -- (69.68172,116.21372) -- (67.8346,118.80153)
  (107.7399,103.52292) -- (132.57576,125.60813)
  (54.2277,133.19509) -- (67.8346,118.80153)
  (72.2601,162.47708) -- (82.90404,171.94217) -- (90,178.25222)
  (125.7723,132.80491) -- (132.57576,125.60813)
  (95.54135,170.48878) -- (90,178.25222)
  (90,178.25222) -- (129.02778,129.99513) -- (132.57576,125.60813)
  (90,178.25222) -- (77.07018,143.57266) -- (67.8346,118.80153)
  (132.57576,125.60813) -- (94.81008,121.63761) -- (78.62479,119.93597) -- (67.8346,118.80153);
\draw[draw={rgb,1:red,0.5;green,0.8098;blue,0.72549},line width={0.65*\leechunit}]
  (90,133) -- (90,178.25222);
\draw[draw={rgb,1:red,0.5;green,0.72353;blue,0.84902},line width={0.65*\leechunit}]
  (90,133) -- (125.4798,126.8401) -- (132.57576,125.60813);
\draw[draw={rgb,1:red,0.91765;green,0.68431;blue,0.5},line width={0.65*\leechunit}]
  (90,133) -- (71.52883,121.16795) -- (67.8346,118.80153);
\path[draw={rgb,1:red,1;green,1;blue,1},line width={0.4*\leechunit},fill={rgb,1:red,0.87318;green,0.51384;blue,0.23}] (112.1654,147.19847) circle[radius=3.35];
\path[draw={rgb,1:red,1;green,1;blue,1},line width={0.4*\leechunit},fill={rgb,1:red,0.23;green,0.57424;blue,0.76749}] (47.42424,140.39187) circle[radius=3.35];
\path[draw={rgb,1:red,1;green,1;blue,1},line width={0.4*\leechunit},fill={rgb,1:red,0.23;green,0.7071;blue,0.57725}] (90,87.74778) circle[radius=3.35];
\path[draw={rgb,1:red,1;green,1;blue,1},line width={0.4*\leechunit},fill={rgb,1:red,0;green,0.61961;blue,0.45098}] (90,178.25222) circle[radius=3.35];
\path[draw={rgb,1:red,1;green,1;blue,1},line width={0.4*\leechunit},fill={rgb,1:red,0;green,0.44706;blue,0.69804}] (132.57576,125.60813) circle[radius=3.35];
\path[draw={rgb,1:red,1;green,1;blue,1},line width={0.4*\leechunit},fill={rgb,1:red,0.83529;green,0.36863;blue,0}] (67.8346,118.80153) circle[radius=3.35];
\node[anchor=base,font={\normalsize\bfseries},text={rgb,1:red,0;green,0.44706;blue,0.69804}] at (67,63) {2};
\node[anchor=base,font={\small},text={rgb,1:red,0.35686;green,0.40784;blue,0.44706}] at (78.5,63) {+};
\node[anchor=base,font={\normalsize\bfseries},text={rgb,1:red,0.83529;green,0.36863;blue,0}] at (90,63) {2};
\node[anchor=base,font={\small},text={rgb,1:red,0.35686;green,0.40784;blue,0.44706}] at (101.5,63) {+};
\node[anchor=base,font={\normalsize\bfseries},text={rgb,1:red,0;green,0.61961;blue,0.45098}] at (113,63) {2};
\node[anchor=base,font={\small},text={rgb,1:red,0.12549;green,0.17255;blue,0.21176}] at (90,48) {Antipodal pairs};
\node[anchor=base,font={\small},text={rgb,1:red,0.35686;green,0.40784;blue,0.44706}] at (90,21) {Pairs permute};
\draw[draw={rgb,1:red,0.88235;green,0.90196;blue,0.91765},line width={0.45*\leechunit}]
  (180,17) -- (180,215);
\node[anchor=base,font={\normalsize\itshape},text={rgb,1:red,0.12549;green,0.17255;blue,0.21176}] at (270,211) {(b)  $O(F)$};
\node[anchor=base,font={\small},text={rgb,1:red,0.35686;green,0.40784;blue,0.44706}] at (270,197) {Cube};
\path[draw={rgb,1:red,0.83529;green,0.87059;blue,0.89804},line width={0.5*\leechunit},fill={rgb,1:red,0.98431;green,0.98824;blue,0.99216}] (270,133) circle[radius=48];
\draw[draw={rgb,1:red,0.81333;green,0.84627;blue,0.87098},line width={0.38*\leechunit}]
  (258.21608,119.3388) -- (258.21608,171.59157)
  (258.21608,119.3388) -- (303.28147,111.51469) -- (307.37832,110.8034)
  (258.21608,119.3388) -- (234.75454,104.31008) -- (232.62168,102.94383)
  (258.21608,171.59157) -- (299.18462,164.47874) -- (307.37832,163.05617) -- (307.37832,110.8034)
  (258.21608,171.59157) -- (245.41888,163.39408) -- (243.28601,162.02784)
  (307.37832,110.8034) -- (296.71399,103.97216);
\draw[draw={rgb,1:red,0.65;green,0.80647;blue,0.89431},line width={0.65*\leechunit}]
  (258.21608,119.3388) -- (307.37832,163.05617)
  (258.21608,119.3388) -- (241.15314,143.244)
  (258.21608,119.3388) -- (273.92797,102.71855)
  (301.1486,162.40121) -- (307.37832,163.05617)
  (307.37832,163.05617) -- (305.24546,157.33553);
\draw[draw={rgb,1:red,0.94235;green,0.77902;blue,0.65},line width={0.65*\leechunit}]
  (258.21608,171.59157) -- (266.40978,161.46021) -- (278.70035,146.26317) -- (286.89405,136.13181) -- (307.37832,110.8034)
  (258.21608,171.59157) -- (247.55174,142.98835) -- (234.75454,108.66447)
  (258.21608,171.59157) -- (266.07203,163.28145)
  (307.37832,110.8034) -- (282.45944,108.18355) -- (238.8514,103.59879)
  (307.37832,110.8034) -- (298.84686,122.756);
\draw[draw={rgb,1:red,0.73333;green,0.78039;blue,0.81569},line width={0.38*\leechunit}]
  (307.37832,163.05617) -- (292.44826,153.49243) -- (281.78392,146.6612)
  (296.71399,103.97216) -- (286.04966,97.14092) -- (281.78392,94.40843)
  (232.62168,107.29823) -- (232.62168,102.94383) -- (261.29965,97.96485) -- (281.78392,94.40843)
  (243.28601,162.02784) -- (232.62168,155.1966)
  (232.62168,111.65262) -- (232.62168,107.29823)
  (232.62168,116.00702) -- (232.62168,111.65262)
  (232.62168,120.36142) -- (232.62168,116.00702)
  (232.62168,124.71582) -- (232.62168,120.36142)
  (232.62168,129.07021) -- (232.62168,124.71582)
  (232.62168,133.42461) -- (232.62168,129.07021)
  (232.62168,137.77901) -- (232.62168,133.42461)
  (232.62168,142.13341) -- (232.62168,137.77901)
  (232.62168,146.4878) -- (232.62168,142.13341)
  (232.62168,150.8422) -- (232.62168,146.4878)
  (232.62168,155.1966) -- (232.62168,150.8422)
  (232.62168,155.1966) -- (261.29965,150.21761) -- (281.78392,146.6612)
  (281.78392,94.40843) -- (281.78392,146.6612);
\draw[draw={rgb,1:red,0.91765;green,0.68431;blue,0.5},line width={0.65*\leechunit}]
  (266.07203,163.28145) -- (281.78392,146.6612)
  (234.75454,108.66447) -- (232.62168,102.94383)
  (232.62168,102.94383) -- (238.8514,103.59879)
  (232.62168,102.94383) -- (240.81538,110.23006) -- (277.68707,143.01808) -- (281.78392,146.6612)
  (298.84686,122.756) -- (281.78392,146.6612);
\draw[draw={rgb,1:red,0.5;green,0.72353;blue,0.84902},line width={0.65*\leechunit}]
  (301.1486,162.40121) -- (257.54056,157.81645) -- (245.08112,156.50653) -- (232.62168,155.1966)
  (241.15314,143.244) -- (232.62168,155.1966)
  (305.24546,157.33553) -- (292.44826,123.01165) -- (281.78392,94.40843)
  (273.92797,102.71855) -- (281.78392,94.40843)
  (232.62168,155.1966) -- (249.00909,134.93387) -- (277.68707,99.47411) -- (281.78392,94.40843);
\path[draw={rgb,1:red,1;green,1;blue,1},line width={0.4*\leechunit},fill={rgb,1:red,0.23;green,0.57424;blue,0.76749}] (258.21608,119.3388) circle[radius=3.35];
\path[draw={rgb,1:red,1;green,1;blue,1},line width={0.4*\leechunit},fill={rgb,1:red,0.87318;green,0.51384;blue,0.23}] (258.21608,171.59157) circle[radius=3.35];
\path[draw={rgb,1:red,1;green,1;blue,1},line width={0.4*\leechunit},fill={rgb,1:red,0.87318;green,0.51384;blue,0.23}] (307.37832,110.8034) circle[radius=3.35];
\path[draw={rgb,1:red,1;green,1;blue,1},line width={0.4*\leechunit},fill={rgb,1:red,0.23;green,0.57424;blue,0.76749}] (307.37832,163.05617) circle[radius=3.35];
\path[draw={rgb,1:red,1;green,1;blue,1},line width={0.4*\leechunit},fill={rgb,1:red,0.83529;green,0.36863;blue,0}] (232.62168,102.94383) circle[radius=3.35];
\path[draw={rgb,1:red,1;green,1;blue,1},line width={0.4*\leechunit},fill={rgb,1:red,0;green,0.44706;blue,0.69804}] (232.62168,155.1966) circle[radius=3.35];
\path[draw={rgb,1:red,1;green,1;blue,1},line width={0.4*\leechunit},fill={rgb,1:red,0;green,0.44706;blue,0.69804}] (281.78392,94.40843) circle[radius=3.35];
\path[draw={rgb,1:red,1;green,1;blue,1},line width={0.4*\leechunit},fill={rgb,1:red,0.83529;green,0.36863;blue,0}] (281.78392,146.6612) circle[radius=3.35];
\node[anchor=base,font={\normalsize\bfseries},text={rgb,1:red,0;green,0.44706;blue,0.69804}] at (258.5,63) {4};
\node[anchor=base,font={\small},text={rgb,1:red,0.35686;green,0.40784;blue,0.44706}] at (270,63) {+};
\node[anchor=base,font={\normalsize\bfseries},text={rgb,1:red,0.83529;green,0.36863;blue,0}] at (281.5,63) {4};
\node[anchor=base,font={\small},text={rgb,1:red,0.12549;green,0.17255;blue,0.21176}] at (270,48) {Two tetrahedra};
\node[anchor=base,font={\small},text={rgb,1:red,0.35686;green,0.40784;blue,0.44706}] at (270,21) {Tetrahedra swap};
\draw[draw={rgb,1:red,0.88235;green,0.90196;blue,0.91765},line width={0.45*\leechunit}]
  (360,17) -- (360,215);
\node[anchor=base,font={\normalsize\itshape},text={rgb,1:red,0.12549;green,0.17255;blue,0.21176}] at (450,211) {(c)  $O(VF)$};
\node[anchor=base,font={\small},text={rgb,1:red,0.35686;green,0.40784;blue,0.44706}] at (450,197) {Cube + octahedron};
\path[draw={rgb,1:red,0.83529;green,0.87059;blue,0.89804},line width={0.5*\leechunit},fill={rgb,1:red,0.98431;green,0.98824;blue,0.99216}] (450,133) circle[radius=48];
\draw[draw={rgb,1:red,0.81333;green,0.84627;blue,0.87098},line width={0.38*\leechunit}]
  (438.21608,119.3388) -- (438.21608,171.59157)
  (438.21608,119.3388) -- (483.28147,111.51469) -- (487.37832,110.8034)
  (438.21608,119.3388) -- (414.75454,104.31008) -- (412.62168,102.94383)
  (472.1654,147.19847) -- (461.37521,146.06403) -- (445.18992,144.36239) -- (434.39973,143.22795) -- (418.21444,141.52631) -- (407.42424,140.39187)
  (472.1654,147.19847) -- (457.38847,107.56467) -- (450,87.74778)
  (472.1654,147.19847) -- (457.38847,167.90097) -- (455.54135,170.48878)
  (472.1654,147.19847) -- (485.7723,132.80491)
  (438.21608,171.59157) -- (479.18462,164.47874) -- (487.37832,163.05617) -- (487.37832,110.8034)
  (438.21608,171.59157) -- (425.41888,163.39408) -- (423.28601,162.02784)
  (407.42424,140.39187) -- (446.45202,92.13478) -- (450,87.74778)
  (407.42424,140.39187) -- (432.2601,162.47708)
  (487.37832,110.8034) -- (476.71399,103.97216)
  (407.42424,140.39187) -- (414.2277,133.19509)
  (450,87.74778) -- (467.7399,103.52292)
  (450,87.74778) -- (444.45865,95.51122);
\draw[draw={rgb,1:red,0.65;green,0.80647;blue,0.89431},line width={0.65*\leechunit}]
  (438.21608,119.3388) -- (487.37832,163.05617)
  (438.21608,119.3388) -- (421.15314,143.244)
  (438.21608,119.3388) -- (453.92797,102.71855)
  (481.1486,162.40121) -- (487.37832,163.05617)
  (487.37832,163.05617) -- (485.24546,157.33553);
\draw[draw={rgb,1:red,0.94235;green,0.77902;blue,0.65},line width={0.65*\leechunit}]
  (438.21608,171.59157) -- (446.40978,161.46021) -- (458.70035,146.26317) -- (466.89405,136.13181) -- (487.37832,110.8034)
  (438.21608,171.59157) -- (427.55174,142.98835) -- (414.75454,108.66447)
  (438.21608,171.59157) -- (446.07203,163.28145)
  (487.37832,110.8034) -- (462.45944,108.18355) -- (418.8514,103.59879)
  (487.37832,110.8034) -- (478.84686,122.756);
\draw[draw={rgb,1:red,0.73333;green,0.78039;blue,0.81569},line width={0.38*\leechunit}]
  (487.37832,163.05617) -- (472.44826,153.49243) -- (461.78392,146.6612)
  (444.45865,95.51122) -- (429.68172,116.21372) -- (427.8346,118.80153)
  (476.71399,103.97216) -- (466.04966,97.14092) -- (461.78392,94.40843)
  (467.7399,103.52292) -- (492.57576,125.60813)
  (414.2277,133.19509) -- (427.8346,118.80153)
  (432.2601,162.47708) -- (442.90404,171.94217) -- (450,178.25222)
  (412.62168,107.29823) -- (412.62168,102.94383) -- (441.29965,97.96485) -- (461.78392,94.40843)
  (423.28601,162.02784) -- (412.62168,155.1966)
  (485.7723,132.80491) -- (492.57576,125.60813)
  (412.62168,111.65262) -- (412.62168,107.29823)
  (455.54135,170.48878) -- (450,178.25222)
  (412.62168,116.00702) -- (412.62168,111.65262)
  (412.62168,120.36142) -- (412.62168,116.00702)
  (412.62168,124.71582) -- (412.62168,120.36142)
  (412.62168,129.07021) -- (412.62168,124.71582)
  (412.62168,133.42461) -- (412.62168,129.07021)
  (412.62168,137.77901) -- (412.62168,133.42461)
  (412.62168,142.13341) -- (412.62168,137.77901)
  (450,178.25222) -- (489.02778,129.99513) -- (492.57576,125.60813)
  (412.62168,146.4878) -- (412.62168,142.13341)
  (450,178.25222) -- (437.07018,143.57266) -- (427.8346,118.80153)
  (412.62168,150.8422) -- (412.62168,146.4878)
  (412.62168,155.1966) -- (412.62168,150.8422)
  (412.62168,155.1966) -- (441.29965,150.21761) -- (461.78392,146.6612)
  (492.57576,125.60813) -- (454.81008,121.63761) -- (438.62479,119.93597) -- (427.8346,118.80153)
  (461.78392,94.40843) -- (461.78392,146.6612);
\draw[draw={rgb,1:red,0.91765;green,0.68431;blue,0.5},line width={0.65*\leechunit}]
  (446.07203,163.28145) -- (461.78392,146.6612)
  (414.75454,108.66447) -- (412.62168,102.94383)
  (412.62168,102.94383) -- (418.8514,103.59879)
  (412.62168,102.94383) -- (420.81538,110.23006) -- (457.68707,143.01808) -- (461.78392,146.6612)
  (478.84686,122.756) -- (461.78392,146.6612);
\draw[draw={rgb,1:red,0.5;green,0.72353;blue,0.84902},line width={0.65*\leechunit}]
  (481.1486,162.40121) -- (437.54056,157.81645) -- (425.08112,156.50653) -- (412.62168,155.1966)
  (421.15314,143.244) -- (412.62168,155.1966)
  (485.24546,157.33553) -- (472.44826,123.01165) -- (461.78392,94.40843)
  (453.92797,102.71855) -- (461.78392,94.40843)
  (412.62168,155.1966) -- (429.00909,134.93387) -- (457.68707,99.47411) -- (461.78392,94.40843);
\path[draw={rgb,1:red,1;green,1;blue,1},line width={0.4*\leechunit},fill={rgb,1:red,0.23;green,0.57424;blue,0.76749}] (438.21608,119.3388) circle[radius=3.35];
\path[draw={rgb,1:red,1;green,1;blue,1},line width={0.4*\leechunit},fill={rgb,1:red,0.23;green,0.7071;blue,0.57725}] (472.1654,147.19847) circle[radius=3.35];
\path[draw={rgb,1:red,1;green,1;blue,1},line width={0.4*\leechunit},fill={rgb,1:red,0.87318;green,0.51384;blue,0.23}] (438.21608,171.59157) circle[radius=3.35];
\path[draw={rgb,1:red,1;green,1;blue,1},line width={0.4*\leechunit},fill={rgb,1:red,0.23;green,0.7071;blue,0.57725}] (407.42424,140.39187) circle[radius=3.35];
\path[draw={rgb,1:red,1;green,1;blue,1},line width={0.4*\leechunit},fill={rgb,1:red,0.87318;green,0.51384;blue,0.23}] (487.37832,110.8034) circle[radius=3.35];
\path[draw={rgb,1:red,1;green,1;blue,1},line width={0.4*\leechunit},fill={rgb,1:red,0.23;green,0.7071;blue,0.57725}] (450,87.74778) circle[radius=3.35];
\path[draw={rgb,1:red,1;green,1;blue,1},line width={0.4*\leechunit},fill={rgb,1:red,0.23;green,0.57424;blue,0.76749}] (487.37832,163.05617) circle[radius=3.35];
\path[draw={rgb,1:red,1;green,1;blue,1},line width={0.4*\leechunit},fill={rgb,1:red,0.83529;green,0.36863;blue,0}] (412.62168,102.94383) circle[radius=3.35];
\path[draw={rgb,1:red,1;green,1;blue,1},line width={0.4*\leechunit},fill={rgb,1:red,0;green,0.61961;blue,0.45098}] (450,178.25222) circle[radius=3.35];
\path[draw={rgb,1:red,1;green,1;blue,1},line width={0.4*\leechunit},fill={rgb,1:red,0;green,0.44706;blue,0.69804}] (412.62168,155.1966) circle[radius=3.35];
\path[draw={rgb,1:red,1;green,1;blue,1},line width={0.4*\leechunit},fill={rgb,1:red,0;green,0.61961;blue,0.45098}] (492.57576,125.60813) circle[radius=3.35];
\path[draw={rgb,1:red,1;green,1;blue,1},line width={0.4*\leechunit},fill={rgb,1:red,0;green,0.44706;blue,0.69804}] (461.78392,94.40843) circle[radius=3.35];
\path[draw={rgb,1:red,1;green,1;blue,1},line width={0.4*\leechunit},fill={rgb,1:red,0;green,0.61961;blue,0.45098}] (427.8346,118.80153) circle[radius=3.35];
\path[draw={rgb,1:red,1;green,1;blue,1},line width={0.4*\leechunit},fill={rgb,1:red,0.83529;green,0.36863;blue,0}] (461.78392,146.6612) circle[radius=3.35];
\node[anchor=base,font={\normalsize\bfseries},text={rgb,1:red,0;green,0.44706;blue,0.69804}] at (427,63) {4};
\node[anchor=base,font={\small},text={rgb,1:red,0.35686;green,0.40784;blue,0.44706}] at (438.5,63) {+};
\node[anchor=base,font={\normalsize\bfseries},text={rgb,1:red,0.83529;green,0.36863;blue,0}] at (450,63) {4};
\node[anchor=base,font={\small},text={rgb,1:red,0.35686;green,0.40784;blue,0.44706}] at (461.5,63) {+};
\node[anchor=base,font={\normalsize\bfseries},text={rgb,1:red,0;green,0.61961;blue,0.45098}] at (473,63) {6};
\node[anchor=base,font={\small},text={rgb,1:red,0.12549;green,0.17255;blue,0.21176}] at (450,48) {Two tetrahedra};
\node[anchor=base,font={\small},text={rgb,1:red,0.12549;green,0.17255;blue,0.21176}] at (450,34) {+ octahedron};
\node[anchor=base,font={\small},text={rgb,1:red,0.35686;green,0.40784;blue,0.44706}] at (450,21) {Tetrahedra swap};
\node[anchor=base,font={\small},text={rgb,1:red,0.35686;green,0.40784;blue,0.44706}] at (450,7) {Octahedron preserved};
\pgfresetboundingbox
\path[use as bounding box] (0,0) rectangle (540,246);
\end{tikzpicture}
\endgroup